\documentclass{article}

\usepackage[numbers]{natbib}
\usepackage[preprint]{arXiv/neurips_2024}  
\usepackage[utf8]{inputenc}
\usepackage[T1]{fontenc}
\usepackage{url}
\usepackage{booktabs}
\usepackage{amsfonts}
\usepackage{nicefrac}
\usepackage{microtype}
\usepackage[dvipsnames]{xcolor}
\usepackage[para, bottom]{footmisc}
\usepackage{subcaption}
\usepackage{graphicx}
\usepackage{amsmath}
\usepackage{multirow}
\usepackage{makecell}
\usepackage[frozencache=true,cachedir=minted-cache]{minted} 
\usepackage[hyperfootnotes=false]{hyperref}  
\usepackage{cleveref}
\usepackage{wrapfig}
\usepackage{bm}  
\usepackage{mathtools}  
\usepackage{amsthm}  
\usepackage{algorithm}
\usepackage{algpseudocode}
\usepackage{placeins}  
\usepackage{listings}

\newcommand{\mup}{µP}
\newcommand{\umup}{u-\mup}

\newcommand{\depthmup}{depth-µP}
\newcommand{\mut}{µTransfer}
\newcommand{\mutable}{µTransferable}

\newcommand{\fanin}{\mathrm{fan{\text -}in}}
\newcommand{\fanout}{\mathrm{fan{\text -}out}}
\newcommand{\depth}{\mathrm{depth}}
\newcommand{\basewidth}{\mathrm{base{\text -}width}}
\newcommand{\basefanin}{\mathrm{base{\text -}fan{\text -}in}}

\newcommand{\basedepth}{\mathrm{base{\text -}depth}}

\newcommand{\one}{\scriptstyle{1}}
\newcommand{\batchsize}{\mathrm{batch{\text -}size}}

\newtheorem{thm}{Theorem}[section]

\newtheorem{lem}[thm]{Lemma}

\definecolor{LightGray}{gray}{0.98}
\newcommand{\codefig}[1]{
    \inputminted[
        fontsize=\small, bgcolor=LightGray, python3=true, xleftmargin=1em
    ]{python}{arXiv/code/#1}
    \vspace{-0.6cm}
}

\title{\umup: The Unit-Scaled Maximal Update Parametrization}

\author{%
  Charlie Blake\thanks{Equal contribution.} \\
  Graphcore \\
  \And
  Constantin Eichenberg\footnotemark[1] \\
  Aleph Alpha \\
  \And
  Josef Dean \\
  Graphcore \\
  \And
  Lukas Balles \\
  Aleph Alpha \\
  \And
  Luke Y. Prince \\
  Graphcore \\
  \And
  Björn Deiseroth \\
  Aleph Alpha \\
  \And
  Andres Felipe\thanks{Work done while at Aleph Alpha.}\\\textbf{Cruz-Salinas} \\
  Cohere \\
  \And
  Carlo Luschi\thanks{Supervisory role.\\Correspondence to: \href{mailto:charlieb@graphcore.ai}{charlieb@graphcore.ai}, \href{mailto:constantin.eichenberg@aleph-alpha-ip.ai}{constantin.eichenberg@aleph-alpha-ip.ai}.} \\
  Graphcore \\
  \And
  Samuel Weinbach\footnotemark[3] \\
  Aleph Alpha \\
  \And
  Douglas Orr \\
  Graphcore \\
}

\begin{document}
\maketitle
\setcounter{footnote}{0}
\vspace{-2em}

\begin{abstract}

The Maximal Update Parametrization (\mup) aims to make the optimal hyperparameters (HPs) of a model independent of its size, allowing them to be swept using a cheap proxy model rather than the full-size target model.
We present a new scheme, \umup, which improves upon \mup\ by combining it with Unit Scaling, a method for designing models that makes them easy to train in low-precision.
The two techniques have a natural affinity: \mup\ ensures that the scale of activations is independent of model size, and Unit Scaling ensures that activations, weights and gradients begin training with a scale of one.
This synthesis opens the door to a simpler scheme, whose default values are near-optimal. This in turn facilitates a more efficient sweeping strategy, with \umup\ models reaching a loss that is equal to or lower than comparable \mup\ models and working out-of-the-box in FP8.

\end{abstract}
\begin{figure}[bh!]
    \vspace{-0.3em}
    \centering
    \begin{subfigure}{\textwidth}
        \centering
        \includegraphics[width=\textwidth]{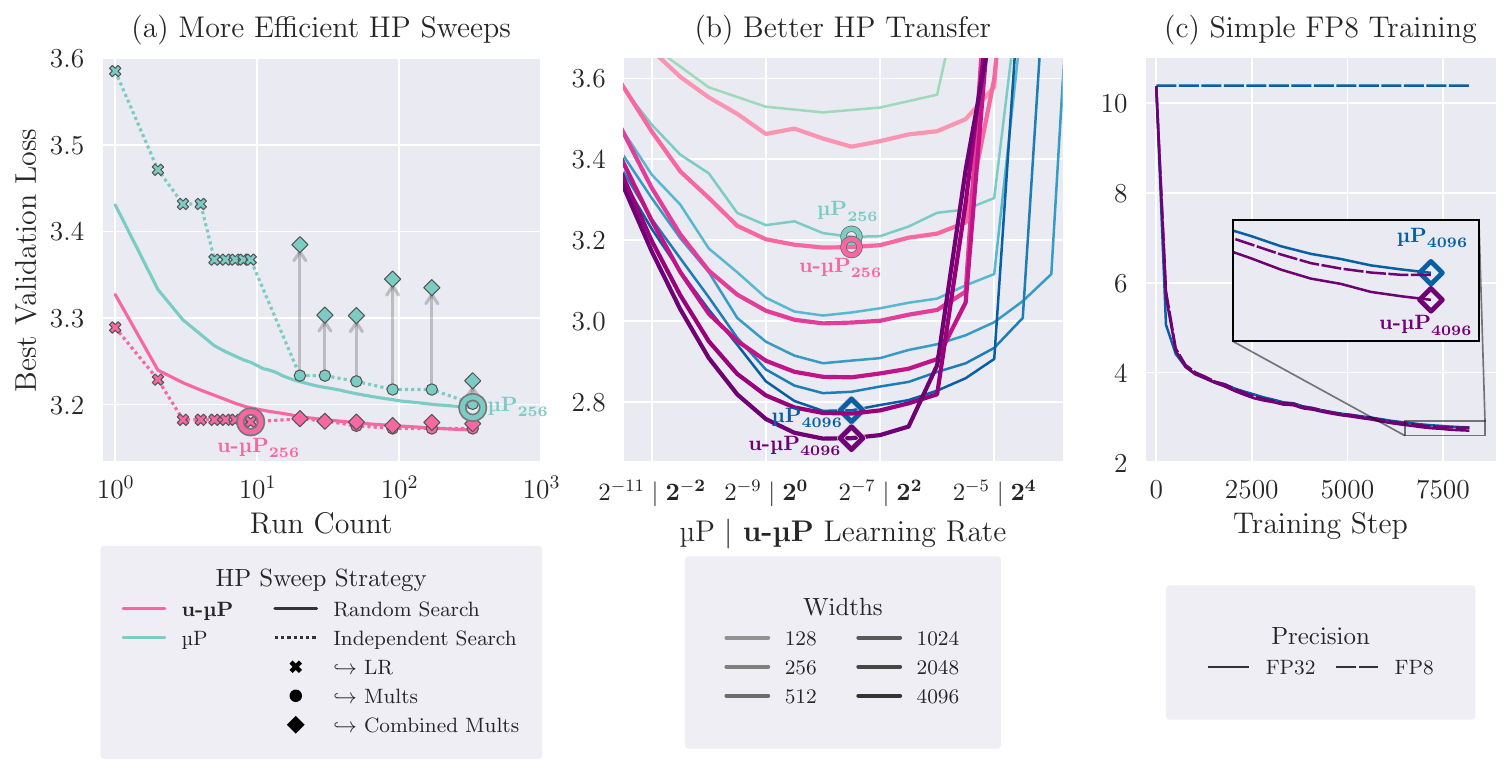}
    \end{subfigure}
    \caption{\textbf{(a)} Two different HP sweeping processes used for \mup\ and \umup\ proxy models. Unlike \mup, \umup\ admits independent (1D) search due to careful HP design. The first part of independent search is an LR sweep, which alone reaches near-optimal loss for \umup. \textbf{(b)} Using the best proxy HPs from (a), we train many models at different widths and LRs. The best LR for width 256 is \char`\~ optimal for 4096, showing LR transfer along with lower loss. \textbf{(c)} We re-train with a simple un-scaled {\texttt{.to(float8)}} cast on matmul inputs. This would fail for other models, but \umup\ trains with minimal degradation.}
    \label{fig:fig1}
    \vspace{-0.7em}
\end{figure}

\section{Introduction} \label{sec:introduction}

The challenges of large-model training extend beyond the domain of engineering; they are also \mbox{\textit{algorithmic}} in nature. Effective approaches for training smaller models are not guaranteed to work at the multi-billion-parameter scale used for today's large language models (LLMs). These difficulties can be framed in terms of stability, which we consider in three forms: 

\begin{enumerate}
    \item feature learning stability, which ensures that parts of the model do not learn too fast or slow relative to each other. 
    \item hyperparameter stability, which ensures that the optimal HPs for small models remain unchanged as the model size grows.
    \item numerical stability, which ensures that floating-point representations during training stay within the range of a given number format.
\end{enumerate}


The Maximal Update Parametrization (\mup)~\citep{Tensor_Programs_IV,Tensor_Programs_V} targets the first two sources of instability. \mup\ defines a set of scaling rules that in principle make a model's optimal HP values consistent across model sizes and ensure `maximal feature learning' in the infinite-width limit. The practical benefits of this are that models continue to improve as they get larger, and that practitioners can re-use a set of HP values (especially the learning rate) found for a small \textit{proxy} version of their model, on a larger \textit{target} model. This is vital for modern LLM training, where the cost of sweeping over candidate HP values for the target model is prohibitive. Consequently, \mup\ has been adopted by several open LLM training efforts \citep{Cerebras_GPT, BTLM, LLM360, MiniCPM} and there are indications of its use in state-of-the-art LLMs\footnotemark.

\footnotetext{The GPT-4 technical report~\citep{GPT4} hints at the use of \mup\ by including \citep{Tensor_Programs_V} in its references, without citing it directly. The multipliers present in the Grok~\citep{Grok} codebase also suggest the use of \mup.}

However, there exists a gap between the extensive theory underpinning \mup\ and its effective use in practice. This relates to issues surrounding efficient HP search, HP transfer, interpretability, ease-of-use and low-precision training. Some of these problems have been observed in the literature \citep{Exploration_Of_Mu_Transfer, Falcon, Tensor_Programs_V}; others we outline here for the first time. As a result, \mup\ does not necessarily provide the kind of simple, stable scaling for which a user might hope. 

To address this, we propose the Unit-Scaled Maximal Update Parametrization (\umup). \umup\ combines \mup\ with another closely-related training innovation, Unit Scaling \citep{Unit_Scaling}. \mup\ ideally provides consistent training dynamics across model sizes, but says little about what those dynamics should be. Unit Scaling addresses this by proposing an ideal principle for dynamics: unit variance for all activations, weights and gradients. Unit Scaling was initially designed to ensure stable numerics, but in the context of \mup\ the principle of unit-scale brings many additional benefits. We show that it provides a foundation upon which the broader range of drawbacks identified for \mup\ can be addressed.

\subsection{Contributions} \label{sec:introduction:contributions}

We focus on LLMs in this work as this is the domain where \mup\ has primarily been used in the literature (though \umup's principles should extend to other architectures). We contribute the following:

\begin{enumerate}
    \item \textbf{Drawbacks of standard \mup:} We show that the way \mup\ is typically applied has several limitations, and does not give effective transfer for Llama-style models (\Cref{sec:the_challenges_with_mup_in_practice}).

    \item \textbf{Simpler scaling rules:} \umup\ is easier to implement in practice than \mup, and removes the unnecessary `base shape' and initialization-scale HPs (\Cref{{sec:umup:combining_mup_with_us}}; \Cref{table:mup_umup_schemes}).

    \item \textbf{Out-of-the-box FP8 training:} \umup\ models generate tensors that lie close to the center of a floating point format's range, meaning that most matrix multiplications can be performed in FP8 via a simple \texttt{.to(float8)} cast without dynamic rescaling.

    \item \textbf{A principled, interpretable \& independent set of HPs:} The set of transferable HPs used in the \mup\ literature is chosen in an inconsistent and arbitrary way. We provide concrete recommendations for a good set of transferable HPs to use with \umup\ (\Cref{sec:umup:principled_hps}).

    \item \textbf{Improved HP transfer:} We identify a problem with the scaling of the embedding layer's LR under \mup. Fixing this for \umup\ gives us better scaling with width (\Cref{sec:umup:emb_lr_rule}).

    \item \textbf{A more efficient approach to HP search:} We show that \umup\ facilitates a cheaper independent search method, attaining near-optimal loss when only sweeping the LR (\Cref{sec:umup_hp_search}).
\end{enumerate}

We provide a guide for using \umup\ in \Cref{app:using_umup_guide}, and a library \citep{us_library} implementing \umup\ functions, layers and optimizers, outlined in \Cref{app:us_lib_guide}.

\section{Background} \label{sec:background}

\subsection{The Maximal Update Parametrization} \label{sec:background:the_maximal_update_parameterization}

Tensor Programs V \citep{Tensor_Programs_V} defines a parametrization as `a rule for how to change hyperparameters when the widths of a neural network change'. They show that \mup\ is the only parametrization that gives `maximal feature learning' in the limit,
whereas standard parametrization (SP) has imbalanced learning (parts of the network blow up or cease to learn).

One consequence of this improved stability is that learning dynamics under \mup\ are ideally independent of model-size, as are optimal HPs. This facilitates a method known as \mut, which describes the process of training many smaller proxy models to evaluate candidate HP values, then using the best-performing ones to train a larger target model. An HP is said to be \mutable\ if its optimal value is the same across model-sizes.

\paragraph{ABC-parametrizations} \mup, SP, and the Neural Tangent Kernel (NTK) \citep{NTK} are all instances of abc-parametrizations. This assumes a model under training where weights are defined as:
\begin{align} \label{eq:abc}
    w_0 &\sim \mathcal{N}(0,B_W^2),
    \\
    W_t &= A_W \cdot w_t, \nonumber
    \\
    w_{t+1} &= w_t + C_W \cdot \Phi_t(\nabla \mathcal{L}_0,...,\nabla \mathcal{L}_t), \nonumber
\end{align}
with $t$ a time-step and $\Phi_t(\nabla \mathcal{L}_0,...,\nabla \mathcal{L}_t)$ the weight update based on previous loss gradients.


A parametrization scheme such as \mup\ is then defined specifying how scalars $A_W,B_W,C_W$ change with model width. This can be expressed in terms of width-dependent factors $a_W, b_W, c_W$, such that $A_W \propto a_W$, $B_W \propto b_W$, $C_W \propto c_W$. The values these factors take are what characterize a particular scheme. For \mup\ these are given in \Cref{table:mup}.
For depth a similar result has been proved using \depthmup\ \citep{Tensor_Programs_VI}, albeit in a restricted setting. When we refer to \mup\ in the paper we assume the \depthmup\ scaling rules (\Cref{table:mup_umup_schemes}, `Residual' column).

A key property of the abc-parametrization is that one can shift scales between $A_W,B_W,C_W$ in a way that preserves learning dynamics (i.e. the activations computed during training are unchanged). We term this \textit{abc-symmetry}. For a fixed $\theta > 0$, the behavior of a network trained with Adam is invariant to changes of the kind:
\begin{align} \label{eq:abc_symmetry}
    A_W \leftarrow A_W \cdot \theta, \quad B_W \leftarrow B_W / \theta, \quad C_W \leftarrow C_W / \theta
\end{align}
(reproduced from Tensor Programs V, Section~J.2.1). This means that parametrizations like \mup\ can be presented in different but equivalent ways. ABC-symmetry is a key component in developing \umup.

\begin{table}[b]
  \vspace{-0.5em}
  \centering
  \caption{The scaling rules defining \mup. The type of a weight is determined by whether $\fanin$ \& $\fanout$ both depend on width (hidden), only $\fanout$ does (input), or only $\fanin$ (output). Hence $\fanin$ is always a multiple of width here.}
  \vspace{0.5em}
  \label{table:mup}
  \begin{tabular}{rl @{\hspace{0.8\tabcolsep}} lccc}
      \toprule
      & \multicolumn{2}{c}{\multirow{2}{*}{ABC-multiplier}} & \multicolumn{3}{c}{Weight ($W$) Type}
      \\
      \rule{0pt}{1em} & & & Input & Hidden & Output
      \\  
      \midrule
      \multirow{3}{*}{\textbf{\mup}} & parameter & ($a_W$) & $\one$ & $\one$ & $\nicefrac \one {\fanin(W)}$
      \\
      & initialization & ($b_W$) & $\one$ & $\nicefrac{\one}{\sqrt{\fanin(W)}}$ & $\one$
      \\
      & Adam LR & ($c_W$) & $\one$ & $\nicefrac \one {\fanin(W)}$ & $\one$
      \\
      \bottomrule
      \\
  \end{tabular}
\end{table}

\paragraph{Transferable HPs} \mup\ focuses on the subset of HPs whose optimal values we expect to \textit{transfer across} axes such as width and depth. We term these \mutable\ HPs. All \mutable\ HPs function as multipliers and can be split into three kinds, which contribute to the three (non-HP) multipliers given by the abc-parametrization: $\alpha_W, \sigma_W, \eta_W$ where $A_W \propto \alpha_W, B_W \propto \sigma_W, C_W \propto \eta_W$. The difference between these multipliers and the ones that define a parametrization is that they are specified by the user, rather than being a function of width. $\alpha_W$ and $ \eta_W$ are rarely introduced outside of the \mup\ literature, but can be valuable to tune for both \mup\ and SP models. In the \mup\ literature the term `HPs' often implicitly refers to \mutable\ HPs. We adopt this convention here, unless specified otherwise.


\paragraph{Base shape} Two additional (non-\mutable) HPs introduced by \mup\ are the $\basewidth$ and $\basedepth$. This refers to a mechanism where a user specifies a particular shape for the model, where its behavior under \mup\ and SP are the same. The \mup\ model still \textit{scales} according to the abc-rules, so for all other shapes the two models will be different. This is implemented by dividing the \mup\ scaling rules for the given model by those of a fixed-shape model at the $\basewidth$ and $\basedepth$.

Putting this together with our abc-parametrization given in \Cref{eq:abc}, and the \mutable\ HPs outlined above, we now derive our final, absolute expressions for $A_W,B_W,C_W$:
\begin{align} \label{eq:abc_mup_absolute}
    A_W \leftarrow \alpha_W \frac{a_W}{a_{W_\textrm{base}}},
    \quad
    B_W \leftarrow \sigma_W \frac{b_W}{b_{W_\textrm{base}}},
    \quad
    C_W \leftarrow \eta_W \frac{c_W}{c_{W_\textrm{base}}}
\end{align}
Though base shapes are necessary for \mup, they are not typically swept. Rather, they are considered a preference of the user, who may wish to retain the behavior of an existing SP model at a given shape.

\paragraph{Choosing HPs to sweep} 

In theory, the search space of \mutable\ HPs includes $\alpha_W, \sigma_W, \eta_W$ for every parameter tensor $W$ in the model. In practice far fewer HPs are swept, with global grouping often used for $\sigma_W$ and $\eta_W$, and many $\alpha_W$s dropped or grouped across layers.

The sets of HPs chosen for sweeps in the \mup\ literature is explored in \Cref{app:additional_background:mup}. Tensor Programs V uses a random search to identify the best HP values, which has become the standard approach to sweeping. The number of runs in a sweep is typically in the low 100s, incurring a non-negligible cost (though usually less than a single training run of the target model). This high number partly owes to dependencies between HPs (shown in \Cref{sec:experiments:hp_independence}), making the search space hard to explore.

\subsection{Low-precision training} \label{sec:background:low_precision_training}

All the major potential bottlenecks of model training---compute, communication and storage---see roughly linear improvements as the bit-width of their number format is reduced.
In modern LLM training, the compute cost of large matrix multiplications (matmuls) means that substantial gains are available if these can be done in low-precision ($<32$ bit) formats.
With the ending of Dennard scaling and Moore's law \citep{Moores_Law_A,Moores_Law_B}, the use of low-precision formats represents one of the most promising avenues towards increased efficiency in deep learning.

Recent AI hardware offers substantial acceleration for the 8-bit FP8 E4 and E5 formats. However the reduced range of these formats means that they cannot directly represent some values generated during training. Various methods have been introduced to address this, such as the per-tensor dynamic re-scaling in Transformer Engine \citep{Transformer_Engine}. However, this comes at the cost of added complexity and potential overheads. For a more in-depth treatment of low-precision formats, see \Cref{app:low_precision_and_its_trade_offs}.

\subsection{Unit Scaling} \label{sec:background:unit_scaling}

An alternative approach to low-precision training is Unit Scaling \citep{Unit_Scaling}, which also uses fine-grained scaling factors to control range, but instead finds these factors via an analysis of expected tensor statistics at initialization. These are fixed factors, calculated independently of the contents of a tensor, at the beginning of training. As such, the method is easy to use and only adds the overhead of applying static scaling factors (which we show to be negligible in \Cref{app:scaled_mm_benchmarking}).

These factors are chosen to ensure the unit variance of activations, weights and gradients at initialization.
This is a useful criterion as it places values around the center of floating-point formats' absolute range. This applies to all tensors, meaning every operation in the network requires a scaling factor that ensures unit-scaled outputs, assuming unit-scaled inputs. Unit Scaling does not provide a mechanism for re-scaling tensors dynamically during training, but due to its ideal starting scale for gradients, activations and weights this may not be required. Empirically this is shown to be true across multiple architectures, though it is not guaranteed.

We provide an example of deriving the Unit Scaling rule for a matmul op in \Cref{app:additional_background:us}, resulting in the scaling factor: $1 / \sqrt{d_\fanin}$. We accompany this example with a full recipe for applying Unit Scaling to an arbitrary model.

\section{The challenges with \mup\ in practice} \label{sec:the_challenges_with_mup_in_practice}


\footnotetext{As in other work, we use \mup\ as a shorthand for the method outlined in Tensor Programs V, including \mut. Strictly speaking, \mup\ ought only to refer to the parametrization outlined in Tensor Programs IV.}

\subsection{Not all training setups give \mut} \label{sec:challenges:mut}

Lingle \citep{Exploration_Of_Mu_Transfer} shows that directly applying \mup\ to a decoder LM fails to provide LR transfer across width. Given that the primary use of \mup\ in the literature has been LM training of this kind, this result suggests a significant limitation. How do we reconcile this with the strong LR transfer across width shown for language models in Tensor Programs V?


\begin{figure}
    \centering
    \begin{subfigure}{\textwidth}
        \centering
        \includegraphics[width=\textwidth]{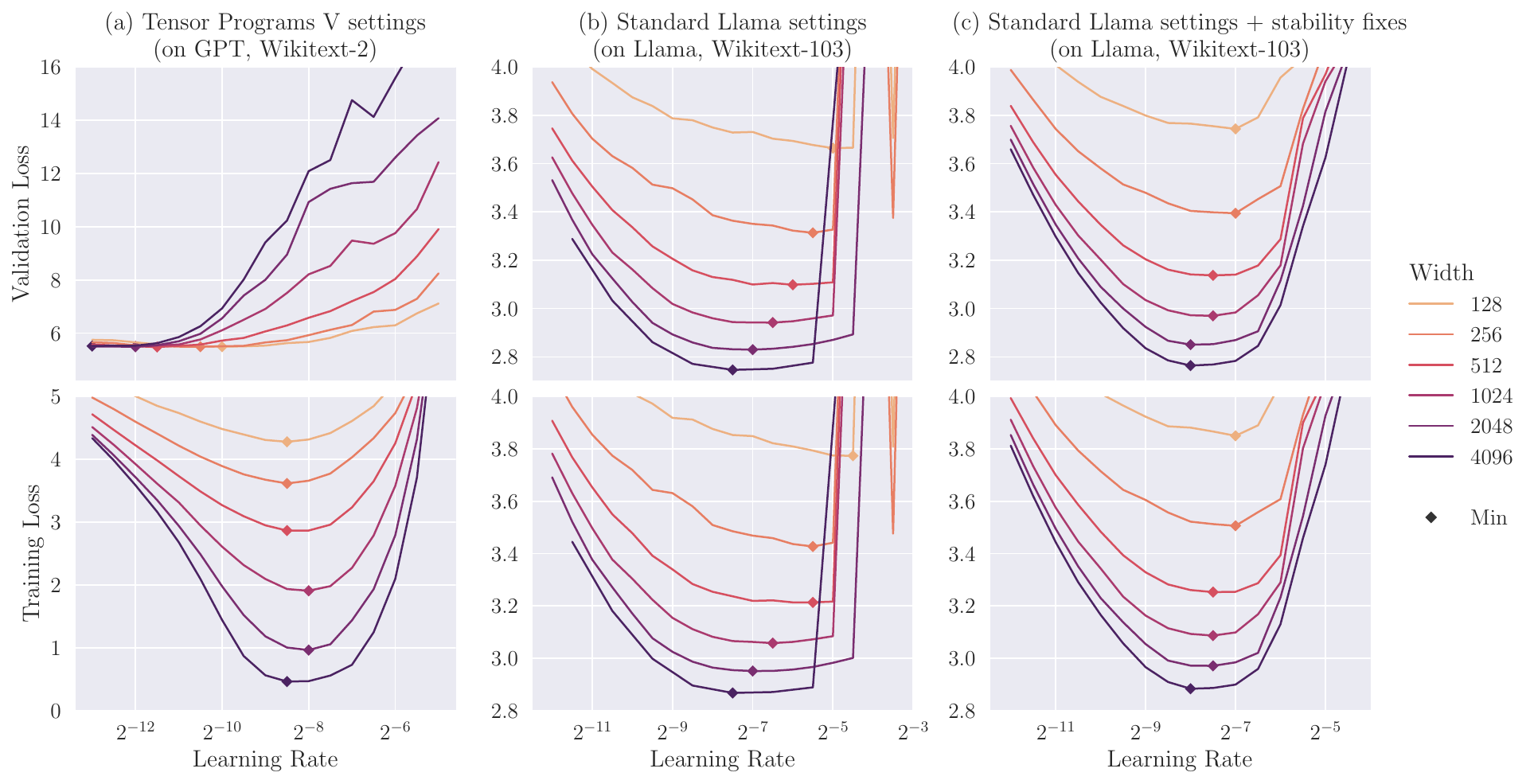}
    \end{subfigure}
    \caption{Effective \mut\ does not hold across all training setups.
    \textbf{(a)} We show strong transfer for the unrealistic setup used in Tensor Programs V (too many epochs; constant LR). \textbf{(b)} Moving to a more standard Llama training setup, transfer breaks down. \textbf{(c)} This is restored by the introduction of two stability fixes: non-parametric norms and independent weight decay.}
    \label{fig:experiments:tpv_setup}
\end{figure}

We answer this in \Cref{fig:experiments:tpv_setup}. The first training setup (a) is aligned with that used in Tensor Programs V (their Figure 4). There are several atypical aspects to their training setup, primarily the use of a constant LR schedule and a high number of epochs; we outline the precise differences between setup (a) and (b) in \Cref{table:experiments:tpv_vs_llama_settings}. This overfitting regime makes validation loss unusable, and transfer misleadingly good. When we remove these and shift to a standard Llama training setup (b), optimal HPs begin to drift with width (see \Cref{fig:experiments:tpv_setup_standard_ablation} for an ablation). This confirms Lingle's findings that standard \mup\ is in fact a poor fit for modern LM training. We fix this (c) by the removal of parameters from LayerNorms/RMSNorms, as suggested by Lingle, and the introduction of \textit{independent} weight decay for AdamW, as suggested by Wortsman et al. \citep{Small_Scale_Proxies} \footnote{Lingle suggests independent weight decay is unstable, but we find it to be more so than Adam or standard AdamW.} (see \citep{AdamW_Weight_Decay} for further analysis). With these changes adopted, we recover the strong transfer shown in Tensor Programs V's experiments.


\subsection{It's not clear which hyperparameters to sweep} \label{sec:challenges:which_hps}

The problem of selecting HPs to sweep can be framed as choosing a subset of the per-tensor $\alpha_W, \sigma_W, \eta_W$ HPs outlined in \Cref{sec:background:the_maximal_update_parameterization}, and grouping across/within layers. As shown in \Cref{table:mup_hps}, \mut\ experiments in the literature have done this in a variety ways. Practitioners have not justified these choices, appearing to rely on a mixture of precedent and intuition. We outline two major downsides to the lack of a principled approach.

Firstly, not all groupings of HPs are suitable. Consider the commonly-used global $\sigma_\mathrm{init}$ HP. At initialization the activations going into the FFN swish function have $\operatorname{std}(x_\mathrm{swish}) \propto \sigma_{W_\mathrm{gate}}$, whereas the self-attention softmax activations have $\operatorname{std}(x_\mathrm{attn}) \propto \sigma_{W_\mathrm{Q}}\sigma_{W_\mathrm{K}}$. A global $\sigma$ HP thus has a linear effect on the FFN and a quadratic effect on attention, suggesting that this grouping may be unideal.

Secondly, not all HPs are independent of one another. The key example of this is the interaction between $\sigma_W$ and $\eta_W$. The relative size of a weight update is determined by the ratio $\eta_W / \sigma_W$, not by either HP individually. Because of this, the optimal values for $\sigma$ and $\eta$ depend on each other, which we demonstrate empirically in \Cref{sec:experiments:hp_independence}. This can make the problem of HP search much harder, and may be why hundreds of random-search runs have been required for sweeps in the literature.


\subsection{Base shape complicates usage}

Most practitioners are unlikely to require alignment with an SP model, in which case it is unclear what $\basewidth$ (and $\basedepth$) should be used. The literature has aligned on a standard $\basewidth$ of $256$ (see \Cref{table:mup_hps}), but this appears to lacking a principled motivation---though the fact that they are not dropped entirely suggests they may be beneficial under \umup.

Implementing base-shape HPs (see \Cref{eq:abc_mup_absolute}) can also add complications from an engineering perspective. The proposed implementation in the $\texttt{mup}$ library \citep{Mup_Library} reflects this, requiring an extra `base' model to be created and the original model to be re-initialized. This can interact awkwardly with other model-transforms for features like quantization, compilation, etc:

\vspace{0.3em}
\codefig{base_shapes.py}


\subsection{\mup\ appears to struggle with low-precision}

Finally, we note an interesting contradiction observed in the relationship between \mup\ and low-precision. One of the stated aims for \mup\ is that its activations have $\Theta(1)$-sized coordinates in the limit \citep[Desiderata J.1]{Tensor_Programs_V}. This desideratum is specifically given in order that values can be represented using finite-range floating-point numbers \citep[Section 3]{Tensor_Programs_IV}.
Yet despite numerical stability being central to the theory underlying \mup, this is not leveraged to ensure that \mup\ models can \textit{actually} be trained in low-precision. Indeed, for the LLM runs in Tensor Programs V the SP model trains successfully in FP16, while the \mup\ model diverges (attributed to underflow of gradients). We remedy this with \umup.
\section{The Unit-Scaled Maximal Update Parametrization} \label{sec:umup}

In this section we show how \mup\ can be adapted to satisfy Unit Scaling, and provide a new set of HPs which---thanks to Unit Scaling---are more interpretable and separable than those commonly used for \mup, unlocking several practical benefits. For those wishing to apply \umup\ to their own models, we provide a user-guide in \Cref{app:using_umup_guide} and an overview of our library implementing \umup\ in \Cref{app:us_lib_guide}.


\subsection{Combining \mup\ with Unit Scaling} \label{sec:umup:combining_mup_with_us}

Whereas Unit Scaling provides rules for scaling all operations, \mup\ only does so for parametrized ones. It's these operations we need to address to arrive at a unified scheme, resolving differences in the scaling rules each recommends. We begin with the expressions for the $A_W,B_W,C_W$ scaling factors in \Cref{eq:abc_mup_absolute}, and substitute in the \mup\ scaling rules defined in \Cref{table:mup}. This results in a complete implementation of \mup, which is shown in the top half of \Cref{table:mup_umup_schemes} (using the \textit{extended} set of \mup\ HPs given in \Cref{table:hp_sets}). We set out to turn this into a valid Unit Scaling scheme, which requires unit initializations ($B_W \leftarrow 1$) and matmuls with the Unit Scaling factor we identified in \Cref{sec:background:unit_scaling} ($A_W \leftarrow 1/\sqrt{\fanin}$).

Our first step is to drop the $\sigma_W$ and $\basefanin$ HPs entirely, and associate the $\alpha_W$ HPs with certain functions instead of weights---decisions we justify in the rest of this section (this results in the simplified intermediate implementation in \Cref{table:mup_to_umup_1}). Our input weights now have unit initializations as desired, and a unit parameter multiplier, which is also the appropriate scaling factor (as input layers here are embedding lookups, not matmuls).

Hidden weights now have the implementation:
\begin{align} \label{eq:mup_comparison}
    A_W \leftarrow 1,
    \quad
    B_W \leftarrow \frac{1}{\sqrt{\fanin}},
    \quad
    C_W \leftarrow \eta \, \frac{1}{\fanin},
\end{align}
which differs from our Unit Scaling criteria. However, using the abc-symmetry outlined in \Cref{eq:abc_symmetry} we can shift scales by a factor of $\sqrt{\fanin}$, arriving at a unit-scaled scheme:
\begin{align} \label{eq:mup_comparison_2}
    A_W \leftarrow \frac{1}{\sqrt{\fanin}},
    \quad
    B_W \leftarrow 1,
    \quad
    C_W \leftarrow \eta \, \frac{1}{\sqrt{\fanin}}.
\end{align}
Finally, our output layers also have unit initialization, but a parameter multiplier of $A_W \leftarrow 1/\fanin$. This differs from the Unit Scaling rule, but in the forward pass this is permissible as there are no subsequent matmuls of a transformer. In the backward pass this mis-scaling would propagate, so we apply the desired $\leftarrow 1/\sqrt{\fanin}$ factor. Using different forward and backward scales in this way is usually not allowed, but is valid for output layers due to the cut-edge rule (\Cref{app:cut_edge_rule}).

The final change we make is to the input LR scaling rule, which we show in \Cref{sec:umup:emb_lr_rule} is more effective if $c_W \leftarrow 1$ is replaced with $c_W \leftarrow 1 / \sqrt{\fanout}$~\footnote{This represents a slight deviation from the Maximal Update Parametrization, though we still refer to our scheme as a form of \mup\ as it conforms in all other aspects.}. With these changes made, we arrive at our final \umup\ scheme, given in \Cref{table:mup_umup_schemes}. It's important to note that the scaling rules in this table must be combined with the standard Unit Scaling rules for other non-matmul operations. These are covered in \Cref{app:additional_unit_scaled_ops}, and implemented in our library (see \Cref{app:us_lib_guide}).

\begin{table}[t]
  \centering
  \caption{The definition of \umup\, along with an implementation of \mup\ (assuming the \textit{extended} HP set in \Cref{table:hp_sets}). \umup\ aims to simplify \mup\ and provide the benefits of Unit Scaling.}
  \vspace{0.2em}
  \label{table:mup_umup_schemes}
  \begin{tabular}{rl @{\hspace{0.8\tabcolsep}} lcccc}
      \toprule
      & \multicolumn{2}{c}{\multirow{2}{*}[-0.2em]{ABC-multiplier}} & & Weight Type \vspace{0.2em} & & \multirow{2}{*}[-0.2em]{Residual}
      \\\cline{4-6}
      \rule{0pt}{1em} & & & Input & Hidden & Output &
      \\
      \midrule
      & parameter & ($A_W$) & $\alpha_{\mathrm{emb}}$ & \hspace{1em} $1$ (or $\alpha_\mathrm{attn}$) & $\alpha_{\mathrm{out}} \frac \basefanin {\fanin}$ & $\sqrt{\frac \basedepth {\depth}}$\textsuperscript{*}
      \\
      \textbf{\mup} & initialization & ($B_W$) & $\sigma_{\mathrm{init}}$ & $\sigma_{\mathrm{init}} \sqrt{\frac \basefanin {\fanin}}$ & $\sigma_{\mathrm{init}} $ & ---
      \\
      & Adam LR & ($C_W$) & $\eta \, \hat \eta_\mathrm{emb}$ & $\eta \, \frac \basefanin {\fanin}$ & $\eta$ & $\sqrt{\frac \basedepth {\depth}}$\phantom{\textsuperscript{*}}
      \\
      \midrule
      & parameter\textsuperscript{†} & ($A_W$) & $1$ & $\frac 1 {\sqrt{\fanin}}$ & $\frac 1 {\fanin}$\textsuperscript{‡} & $\frac 1 {\sqrt{\depth}}$\textsuperscript{*}
      \rule{0pt}{1.2em}\\[0.75em]
      \textbf{\umup} & initialization & ($B_W$) & $1$ & $1$ & $1$ & ---
      \\[0.65em]
      & Adam LR & ($C_W$) & $\eta \, \frac 1 {\sqrt{\fanout}}$ & $\eta \, \frac 1 {\sqrt{\fanin}}$ & $\eta$ & $\frac 1 {\sqrt{\depth}}$\phantom{\textsuperscript{*}}
      \\[0.25em]
      \bottomrule
      \vspace{-0.4cm}
      \\
      \multicolumn{7}{l}{\small{\textsuperscript{*}Residual multipliers are applied to the end of each branch, rather than the output of linear layers.}}
      \rule{0pt}{1.2em}\\
      \multicolumn{7}{l}{\small{\textsuperscript{†}\umup's $\alpha$ HPs are associated with operations, not weights, so are not included here (see \Cref{sec:umup:principled_hps}).}}
      \\
      \multicolumn{7}{l}{\small{\textsuperscript{‡}To maintain unit scale we apply $ 1 / \sqrt{\fanout}$ scaling in the backward pass (see \Cref{app:cut_edge_rule}).}}
      \\
  \end{tabular}
  \vspace{-0.4em}
\end{table}

\subsection{Out-of-the-box low-precision training} \label{sec:umup:low_prec_training}

By applying the principles of Unit Scaling to \mup, \umup\ gains a key feature: the easy use of low-precision number formats during training. We can attribute the difficulties \mup\ has with low precision to the fact that it ignores constant factors (along with weight and gradient-scaling), only ensuring that activations are \textit{of order} $\Theta(1)$. The stricter condition of unit scale across all tensors at initialization provides a way of leveraging \mup's rules in order to make low-precision training work.


When training a transformer model with \umup\, most scales in the model stabilize while certain tensors exhibit scale growth that potentially pushes them out of FP8 range. We empirically identify these \textit{critical tensors} to be the inputs to the attention dense projection and final FFN matmul as well as the weight of the decoder head (for details see \Cref{app:fp8_training}). The latter becomes negligible in terms of model flops as width and depth of the model increase, so we generally keep this operation in higher precision. 

Following these observations, we propose the following FP8 mixed precision scheme for \umup\ transformer models:
\begin{itemize}
    \item For non-critical matmul operations, we cast the input and weight to \texttt{E4M3}, and the gradient with respect to the output to \texttt{E5M2}. This is done in the forward computation, as well as the two backward computations (for the gradient w.r.t. the weight, respectively the input). Non-critical layers are query, key, value as well as the input layer(s) to the FFN.
    \item All layers involving critical tensors, as well as embedding layer, residual addition and nonlinear functions are performed in higher precision. This also means that we directly aggregate into higher precision in each FP8 matmul. We keep optimizer states in FP32, as is usually the case in mixed precision training. 
\end{itemize}
We note that in some cases one can deal with the critical tensors by casting them to \texttt{E5M2} instead of \texttt{E4M3}, however we observed some instabilities applying this in a large scale setting, possibly due to loss of precision. In small scale scenarios we also empirically find that applying the \texttt{E4M3} format instead of \texttt{E5M2} for the gradients is possible, but becomes problematic in a more realistic setting where gradients require a higher dynamic range. 

With our proposed mixed precision scheme, about $70$\% of the matmul computations in the transformer block are performed natively in FP8 (assuming a standard architecture, e.g. Llama). If desired, a dynamic per-tensor scaling could still be applied to the critical tensors.



\subsection{A principled approach to hyperparameters} \label{sec:umup:principled_hps}

We saw in \Cref{sec:challenges:which_hps} that approaches for selecting which HPs to sweep are poorly motivated in the literature. Our objective in \umup\ is to find a simple, well-justified and effective alternative. To this end, we propose the following ideal criteria:

\begin{enumerate}
    \item \textbf{Minimal cardinality}: the use of as few HPs as possible.
    \item \textbf{Maximal expressivity}: the ability to still express any model defined using the per-tensor $\alpha_W, \sigma_W, \eta_W$ HPs outlined in \Cref{sec:background:the_maximal_update_parameterization} (in practice, we relax this slightly).
    \item \textbf{Minimal interdependency}: the optimal value of each HP should not depend on the value of other HPs, simplifying the search space.
    \item \textbf{Interpretability}: there should be a clear explanation for what an HP's value `means' in the context of the model.
\end{enumerate}

\begin{wraptable}{r}{17.6em}
    \vspace{-0.75em}
    \caption{Typical transformer HPs used under different schemes. \textit{Basic} HPs in \textbf{bold} are considered most impactful and are commonly swept. \textit{Extended} HPs in non-bold are not always swept, often set heuristically or dropped.}
    \label{table:hp_sets}
    \vspace{-1em}
    \begin{tabular}{ccc}\\\toprule  
        SP & \mup\ & \umup\ \\\midrule
        {$\boldsymbol{\eta}$} & {$\boldsymbol{\eta}$} & {$\boldsymbol{\eta}$}
        \\ 
        $\sigma\text{-}\mathrm{scheme}$ & $\boldsymbol{\sigma_\mathrm{init}}$ &
        \\
          & $\boldsymbol{\alpha_\mathrm{emb}}|\boldsymbol{\eta_\mathrm{emb}}$ & $\alpha_\mathrm{ffn\text{-}act}$
        \\
          & $\alpha_\mathrm{attn}$ & $\alpha_\mathrm{attn\text{-}softmax}$
        \\
          & $\alpha_\mathrm{out}$ & $\alpha_\mathrm{res}$
        \\
          & $\basewidth$ & $\alpha_\mathrm{res\text{-}attn\text{-}ratio}$
        \\
          & $\basedepth$ & $\alpha_\mathrm{loss\text{-}softmax}$
        \\
        \bottomrule
    \end{tabular}
    \vspace{-2em}
\end{wraptable}

The \umup\ HPs given in \Cref{table:hp_sets} are designed to satisfy these criteria, to the fullest extent possible. The placement of these HPs in the model is given in \Cref{tab:ops_compendium}.

\paragraph{Cardinality \& expressivity} We arrive at our set of HPs in three steps, starting with the full $\alpha_W, \sigma_W, \eta_W$ for each weight tensor $W$. Firstly, we can choose to `drop' any one of these three HPs by permuting under abc-symmetry, such that one HP $=1$. As we want our weights to begin with unit scale, we choose $\sigma_W$ (i.e. $\theta = \sigma_W$ in \Cref{eq:abc_symmetry}), leaving just $\alpha_W, \eta_W$.

Secondly, we observe that several of the $\alpha_W$ HPs combine linearly with other $\alpha_W$ HPs, providing an opportunity to re-parametrize with a single HP. For instance, we noted in \Cref{sec:the_challenges_with_mup_in_practice} that the scale of self-attention softmax activations is proportional to the product of $\sigma_W$ multipliers, and the same is true for $\alpha_W$ multipliers: $\operatorname{std}(x_\mathrm{attn}) \propto \alpha_{W_\mathrm{Q}}\alpha_{W_\mathrm{K}}$. In this instance it appears more natural to use a single $\alpha$ parameter and associate it with the attention operation, rather than the weights. We term this $\alpha_\mathrm{attn\text{-}softmax}$.

We apply the same principle to the rest of the model, associating $\alpha$ HPs with operations instead of weights. This applies to all operations, unless they are unary and $k$-homogeneous for $k \ge 0$, in which case they propagate scale and don't require an HP (see \Cref{app:non-homogeneous}). This results in the set of HPs shown, with their placement in the model given in \Cref{tab:ops_compendium}.

Thirdly, we use a single global $\eta$ and group $\alpha$ HPs across layers. This breaks our expressivity criterion, but we argue represents the best trade-off between expressivity and cardinality. We show in \Cref{app:umup_lr_mults} that having tuned a global $\eta$ HP and our extended $\alpha$ HPs, the further benefits of tuning per-tensor $\hat{\eta}_W$ HPs (which modify the global $\eta$) is minimal, justifying our decision to only use one global $\eta$.

\paragraph{Interdependency}

The second stage above, moving $\alpha$ HPs from weights into subsequent operations, not only reduces the number of HPs, but also minimizes the interdependence between those that remain. Interactions between HPs are complex and unlikely to be entirely separable, but we find that \umup's optimal HP values depend less on each other than under \mup\ (see \Cref{sec:experiments:hp_independence}).

\paragraph{Interpretability} 

The combination of unit scale and reduced dependencies between HPs means that each $\alpha$ can be interpreted as determining some fundamental property of the model at initialization. For example, the $\alpha_\mathrm{loss\text{-}softmax}$ HP defines the (inverse of) the softmax's \textit{temperature} for a unit-scaled input. We also introduce a new scaling scheme (defined in \Cref{subsubsec:umup_residual_in_full}) for residual connections, designed to give HPs independence and a clear interpretation: $\alpha_\mathrm{res}$ defines the contribution of the residual connections to the output scale, and $\alpha_\mathrm{res\text{-}attn\text{-}ratio}$ defines the relative contribution of attention versus FFN branches.
Finally, we choose not to include base shape HPs in \umup. They do not add to expressivity, lack a clear interpretation (besides alignment to a base model at a particular shape), break the interpretations of other HPs (as given above), and complicate implementation.

\subsection{A new embedding LR rule} \label{sec:umup:emb_lr_rule}

\begin{figure}[t]
    \centering
    \begin{subfigure}{\textwidth}
        \centering
        \includegraphics[width=\textwidth]{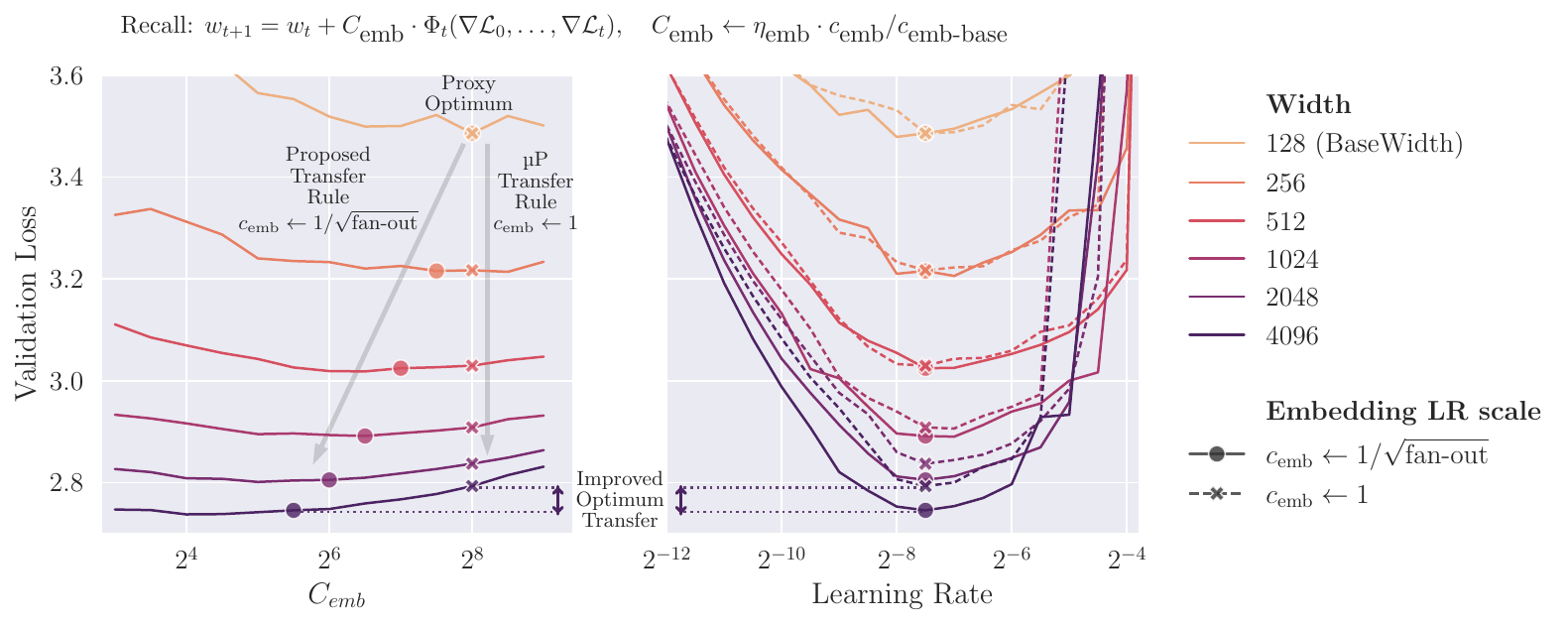}
    \end{subfigure}
    \caption{(Left) holding the embedding LR ($\hat{\eta}_\mathrm{emb}$) constant, vs. scaling with $\sqrt{\nicefrac{\basewidth}{\mathrm{width}}}$, both with a fixed global LR. This suggests the \mup\ embedding LR rule ($c_\mathrm{emb}$) should follow the latter scaling. (Right) we test this by sweeping the global LR under the two scaling rules. The new rule leads to lower loss on large models. (Dot/cross markers represent the same runs across both graphs).}
    \label{fig:umup:emb_lr_rule}
\end{figure}

Although theoretical transfer properties have been proved for \mup, not all its HPs have had \mut\ shown empirically. We do so for the \textit{extended} \mup\ transformer HPs in \Cref{{fig:experiments:hp_transfer_over_width}}, where we observe poor transfer across width for the embedding LR multiplier $\hat{\eta}_\mathrm{emb}$. The associated scaling rule for the embedding LR is constant in width ($c_\mathrm{emb} = 1$), but this poor multiplier transfer suggests the rule is mis-specified. We show in \Cref{fig:umup:emb_lr_rule} (left) that a more effective rule is $c_\mathrm{emb} = 1 / \sqrt{\fanout}$. 

This keeps the optimal value of $\hat{\eta}_\mathrm{emb}$ the same regardless of width. \Cref{fig:umup:emb_lr_rule} (right) shows that a constant scaling rule leads to diminishing returns as width increases, whereas our new rule continues to work well at scale, attaining the same loss at 2048-width that constant scaling attains at 4096-width.
Our adoption of this change is a key factor in the improved performance of \umup\ over \mup\ in \Cref{fig:fig1}. 
We offer no theoretical justification for our rule, which we leave to further work.

\subsection{Hyperparameter search} \label{sec:umup_hp_search}

As shown in section \Cref{sec:background:the_maximal_update_parameterization}, the standard approach to HP search for \mut\ is via a random sweep over all HPs simultaneously. Sweeping individual HPs separately is challenging due to the dependencies between them. In contrast, \umup's HPs are designed to admit such a strategy due to our interdependence criterion. Because of this, we propose a simpler sweeping strategy for \umup\ which we term \textit{independent search} (outlined in detail in \Cref{app:umup_hp_search_algorithm}).

Independent search involves a sweep of the LR, followed by a set of one-dimensional sweeps of the other HPs (which can be run in parallel). The best results from the individual sweeps are combined to form the final set of HP values. We also consider an even simpler scheme, which only sweeps the LR, leaving other HP values at 1 (i.e. dropping them). For caution, we recommend the full approach, but in practice we find that only sweeping the LR is surprisingly effective, as shown in \Cref{fig:fig1} (a). This indicates that not only is the principle of unit scale good for numerics, but also for learning dynamics where it provides near-optimal scaling.

\section{Experiments}

\subsection{Experimental setup}

Our experiments all use the Llama \citep{Llama} architecture trained on WikiText-103~\citep{WikiText103} (excepting the large-scale runs in \Cref{sec:fp8_training}). We apply current best-practice LLM training techniques from the literature (full settings are given in \Cref{tab:experiment_defaults}). In accordance with our analysis of settings for \mut\ in \Cref{sec:challenges:mut}, we remove parameters from norm layers, use independent AdamW, and avoid training on too many epochs for both \umup\ and \mup\ for the sake of fair comparison.

\subsection{Quantifying hyperparameter interdependence} \label{sec:experiments:hp_independence}

Our principled approach to HPs (\Cref{sec:umup:principled_hps}) contains the requirement that their optimal values should depend minimally on the value of other HPs. We now investigate this empirically, conducting a 2D sweep over every pair of HPs for \mup\ and \umup, shown in \Cref{fig:additional_experiments:mult_grid_mup,,fig:additional_experiments:mult_grid_umup} respectively.

\begin{figure}[t]
    \vspace{-1em}
    \centering
    \includegraphics[width=\textwidth]{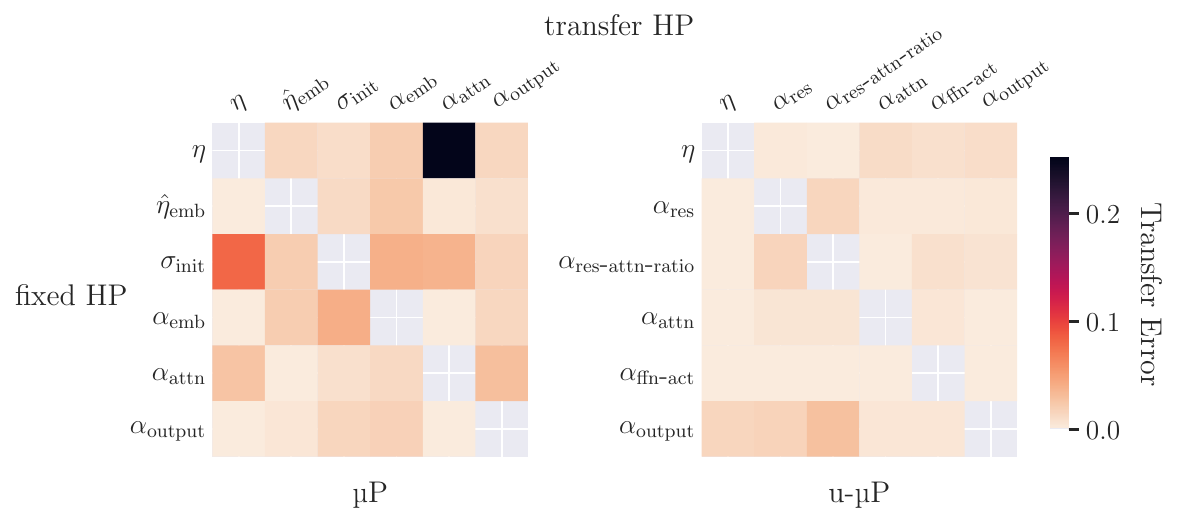}
    \caption{A visualization of the dependencies between pairs of HPs under each scheme. Transfer error measures the extent to which the optimal value of the transfer HP depends on the fixed HP (see \Cref{alg:transfer_error}). On average, \mup\ has a transfer error of 0.03, whereas \umup\ has 0.005.}
    \label{fig:experiments:transfer_error}
\end{figure}

To derive an empirical measure of HP dependency, we introduce the notion of \textit{transfer error} (see \Cref{alg:transfer_error}). This considers a pair of HPs, with one `fixed' and the other for `transfer'. We take the best value of the transfer HP for each non-optimal value of the fixed HP, and use it with the optimal value of the fixed HP. The transfer error is the difference between the losses obtained and the minimum loss. \Cref{fig:experiments:transfer_error} shows this measure for each pair of HPs under \mup\ and \umup, reflecting the improvement in HP dependency as a result of our scheme. This gives \umup\ a reduced risk of small transfer errors leading to large degradations, and the potential to sweep HPs in a more separable way.

\subsection{Hyperparameter search} We now leverage this improved separability of HPs for the purpose of efficient sweeping. In \Cref{fig:fig1}~(a) we conduct a standard random search for \mup\ and \umup, along with the independent search outlined in \Cref{sec:umup_hp_search} (and \Cref{app:umup_hp_search_algorithm}). We observe the following:

\begin{enumerate}
    \item For \umup\ the LR-sweep phase of independent search alone is sufficient to reach near-optimal loss (totaling 9 runs). During this phase other HPs are fixed at 1, which for \umup\ means that the inputs to operations are generally unit-scaled.

    \item Consequently, we conclude that unit scale at initialization is close to the ideal scaling for effective learning here. This is not a property we asserted a priori, nor do we argue that it necessarily holds for other training setups and models; hence why we still provide a set of extended HPs to be swept.

    \item In contrast \mup\ still requires non-LR HPs to be swept to attain a reasonable loss. Unlike \umup, fixing HPs at 1 results in arbitrarily-scaled inputs, which appear to result in worse training.

    \item The `combined mults' phase causes the loss to spike for \mup. This is due to the HP dependencies shown in \Cref{fig:experiments:transfer_error}, which mean HPs cannot be swept independently and used together. Conversely, lower dependence means this can be done for \umup, making random search unnecessary.
\end{enumerate}

\subsection{Hyperparameter transfer}

\begin{figure}[t]
    \centering
    \begin{subfigure}{\textwidth}
        \centering
        \includegraphics[width=\textwidth]{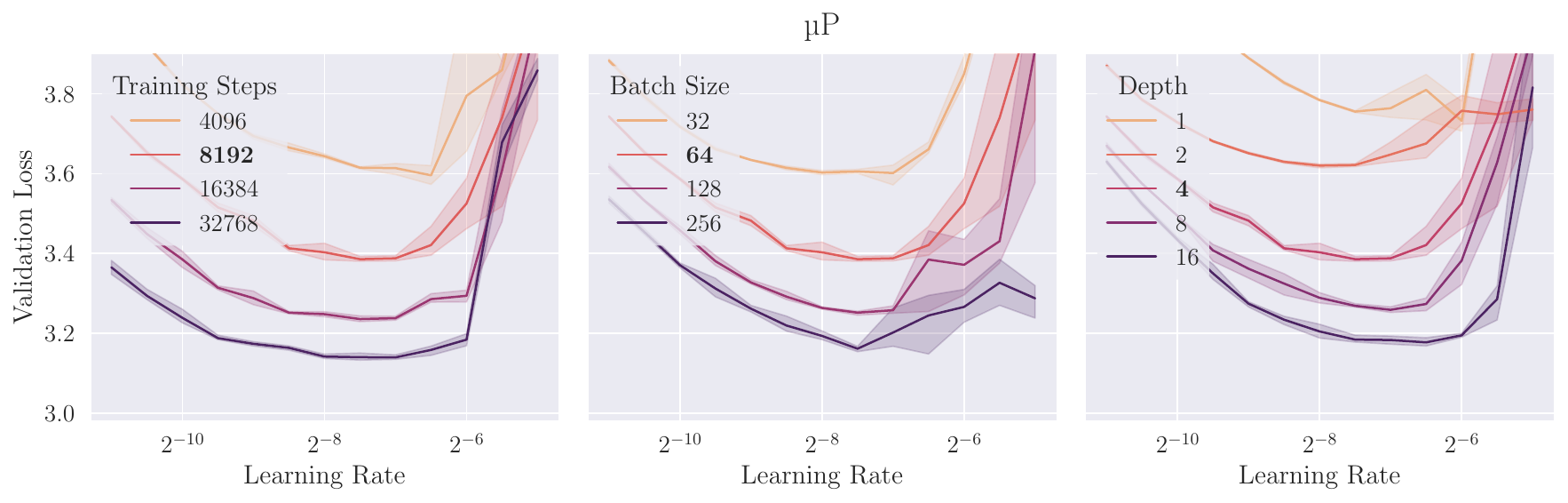}
    \end{subfigure}
    \begin{subfigure}{\textwidth}
        \centering
        \includegraphics[width=\textwidth]{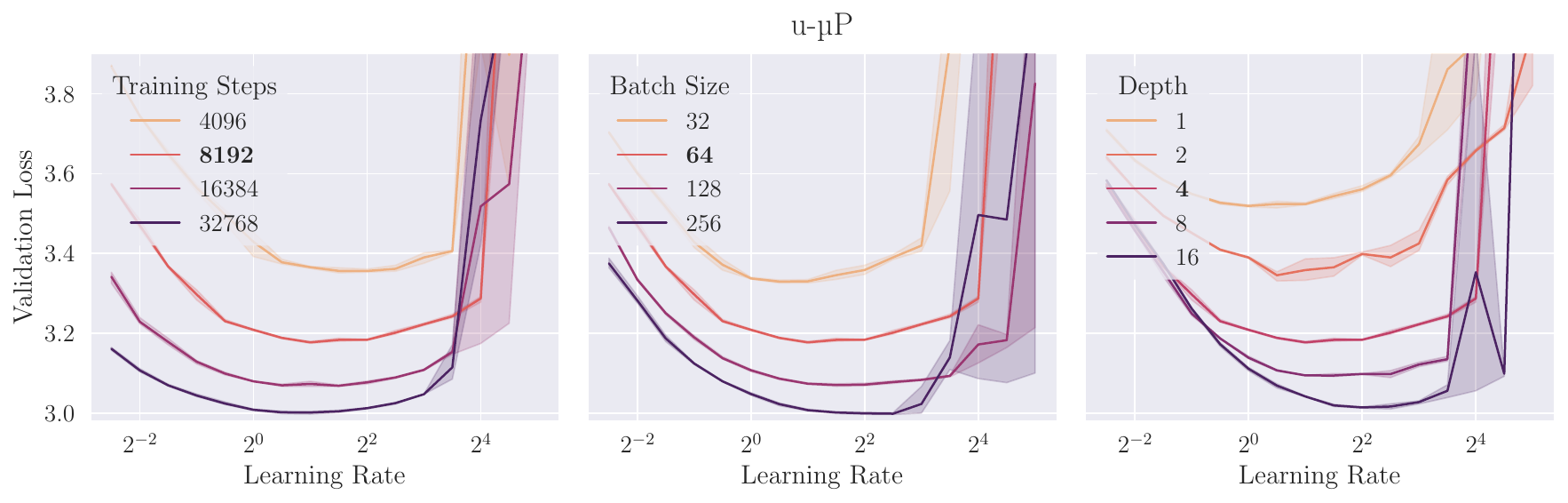}
    \end{subfigure}
    \caption{Learning rate transfer for \mup{} (top) and \umup{} (bottom), over training steps, batch size and depth. See \Cref{fig:fig1}~(b) for transfer over width. The \textbf{default} shape parameter for other panels is shown in bold. The shaded area shows the $95\%$ confidence interval for the mean.}
    \label{fig:lr_transfer}
\end{figure}

We train many models and plot transfer of LR across width (\Cref{fig:fig1}~(b)), steps, batch size and depth (\Cref{fig:lr_transfer}), and transfer of other HPs across width (\Cref{fig:experiments:hp_transfer_over_width}). Note that \umup\ (building on \mup) is designed to give transfer over width\footnote{As we use \depthmup\ this could be said about depth as well, but as \citep{Tensor_Programs_VI} show that transformers don't attain depth-transfer under \depthmup\ we do not expect strong transfer across depth.}; the other axes we report for practical purposes. We find that:

\begin{enumerate}
    \item The optimal LR is constant across width under \umup. There is a small drift for training steps and batch size, and a larger one with depth. Hence we recommend proxy models which primarily differ in width, moderately in steps and batch size, and least in depth.

    \item The optimal LR is also approximately constant for training steps, batch size and depth. This means we can scale our proxy model down across all these axes and maintain LR transfer. Of these, width appears the most stable and depth the least.

    \item Whereas \mup\ sees diminishing returns for larger widths, \umup\ continues to benefit from width, with the 2048 \umup\ model matching the 4096 \mup\ model. We attribute this primarily to our improved embedding LR rule.

    \item Non-LR HPs also have approximately constant optima across width under \umup. This is not true for \mup, where $\hat{\eta}_\mathrm{emb}$ has poor transfer due to the embedding scaling rule issue identified in \Cref{sec:umup:emb_lr_rule}, along with $\sigma_\mathrm{init}$ which in \Cref{sec:challenges:which_hps} we argue should not be grouped across all weights (and drop from the \umup\ HP scheme).

    \item The optimal values found for non-LR HPs are all close to 1. In practice this means that dropping these HPs entirely is potentially viable for similar models and training setups.
\end{enumerate}

\subsection{FP8 training} \label{sec:fp8_training}

In this section we justify the simple mixed-precision scheme described in \Cref{sec:umup:low_prec_training} and demonstrate that it can be used to train \umup\ models out-of-the-box.

\begin{figure}[t]
    \centering
    \begin{subfigure}{.46\textwidth}
        \centering
        \includegraphics[width=\textwidth]{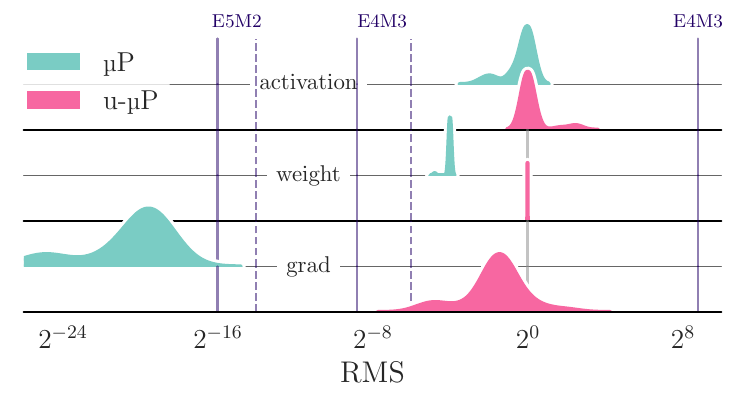}
    \end{subfigure}
    \begin{subfigure}{.46\textwidth}
        \centering
        \includegraphics[width=\textwidth]{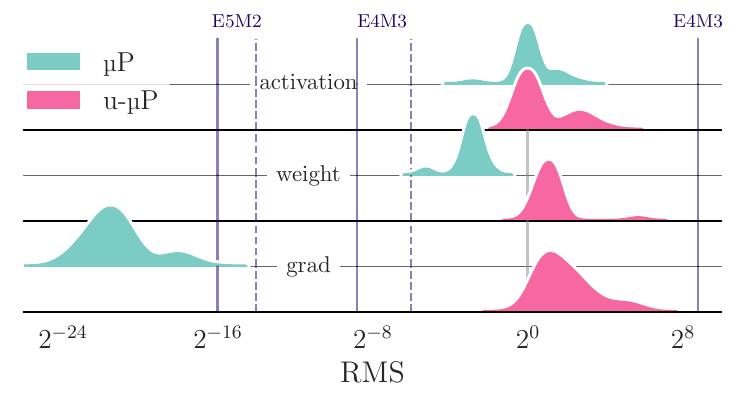}
    \end{subfigure}
    \caption{Per-tensor $\mathrm{RMS} = \sqrt{\sigma^2 + \mu^2}$ across \umup\ and \mup\ models at initialization (left) and after training (right). \umup\ tensors have RMS that starts close to $1$ and remains within E4M3 range at the end of training. Dashed and solid red lines show each format's min. normal and subnormal values.}
    \label{fig:numerics:scale}
\end{figure}

\paragraph{Proof-of-concept} \Cref{fig:numerics:scale} shows the RMS of all linear layer inputs for a moderately sized transformer. RMS captures the larger of the mean and scale of a distribution, and as such is a good test of whether a tensor is likely to suffer over/underflow in low-precision. We observe that \umup\ tensors largely have RMS starting close to $1$ and remaining so at the end of training, supporting our scheme.

\Cref{fig:numerics:rms_during_training} demonstrates the scale-growth of critical tensors which our scheme is designed to accommodate, showing RMS on a per-tensor basis over steps. \Cref{fig:numerics:scale_scaling} provides further insight into this issue, showing the effect of LR, width, depth, steps and batch size on the RMS of critical tensors.

As an initial proof-of-concept we train a \umup\ model using our FP8 scheme over 8k steps, using HPs from a proxy model, as shown in \Cref{fig:fig1}~(c). We see only a small degradation versus FP32, and at this scale critical tensors can still be cast to FP8 using \texttt{E5M2}, while gradients can even use \texttt{E4M3}.

\paragraph{Larger scale} Next we consider a more realistic training scenario \footnote{The training codebase used for our larger-scale experiments can be found at the following url \url{https://github.com/Aleph-Alpha/scaling}. We have also released model checkpoints, which are available at \url{https://huggingface.co/Aleph-Alpha}.}.
Using the same architecture, and following the steps set out in our \umup\ user-guide (\Cref{app:using_umup_guide}), we train our target models on 300B tokens of the SlimPajama dataset \citep{SlimPajama} (see \Cref{app:large_model_training} for training details).

We begin with an independent search (\Cref{sec:umup_hp_search}) over our \umup\ proxy model's HPs. Here we make the following observations:
\begin{enumerate}
    \item When using a relatively small proxy model (8 layers and 512 width), the HP-loss landscape is rather noisy. By doubling the width we can discern optimal HP values more clearly.
    \item The most important HPs are $\eta$ and $\alpha_\mathrm{res\text{-}attn\text{-}ratio}$. All others can be left at the default of $1$.
    \item The optimal values of these HPs are $\eta = 2^{3.5}$ and $\alpha_\mathrm{res\text{-}attn\text{-}ratio} = 2^{-2.0}$ and thus differ non-trivially from the observed HPs in our smaller-scale experiments.
\end{enumerate}

\begin{figure}[t]
    \vspace{-1.5em}
    \centering
    \begin{subfigure}{0.4\textwidth}
        \centering
        \includegraphics[width=\textwidth]{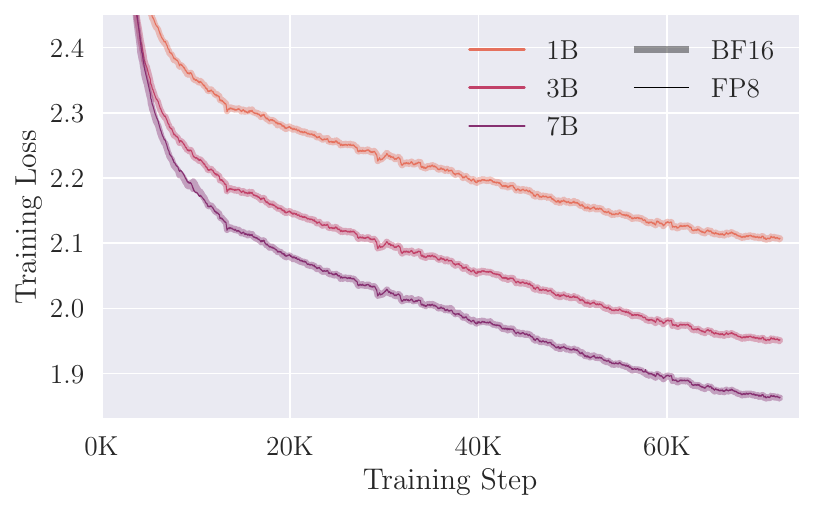}
    \end{subfigure}
    \hspace{2em}
    \begin{subfigure}{0.4\textwidth}
        \centering
        \includegraphics[width=\textwidth]{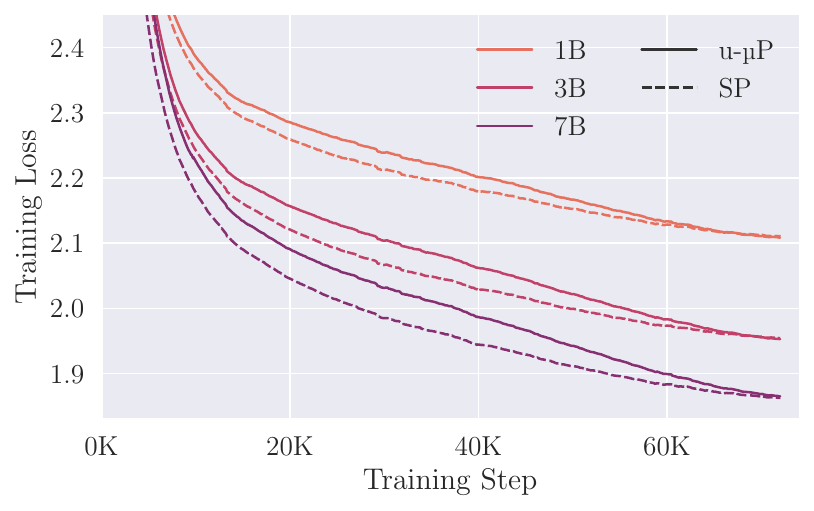}
    \end{subfigure}
    \vspace{-0.5em}
    \caption{Large-scale training runs. (Left) \umup\ BF16 vs \umup\ FP8. (Right) \umup\ BF16 vs SP BF16.}
    \label{fig:scaleup}
\end{figure}

We then train \umup\ models of approximately 1B, 3B and 7B parameters, using our FP8 mixed-precision scheme (see \Cref{sec:umup:low_prec_training}). We also train two baselines at each size: the first is a BF16 version of our \umup\ models, and the second is a set of SP models using the weight init scheme from the Pythia model family~\citep{Pythia} and the LR scheme from Llama 3~\citep{LLAMA3}, scaling inversely with width and using a LR of 3e-4 at 7B scale.
The loss curves are shown in \Cref{fig:scaleup}. All FP8 runs converge and show no significant loss degradation. In comparison to SP, the \umup\ models have a qualitatively different training curve with a higher loss for most of training that catches up in latter stages, hinting at a fundamentally different optimization trajectory. In terms of downstream performance, both of the \umup\ 7B models are competitive with SP. In particular, the scores of the FP8 model are mostly on par with the BF16 models (see \Cref{tab:eval_results}).

\begin{table}[t] 
  \centering
  \caption{0-shot benchmark results at 7B scale.}
\begin{tabular}{llcccccc}
\toprule
Scheme & Format & MMLU & HellaSwag & OpenBook QA & PIQA & TriviaQA & WinoGr \\
\midrule
SP & BF16 & 29.6 & 52.4 & 27.8 & 76.5 & 22.2 & 63.3 \\
\umup\ & BF16 & 29.0 & \textbf{53.4} & \textbf{31.6} & 77.1 & \textbf{23.4} & 63.7 \\
\umup\ & FP8 & \textbf{31.2} & \textbf{53.4} & 29.6 & \textbf{77.6} & 21.3 & \textbf{65.7} \\
\bottomrule
\label{tab:eval_results}
\vspace{-1.5em}
\end{tabular}
\end{table}




\section{Related Work}

\paragraph{Low-precision training}

Techniques introduced to facilitate FP8 training include those covered in \Cref{app:low_precision_and_its_trade_offs} and more \citep{Training_And_Inference_Using_8_Bit,DNNs_With_8_Bit,Mixed_Precision_8_Bit}. These largely concern the quantizing of activations, weights and gradients, though \citep{FP8-LM} also explore FP8 optimizer states and cross-device communication, which we consider interesting avenues of further exploration. Recently, stable training has been demonstrated for the MX family of formats which use a shared block-exponent \citep{MX1,MX2}, and even for the ternary BitNet format \citep{BitNet1,BitNet2,BitNet3}. Again, we consider these formats for follow-up work.

\paragraph{Stability features}

Another recent research trend has been the analysis of features that contribute to (or resolve) numerical and algorithmic instability. \citep{Small_Scale_Proxies} show that unstable training dynamics can result from attention logit growth (fixed by QK-norm \citep{Scaling_Vision_Transformers}) and from divergent output logits (fixed by z-loss \citep{Palm}). \citep{Stable_And_Low_Precision_Training} find large feature magnitudes can be avoided by zero-initialization, and loss spikes avoided via a modified AdamW, specifically for low-precision training. \citep{Intriguing_Properties} investigate how pre-training settings affect instabilities revealed during post-training quantization. \citep{Dynamics_Of_Diffusion_Models} apply a similar philosophy to Unit Scaling for the training of diffusion models, to address uncontrolled magnitude changes. Extreme activation values seen in large models \citep{LLM_INT8,SmoothQuant} have been addressed by softmax-clipping \citep{Quantizable_Transformers}, and by the addition of extra terms \citep{Massive_Activations} or tokens \citep{Vision_Transformers_Need_Registers} to bias the attention computation. We do not adopt these features in our experiments to avoid confounding effects, but we expect them to benefit \umup\ and hope to explore their usage.

\paragraph{Learning dynamics}

Several recent efforts have tried to improve \mup\ from different angles.
\citep{Modula} introduces the notion of the \textit{modular norm} over the full weight-space, which like \mup\ aims to ensure stable updates that provide LR transfer, and like \umup\ is implemented via modules designed to ensure stable training.
Challenging the assumptions underpinning \mup, \citep{Scaling_Exponents} explores the notion of \textit{alignment} between parameters and data, demonstrating that other parametrizations with per-layer learning rates can outperform standard \mup. We consider comparing these parametrizations against \umup\ and trying unit-scaled versions valuable future work. Recent applications of \mup\ to the problems of weight sparsity \citep{Sparse_Mup} and structured matrices \citep{Compute_Better_Spent} are also interesting candidates for \umup.

\section{Conclusions}

We introduce \umup, a modified and improved version of \mup\ that satisfies Unit Scaling. Through careful analysis guided by first principles we identify an interpretable set of HPs that has minimal interdependencies and facilitates an efficient independent sweeping strategy. We show that the stability properties of \mup\ combined with Unit Scaling enable a simple and robust FP8 mixed precision scheme that works in a realistic large scale training scenario.
\umup\ provides further evidence that the principle of Unit Scaling is beneficial for model design.

\paragraph{Limitations and future work} Some choices like the modified embedding LR rule are only justified by empirical observations, and currently lack a theoretical explanation. Additionally, neither \mup\ nor Unit Scaling give guarantees for network quantities to be well-behaved over the course of training. In particular we would like to understand the issue (or feature) of scale growth in the critical layers better and look into possible mitigations. We also believe that low-precision training techniques can be pushed further, with \umup\ offering an ideal starting point for future optimizations.

\section{Acknowledgments}

We would like to thank Paul Balança, Andrew Fitzgibbon, Steve Barlow, Mark Pupilli, Jeremy Bernstein, Tim Large and Lucas Lingle for their feedback and discussions around \umup\ at the various stages of its development.

\clearpage
\bibliographystyle{unsrt}
\bibliography{main_arxiv}
\clearpage
\tableofcontents
\clearpage
\newpage
\appendix

\section{Additional experimental details} \label{app:additional_experimental_results}

\subsection{Experimental Setup} \label{app:additional_experimental_results:experimental_details}

Our experimental analysis of \umup\ was conducted by adapting the codebase used for Tensor Programs V, allowing us to compare \mup\ and \umup\ in the same setting. We change various experimental settings from the original paper to make our experiments better reflect standard training procedures, particularly the dataset which we switch from WikiText-2 to the larger WikiText-103 \citep{WikiText103}. Where not specified otherwise, the default setting used in our experiments are given in \Cref{tab:experiment_defaults}. These also represent the settings of our proxy model.

\begin{table}[h] 
    \centering
    \renewcommand{\arraystretch}{1.25}
    \begin{tabular}{|r|p{10cm}|}
    \hline
        Dataset & WikiText-103 \citep{WikiText103} \\
        Sequence length & $256$ \\
        Vocab size & $32000$ \\
        Training set tokens & $138\mathrm{M}$ \\
        \hline
        Architecture & Llama \citep{Llama} \; (Transformer, PreNorm, RMSNorm, SwiGLU, RoPE, ``untied'' embeddings), non-trainable RMSNorm parameters. \\
        Width & $256$ \; (scaled up to $4096$) \\
        Depth & $4$ \\
        Number of heads & $4$ \; (scaled up to $64$) \\
        Head dimension & $64$ \\
        Total parameters & $19.5M$ \; (scaled up to $1.07\mathrm{B}$)\\
        \hline
        Batch size & $64$ \\
        Training steps & $8192$ ($0.97$ epochs) \\
        LR schedule & Cosine to $10\%$, $2000$ steps warm-up \\
        Optimizer & AdamW $(\beta_1, \beta_2, \epsilon) = (0.9, 0.999, 10^{-8})$ \\
        Weight decay & $2^{-13}$, independent \citep{Independent_WD} \\
        Dropout & $0.0$ \\
        \hline
         \mup{} HP search range &
        $\eta \in [2^{-10}, 2^{-6}]$\\
        &$\hat{\eta}_{\mathrm{emb}} \in [2^{0}, 2^{8}]$\\
        &$\sigma_{\mathrm{init}},
        \alpha_{\mathrm{emb}},
        \alpha_\mathrm{{attn}},
        \alpha_{\mathrm{output}} \in [2^{-2}, 2^{2}]$\\
        \umup{} HP search range &
        $\eta \in [2^{-1}, 2^{3}]$\\
        &$\alpha_{\mathrm{attn}} \in [2^{-2}, 2^{2}]$\\
        &$\alpha_{\mathrm{residual}},\alpha_{\mathrm{residual{\text -}attn{\text -}ratio}},\alpha_{\mathrm{ffn{\text -}act}},\alpha_{\mathrm{output}} \in [2^{-3}, 2^{3}]$\\
         \hline
        \mup{} HP defaults &
        $\sigma_{\mathrm{init}}=
        \alpha_{\mathrm{emb}}=
        \alpha_\mathrm{{attn}}=
        \alpha_{\mathrm{output}}=
        \hat{\eta}_{\mathrm{emb}}=1$\\
        \umup{} HP defaults &
        $\alpha_{\mathrm{residual}}=
        \alpha_{\mathrm{residual{\text -}attn{\text -}ratio}}=
        \alpha_{\mathrm{ffn{\text -}act}}=
        \alpha_{\mathrm{output}}=
        \alpha_{\mathrm{attn}}=1$\\
    \hline
    \end{tabular}
    \caption{Default hyperparameters and training settings.}
    \label{tab:experiment_defaults}
\end{table}

\FloatBarrier
\clearpage

\subsection{Further analysis of \mut\ failure modes}

In \Cref{table:experiments:tpv_vs_llama_settings} we provide exact details of the experimental differences between setups (a) and (b) from \Cref{fig:experiments:tpv_setup}, for readers wishing to understand and reproduce this result. We also provide a step-by-step ablation of the various changes made between these setups in \Cref{fig:experiments:tpv_setup_standard_ablation}.

\begin{figure}
    \centering
    \begin{subfigure}{\textwidth}
        \centering
        \includegraphics[width=\textwidth]{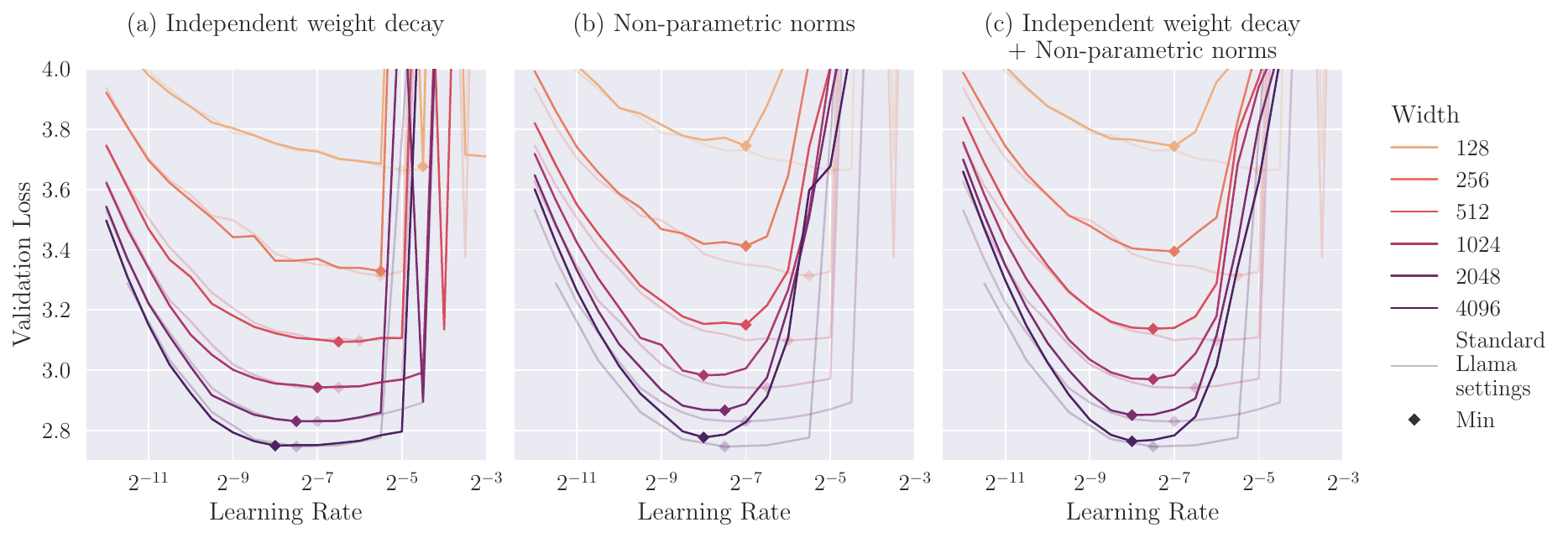}
    \end{subfigure}
    \caption{The effect of the individual transfer stability fixes from \Cref{fig:experiments:tpv_setup}.
    \textbf{(a)} In this setting switching from non-independent to independent weight decay has only a minor effect, though \citep{Small_Scale_Proxies} Figure 6 suggests it may be highly valuable in other settings. \textbf{(b)} Non-parametric norms give a narrower learning rate basin, leading to better transfer. \textbf{(c)} The combination of these, for comparison, matching \Cref{fig:experiments:tpv_setup} (c).}
    \label{fig:experiments:tpv_setup_stability_fixes}
\end{figure}

\begin{figure}
    \centering
    \begin{subfigure}{\textwidth}
        \centering
        \includegraphics[width=\textwidth]{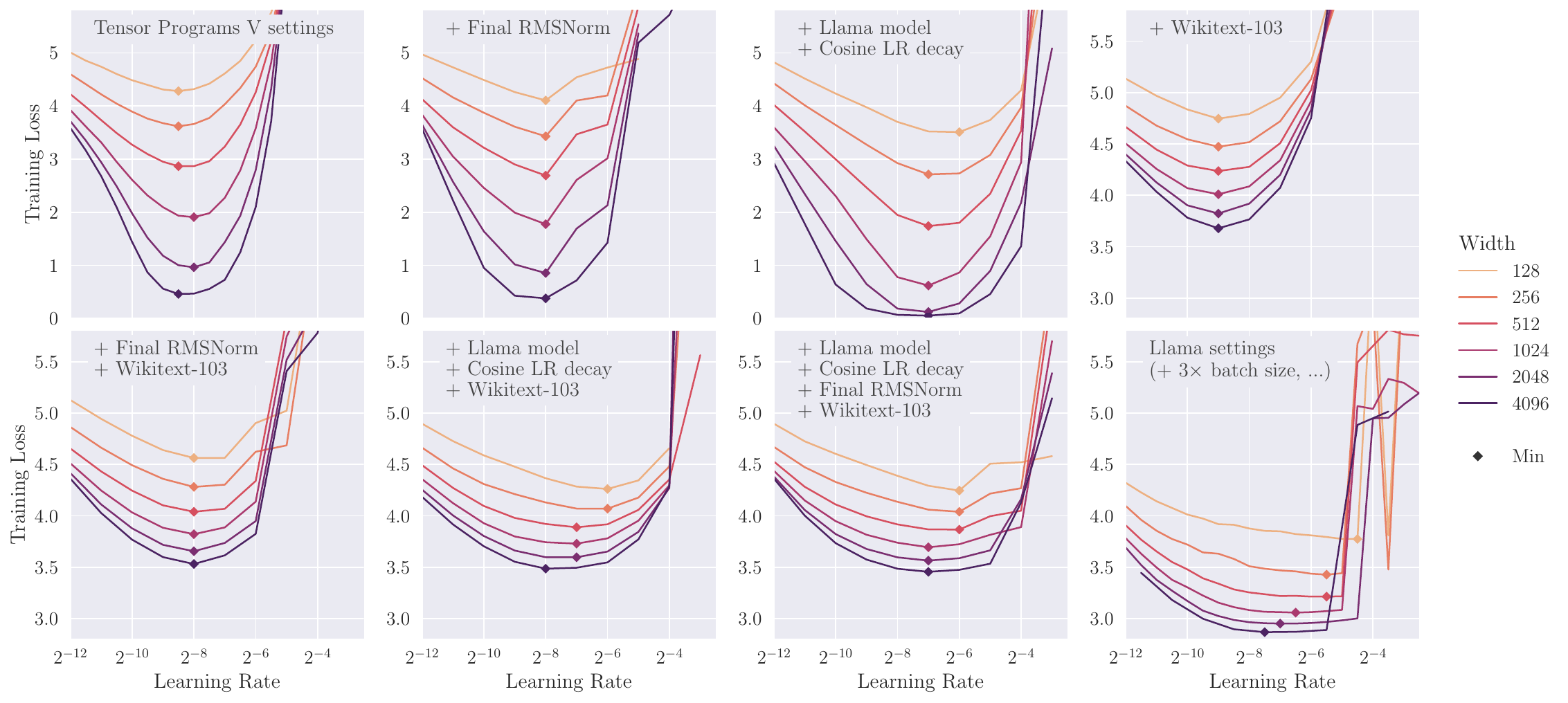}
    \end{subfigure}
    \caption{An ablation of the more standard Llama training settings against the Tensor Programs V settings from \Cref{fig:experiments:tpv_setup}. This shows that the flat basins with poor transfer are not due to a single change, but the combination of a larger dataset (training $<\!1$ epoch) and the stronger Llama model are largely responsible. Note that `Llama model' here indicates a group of changes: rms norm, rotary embeddings \& swiglu FFN.}
    \label{fig:experiments:tpv_setup_standard_ablation}
\end{figure}

For setup (c), which shows how our two combined stability fixes mitigate the problem of poor transfer, both changes are evaluated independently in \Cref{fig:experiments:tpv_setup_stability_fixes}, which shows that the dominant effect is a narrowing of the learning basin due to non-parametric RMSNorms, leading to better learning rate transfer.

\begin{table}[h!]
\caption{Comparison of Tensor Programs V's standard settings (as best we can tell) and our Standard Llama setup, corresponding to (a) and (b) in \Cref{fig:experiments:tpv_setup}.}
\vspace{1em}
\centering
\begin{tabular}{lcc}
\toprule
\textbf{Feature} & \textbf{Tensor Programs V} & \textbf{Standard Llama} \\
\midrule
Dataset & wikitext-2 & wikitext-103 \\
Vocab Size & 33278 & 32000 \\
Nsteps & 10000 & 8192 \\
Batch Size & 20 & 64 \\
Optimizer & adam & adamw \\
LR Schedule & constant & cosine \\
Weight Decay & 0 & 0.00012 \\
Positional Encoding & absolute & rotary \\
Norm & layer\_norm & rms\_norm \\
Dropout & 0.2 & 0 \\
NLayers & 2 & 4 \\
Use Gated FFN & False & True \\
Activation FN & relu & swish \\
FFN Ratio & 4 & 2.75 \\
Final Norm & False & True \\
Base Depth & 1 & 4 \\
Zero Init Readout & True & False \\
\bottomrule
\end{tabular}
\label{table:experiments:tpv_vs_llama_settings}
\end{table}

\subsection{Validating our experimental setup}\label{app:valid_experiments}

In this section we run a series of ablations to validate decisions relating to our experimental setup given above. In particular, we examine the effect of using repeated data, the effect of using a shorter warmup duration, and the effect of different final learning rates at the end of decay.

\subsubsection{Repeated data}

As outlined in \Cref{tab:experiment_defaults}, our standard training setup uses 0.97 epochs of the WikiText-103 dataset (50x larger than the WikiText-2 dataset used in Tensor Programs V). However on our batch size and training steps scaling experiments in \Cref{fig:lr_transfer} we train on up to $4 \times$ the amount of data than in our standard setup, and hence use up to $4$ epochs.

Though this is still a small level of repeated data, this moves our training slightly into the over-fitting regime. Based on this change, we here investigate the hypothesis that this regime has better or worse transfer of the optimal LR than the non-overfitting regime, and hence our results could be misleading. To do so, we repeated these experiments with the same number of tokens, but using the much larger SlimPajama dataset \citep{SlimPajama} where we use $< 1$ epoch.

\begin{figure}[h]
    \centering
    \includegraphics[width=\textwidth]{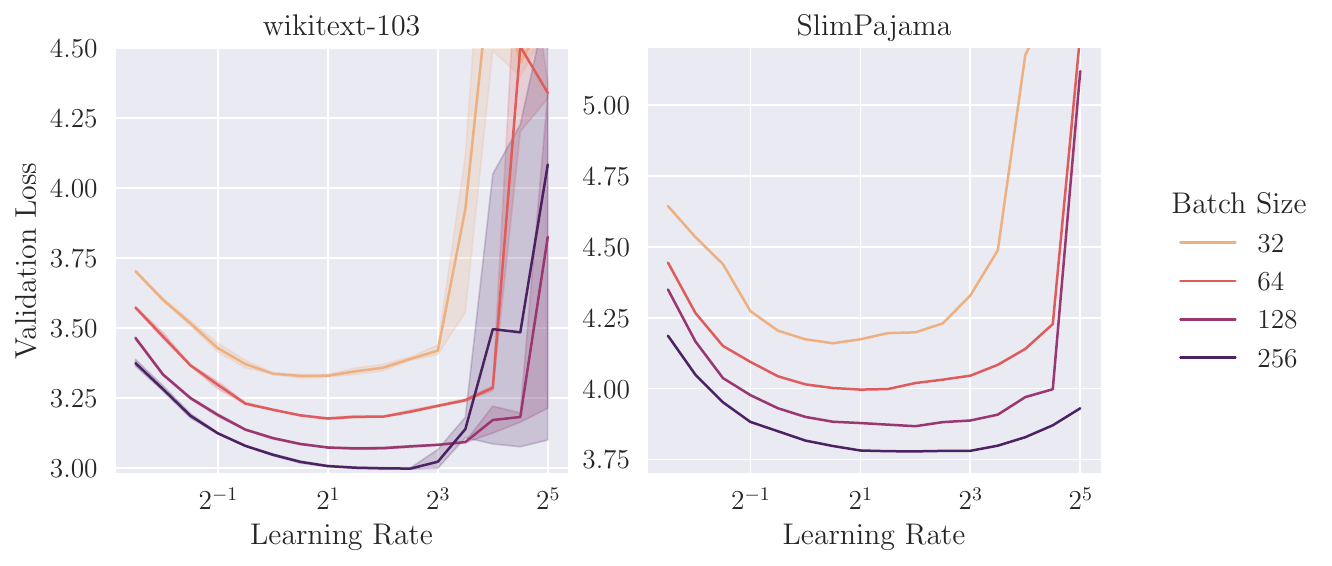}
    \includegraphics[width=\textwidth]{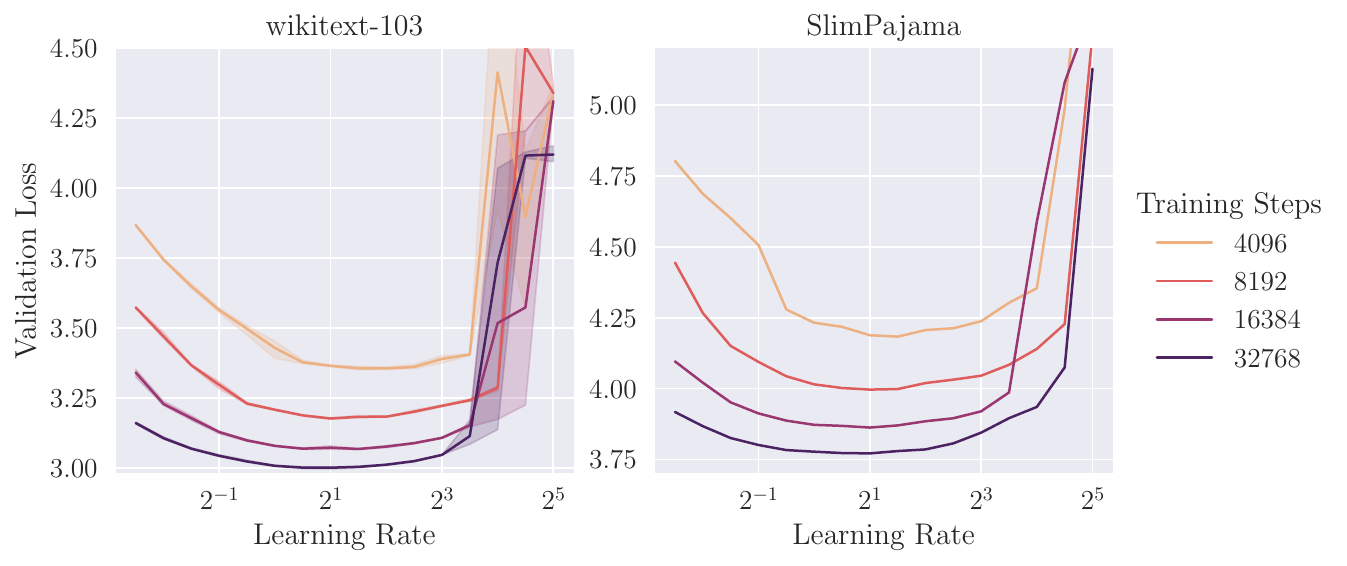}
    \caption{A repeat of the batch size and training steps experiments in \Cref{fig:lr_transfer}, but using the larger SlimPajama dataset where no data is repeated. In both settings our validation loss basins take the same shape, indicating that our analysis using the WikiText-103 dataset holds.}
    \label{fig:additional_experiments:sub_epoch}
\end{figure}

The results for this experiment are seen in \Cref{fig:additional_experiments:sub_epoch}. The shape of curves is very similar across the two datasets, for both batch size and training steps (albeit with a higher loss, due to the more varied nature of SlimPajama). From this we conclude that the effect of repeated data from our use of WikiText-103 is not significant.

\subsubsection{Warmup duration}

For our experimental setup (\Cref{tab:experiment_defaults}) we use a longer duration of warmup than in our large-scale setup (\Cref{tab:large_scale_experiment_defaults}). 
We do so out of caution, as we use fewer tokens-per-batch for the smaller-scale experiments and so may require longer warmup. However, doing so also creates the risk of spending too large a proportion of training doing warmup, which could affect transfer.

\begin{figure}[h]
    \centering
    \includegraphics[width=0.48\textwidth]{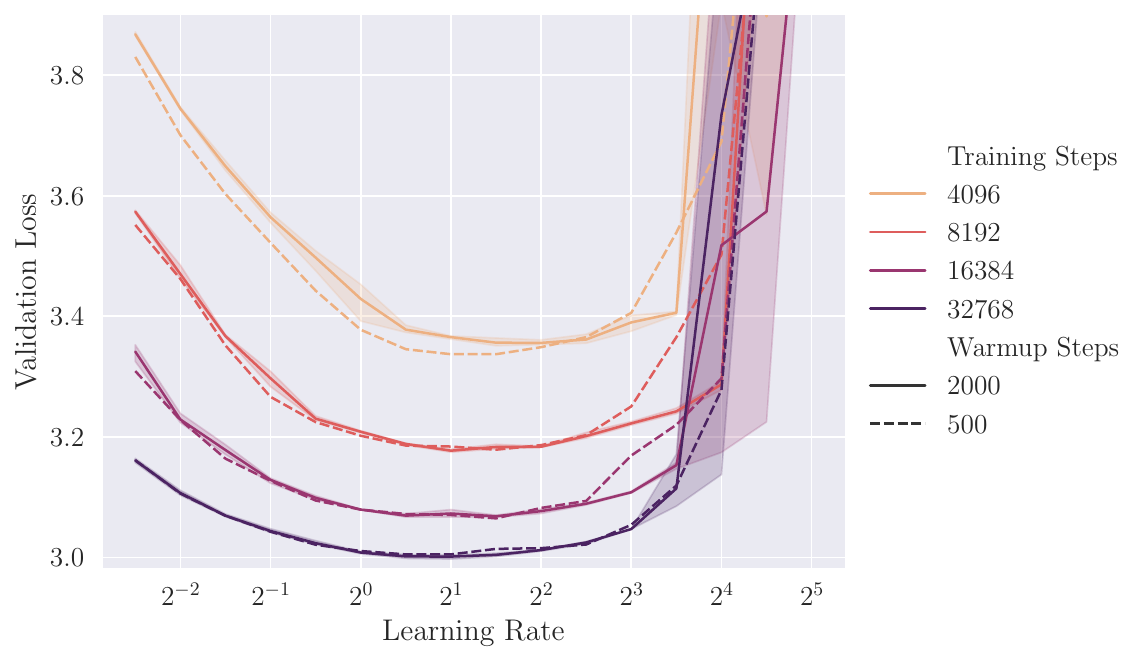}
    \includegraphics[width=0.48\textwidth]{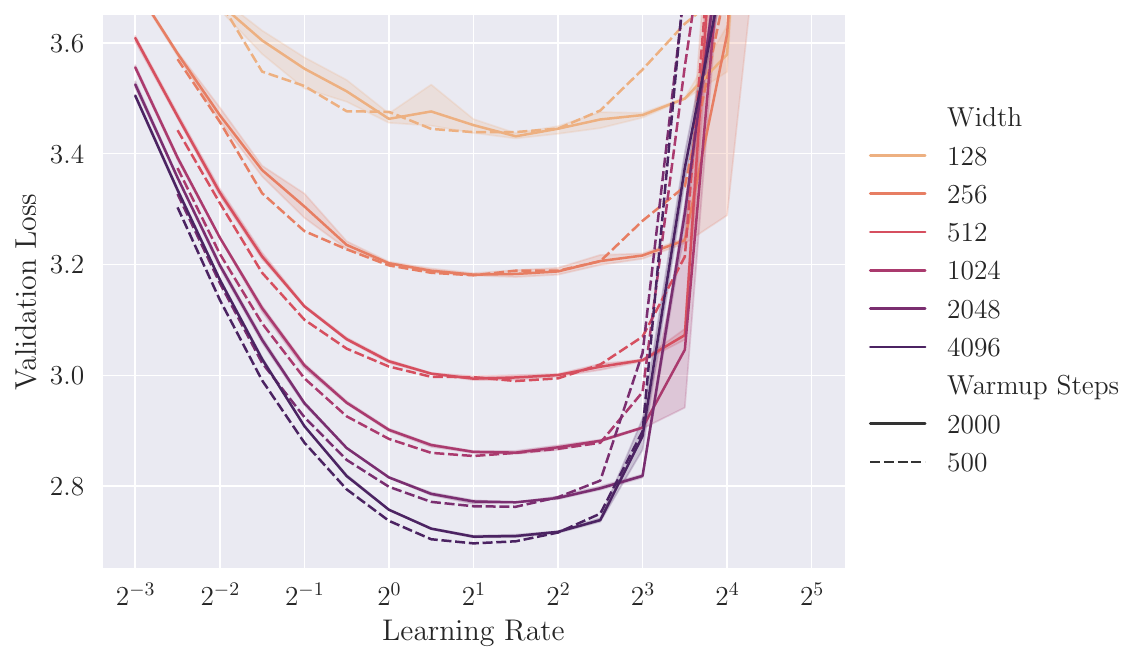}
    \caption{(Left) Learning rate transfer across training steps under different numbers of warmup steps. (Right) Learning rate transfer across width under different numbers of warmup steps. In this setting (training steps = 8192) the optimal LR is consistent, meaning either warmup regime can be used, though the longer gives wider basins.}
    \label{fig:additional_experiments:warmup__transfer_lr_nsteps_width}
\end{figure}

To investigate this effect, we run two experiments. Firstly, we re-run the experiment for LR transfer over training steps, shown in \Cref{fig:lr_transfer}, on a quarter of the warmup steps. This is shown in \Cref{fig:additional_experiments:warmup__transfer_lr_nsteps_width} (left). The main effect appears to be higher loss for larger learning rates, but the optima are largely unchanged. The only exception is the 4096-step run, where the optimum shifts left and the loss improves slightly. This appears to now align the optimum better with the other training durations, but leads to narrower basins as a result, suggesting a trade-off for this particular experiment.

However, all our other experimental runs use the 8192-step configuration, which has a consistent optimum regardless of warmup duration here. To investigate the effect of reduced warmup on width transfer at this particular step-count, we re-run our experiment in \Cref{fig:fig1} (b) under the shorter warmup duration, shown in \Cref{fig:additional_experiments:warmup__transfer_lr_nsteps_width} (right). The only significant impact of this change is to narrow the basins, inducing no significant change in the optimal LR. As such, we conclude that using 2000 steps of warmup in our experimental setup is a reasonable choice, and both give the same width transfer.

\subsubsection{Learning rate decay target}

In all our experiments we use a cosine decay of our learning rate down to 10\% of the maximum. This follows the standard approach taken by most LLM training projects \citep{Llama_2, Pythia, Olmo, Falcon, Qwen2}. However, recent research has indicated that this may not be the optimal decay target, with implications for LR transfer. \citep{BeyondFixedTraining} show that the choice of target percentage can alter the shape of transfer curves and potentially shift the optimum value (Figure 21, right). They also suggest that using a fixed target value may work better than a percentage (Figure 22, right), which could be swept separately. \citep{StraightToZero} separately suggest that linear decay to zero is the most effective scheme.

\begin{figure}[h]
    \centering
    \includegraphics[width=0.48\textwidth]{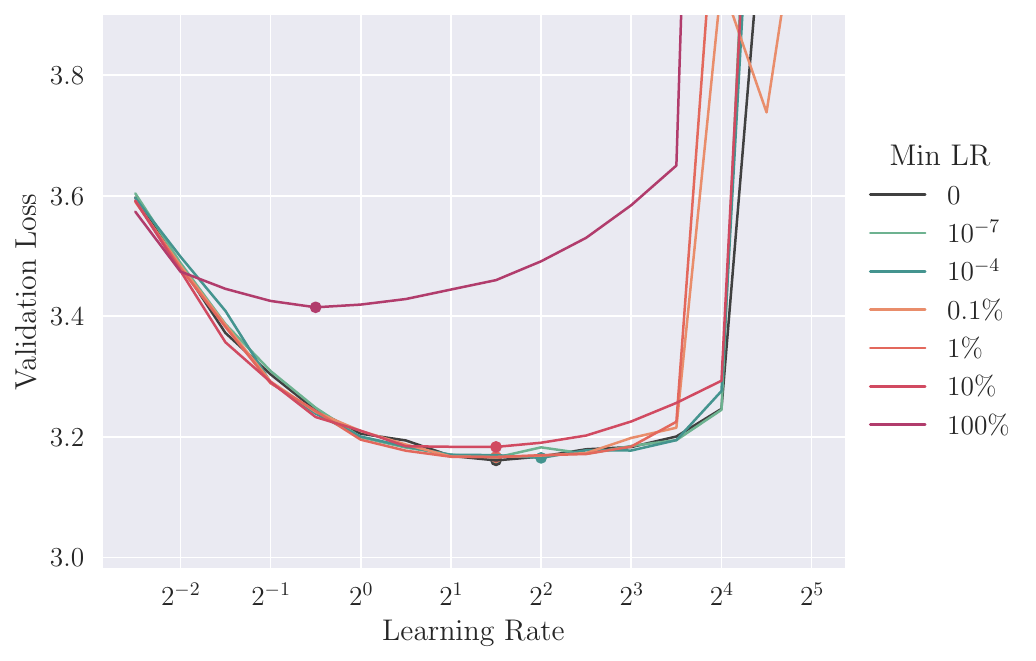}
    \includegraphics[width=0.48\textwidth]{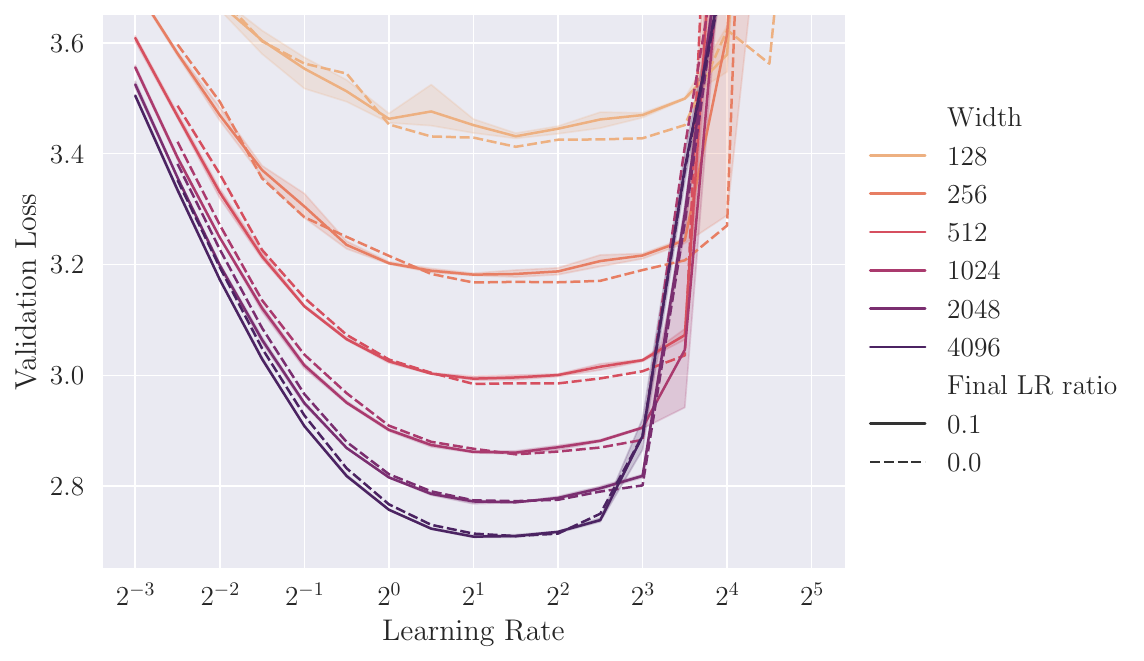}
    \caption{(Left) A learning rate sweep over LR targets of different types (percentage, fixed and zero) on our standard model. (Right) Using the zero and 10\% learning rate targets, LR transfer over width.}
    \label{fig:additional_experiments:lrdecay__sweep_lr}
\end{figure}

Though using the optimal decay scheme is not necessarily essential to the validity of our method, any implications of different schemes on transfer properties should be investigated. To do so, we run two experiments. The first sweeps the learning rate for our standard model at various percentages and fixed values of cosine decay target, including zero, in \Cref{fig:additional_experiments:lrdecay__sweep_lr} (left). Lower decay targets perform better here, including zero, suggesting that this simple rule may be ideal.

We then re-run our width transfer experiment from \Cref{fig:fig1} (b) but with our LR decaying to 0, and plot the result in \Cref{fig:additional_experiments:lrdecay__sweep_lr} (right). This leads to slightly better results for large learning rates, though for large models this difference diminishes. Fortunately the effect this decay target has on the shape of curves (and hence optimal LR transfer) is minimal, indicating that our conclusions are not effected significantly by the choice of decay target.

\FloatBarrier
\clearpage

\subsection{Per-tensor learning rates}\label{app:umup_lr_mults}

\begin{figure}[h]
    \centering
    \includegraphics[width=\textwidth]{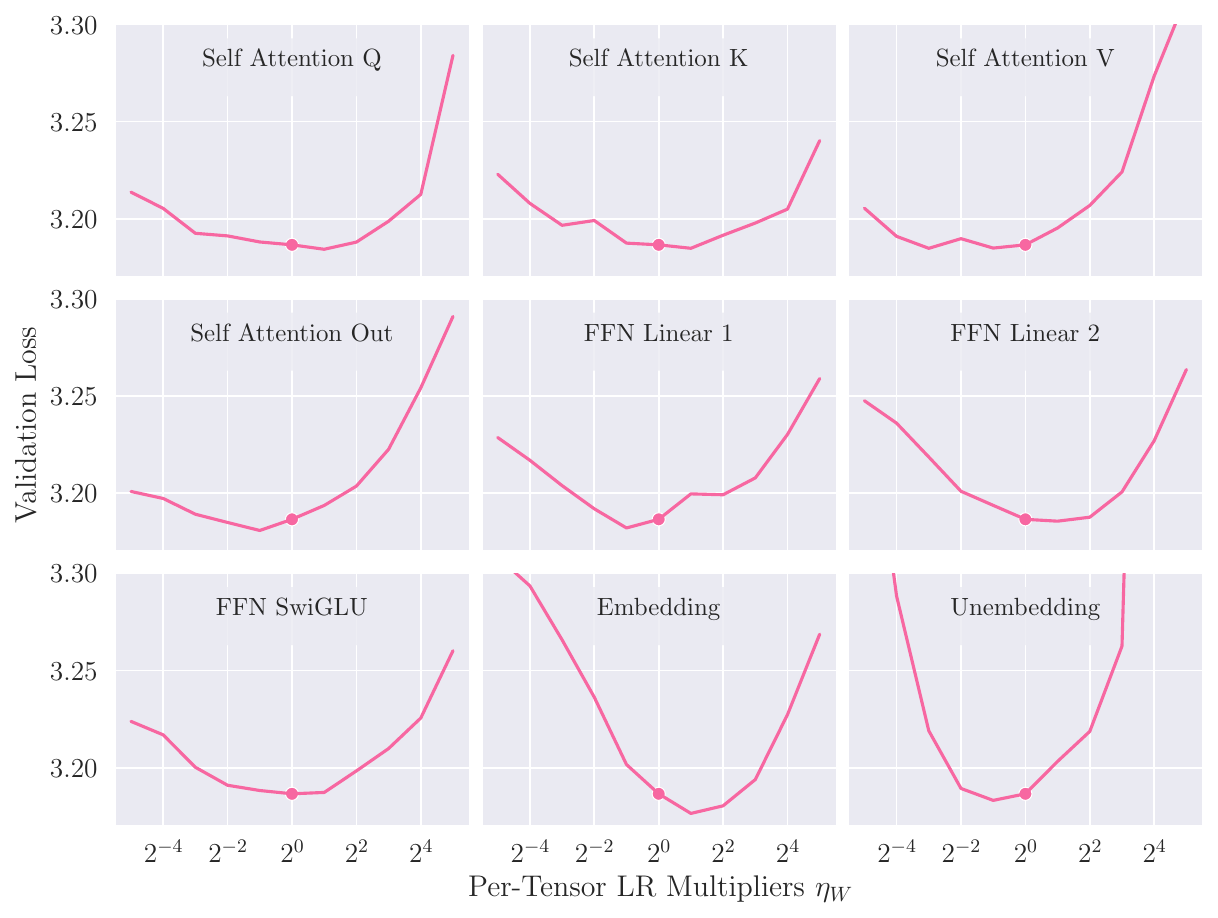}
    \caption{Independently varying per-tensor learning rate multipliers $\eta_W$, using the \umup\ model of width 256 from \Cref{fig:fig1} with optimized global learning rate $2^{1.5}$ as the starting point. Where applicable, the same multiplier is used for tensors of the same name across transformer layers. Each subplot fixes all but one multiplier at 1, therefore the midpoint of each subplot is precisely the \umup{}$_{256}$ model from \Cref{fig:fig1}.}
    \label{fig:additional_experiments:lr_mults}
\end{figure}

In \Cref{sec:umup:principled_hps} we relax the requirement for each weight tensor in the \umup\ model to have an associated tuneable learning-rate multiplier on top of the global learning rate. Whilst this does reduce the theoretical expressivity of the \umup\ scheme, \Cref{fig:additional_experiments:lr_mults} shows that using a single globally optimized learning rate is already at or close to the optimal choice for all weight tensors, and therefore it is reasonable to drop these multipliers in favor of reducing the number of HPs. However, a practitioner attempting to absolutely maximize the task performance of their model could experiment with tuning a few key per-tensor LRs, in particular the embedding table.

\subsection{Hyperparameter independence}

\begin{figure}[h]
    \centering
    \includegraphics[width=\textwidth]{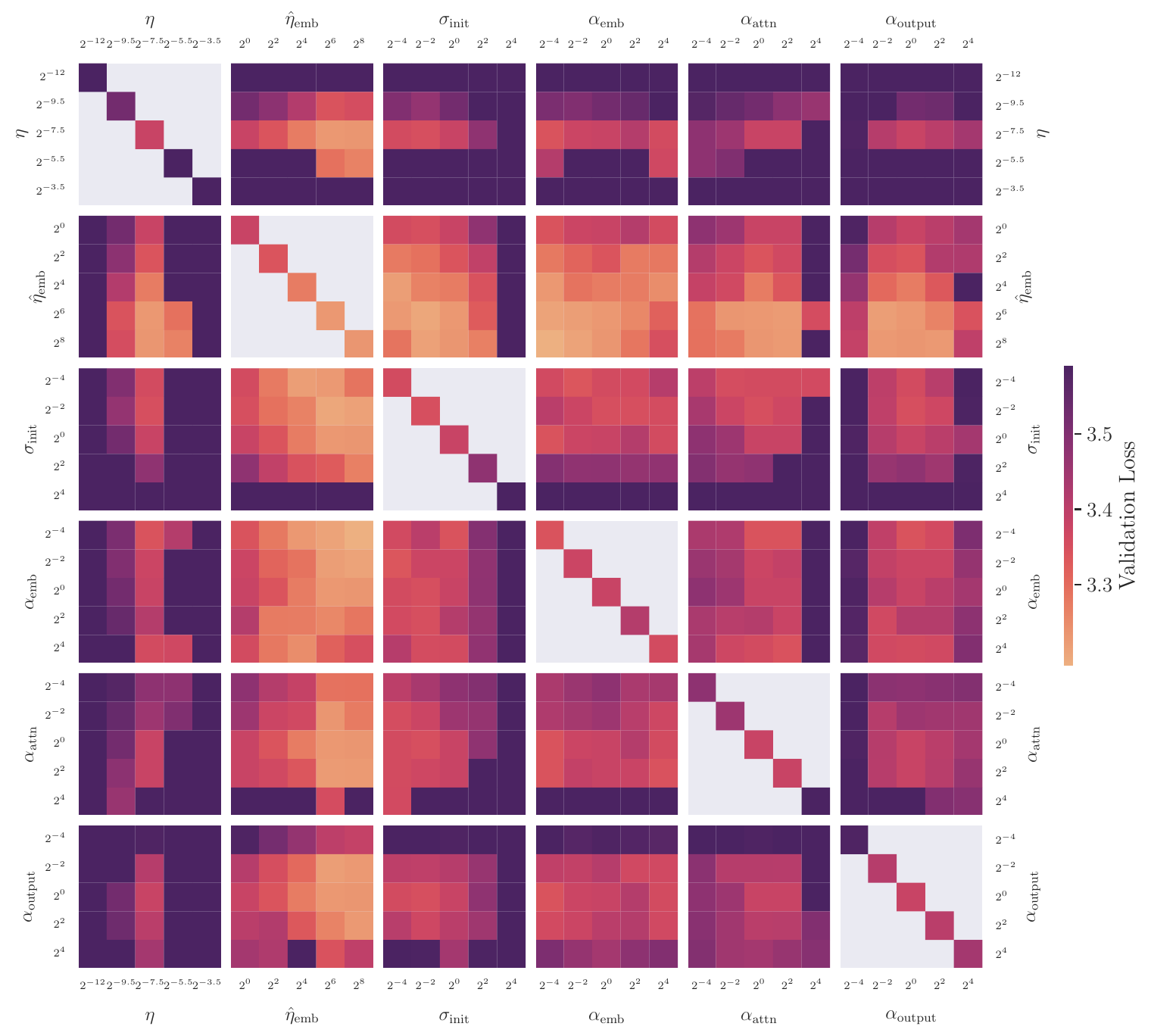}
    \caption{Hyperparameter coupling sweep for \mup{}. Note strong coupling between optima, e.g. in the cases of $(\hat{\eta}_{\mathrm{emb}}, \sigma_{\mathrm{init}})$ and $(\eta, \alpha_{\mathrm{attn}})$. See also: \umup{}, \Cref{fig:additional_experiments:mult_grid_umup}.}
    \label{fig:additional_experiments:mult_grid_mup}
\end{figure}

\begin{figure}[h]
    \centering
    \includegraphics[width=\textwidth]{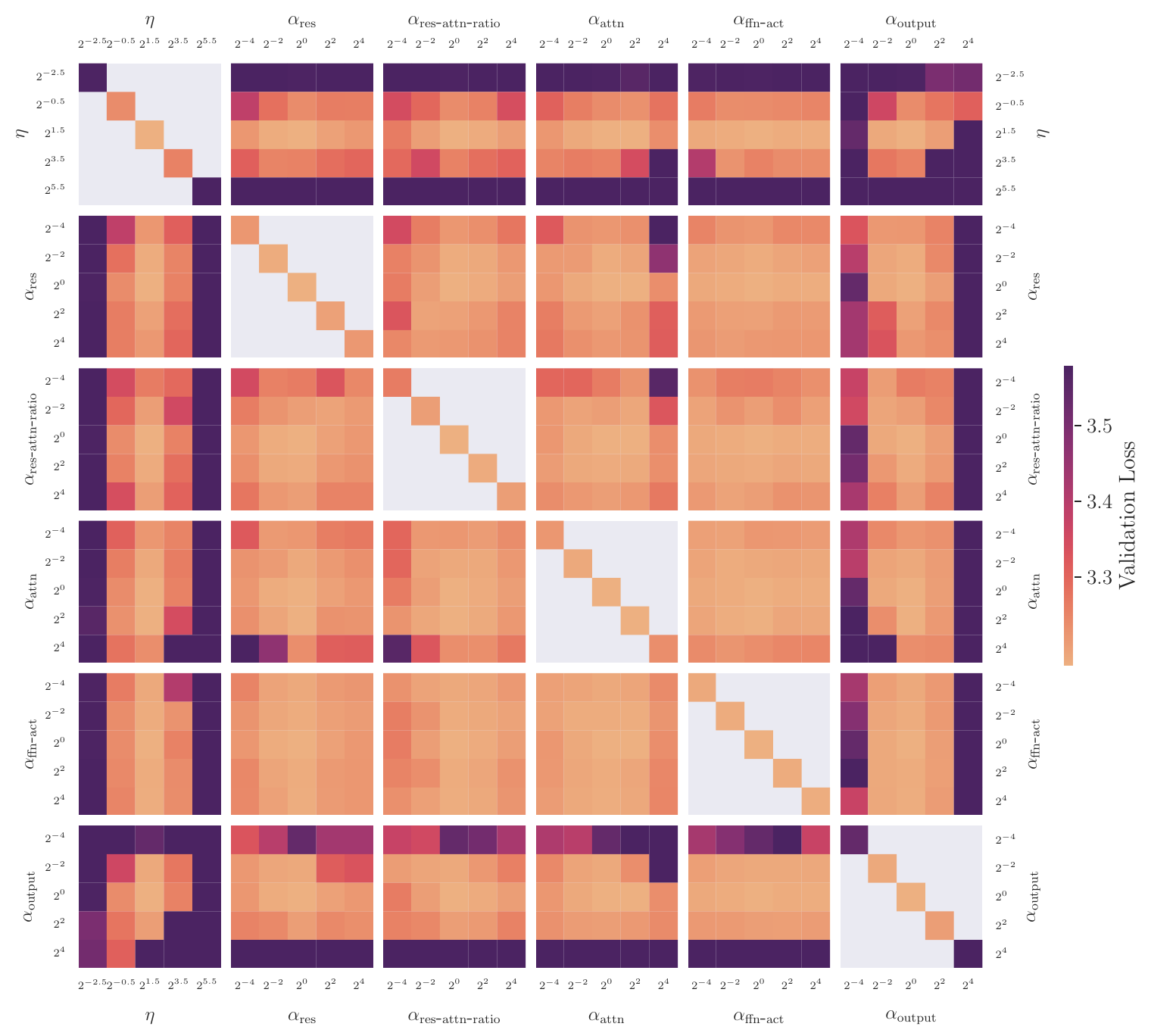}
    \caption{Hyperparameter coupling sweep for \umup{}. Note less coupling than with \mup{}, see \Cref{fig:additional_experiments:mult_grid_mup}.}
    \label{fig:additional_experiments:mult_grid_umup}
\end{figure}

In \Cref{sec:experiments:hp_independence} we explore the question of HP independence under \mup\ and \umup. The following plots in \Cref{fig:additional_experiments:mult_grid_mup,,fig:additional_experiments:mult_grid_umup} show the result of a 2D sweep over every pair of HPs under each scheme. All other HPs are held at 1 when not swept, except the $\eta$ which is held at $2^{-7.5}$ for \mup\ and $2^{1.5}$ for \umup, and $\hat{\eta}_\mathrm{emb}$ which is held at $2^4$ for \mup.

These results show visual dependence between \mup\ hyperparameters as a diagonal structure in the grids, such as $(\hat{\eta}_{\mathrm{emb}}, \sigma_{\mathrm{init}})$ and $(\eta, \alpha_{\mathrm{attn}})$. We quantify this in the plot in \Cref{fig:experiments:transfer_error}, where we use a measure of HP dependence termed transfer error. This is explained verbally in \Cref{sec:experiments:hp_independence}, and we provide an algorithmic description in \Cref{alg:transfer_error}. We note that differences in transfer error between the two methods may also be influenced by the flatness of the optimum. The HP and loss values used for our transfer error calculations are those in \Cref{fig:additional_experiments:mult_grid_mup,,fig:additional_experiments:mult_grid_umup}.

\begin{algorithm}
\caption{Transfer Error} \label{alg:transfer_error}
\begin{algorithmic}
\Require A `fixed' HP with candidate values $F = \{f_1, \cdots, f_n\}$,
a `transfer' HP with candidate values $T = \{t_1, \cdots, t_m\}$, a function that gives the final validation loss for the pair of HPs $L : F \times T \to \mathbb{R}$ (assuming all other HPs are fixed at default values).
\\
\State $\mathrm{err} \gets 0$
\State $f^*, t^* \gets \operatorname{argmin}(L)$
\For{$f$ in $F$}
    \If{$f \neq f^*$}
        \State $t \gets \operatorname{argmin}(L(f))$
        \State $\mathrm{err} \mathrel{+}= L(f^*, t) - L(f^*, t^*)$
    \EndIf
\EndFor
\\
\Return $\mathrm{err}/ (n - 1)$
\end{algorithmic}
\end{algorithm}

\FloatBarrier
\clearpage

\subsection{Hyperparameter search} \label{app:umup_hp_search_algorithm}

Here we outline the particular search processes used for our \mup\ and \umup\ HP sweeps in \Cref{fig:fig1}~(a). The \textit{random search} samples uniformly from a grid defined over all \textit{extended} HPs (extended HP sets are defined in \Cref{table:hp_sets}, with grid values defined in \Cref{tab:experiment_defaults}). We perform the random search over 339 runs, each of which is a full training of the width-256 proxy model. We then simulate the effect of shorter searches at various run-counts by taking a random sample of the results, resulting in the smooth curve over run-count shown.

The \textit{independent search} consists of the following phases:

\begin{enumerate}
    \item Perform a 1D line search for an optimal learning rate, with other hyperparameters set to their default values ($9$ runs).
    \item For each hyperparameter in parallel, perform a 1D line search ($330$ runs).
    \item Combine the best settings from step 2, and re-evaluate ($6$ runs).
\end{enumerate}

The number of runs in the 1D line search is an order of magnitude higher than is required in practice. We do so to form a fair comparison with the random search, which benefits from this large number of runs. The number of runs for the 1D line search could be reduced further by using binary search, though this would require sequential runs and limit the extent of parallelism.

\FloatBarrier

\subsection{Hyperparameter transfer experiments} \label{app:transfer_experiments}

\paragraph{LR transfer over width} The transfer experiments shown in \Cref{fig:fig1} (b) use the non-LR HPs found in \Cref{fig:fig1} (a) (indicated by the circled points), rather than using default HP values. For the \umup\ sweep we take the HPs at the end of the LR portion of the independent search, as these are already close-to-optimal, and means only 9 runs were required in the sweep. In contrast, for \mup\ it is necessary to use the results of the random search over a large number of runs.

\paragraph{LR transfer over other axes} For the training steps, batch size and depth transfer experiments in \Cref{fig:lr_transfer}, all HP values are fixed to 1 except LR which is swept. As with width transfer, \umup\ outperforms \mup\ here using these default HP values. Reducing training steps is done by fixing the number of warm-up steps (at 2000) and still cosine-decaying the learning rate to $10\%$; all that changes is the number of post-warm-up steps. We found this to be more effective than cutting-short the decay schedule.
For both \Cref{fig:fig1} (b) and \Cref{fig:lr_transfer} we sweep the LR over a logarithmically-spaced grid of step $2^{1/2} \times$, with 3 runs for each point.

\begin{figure}[h]
    \includegraphics[width=\textwidth]{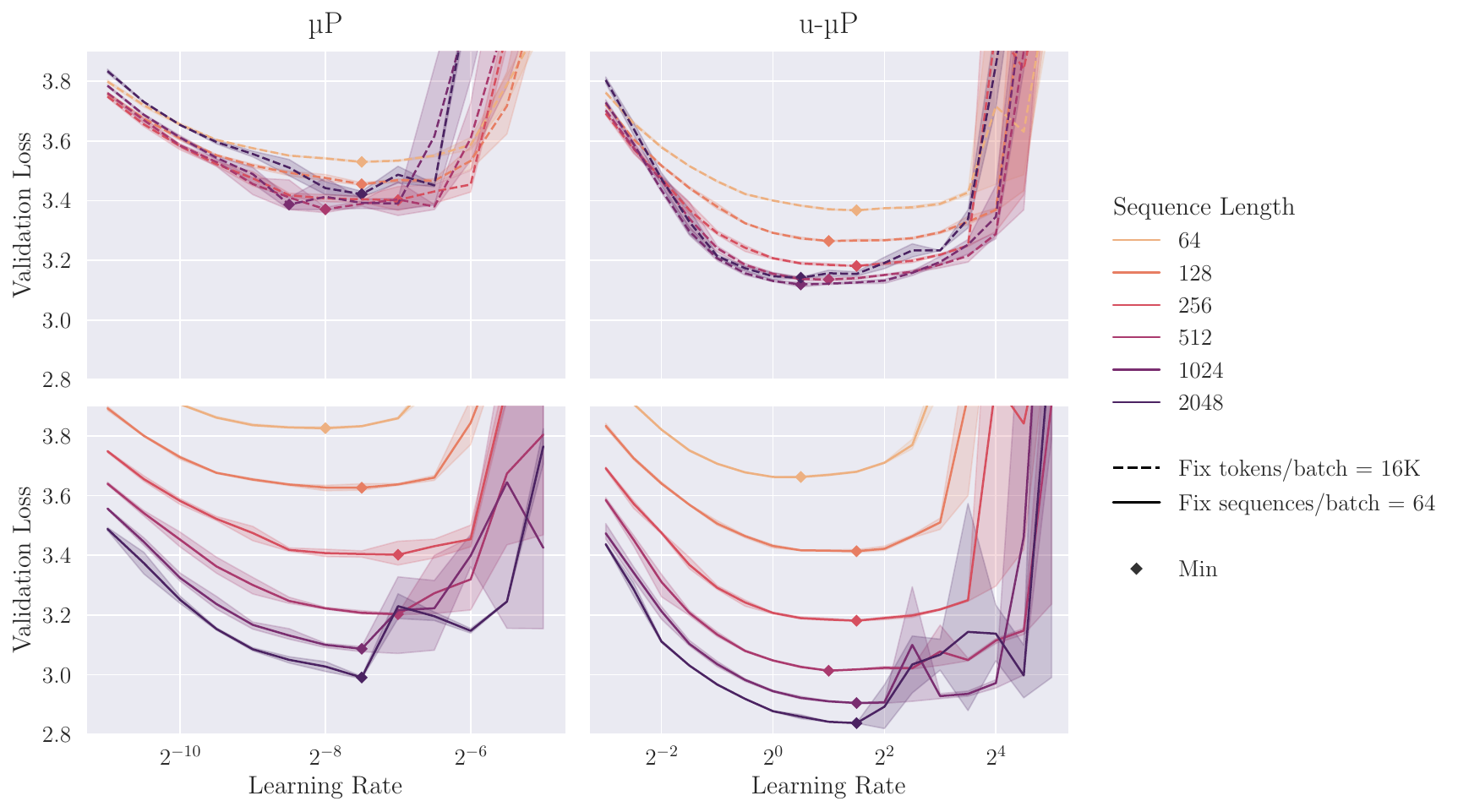}
    \caption{Transfer of learning rate over sequence length for \mup{} (left) and \umup{} (right). As sequence length varies, we can fix the number of tokens per batch by inversely varying the number of sequences per batch (top). Alternatively we can fix the sequences per batch and allow the number of tokens per batch to vary with sequence length (bottom). In the latter case, larger sequence lengths mean the model sees more tokens during training, though as per \Cref{tab:experiment_defaults} this translates to >1 epoch on WikiText-103 when sequence length goes above 256.}
    \label{fig:additional_experiments:lr_transfer_over_seqlen}
\end{figure}

Additionally, in \Cref{fig:additional_experiments:lr_transfer_over_seqlen} we show learning rate transfer over sequence length for both \mup\ and \umup\, fixing either tokens per batch or sequences per batch. In both scenarios \umup\ shows not only better absolute training performance, but also better transfer behaviour as sequence length increases. Since our default proxy sequence length is 256, using \mup\ to transfer to sequence length 2048 would result in minimal improvements or even a degradation in validation loss, whereas the \umup\ shows much greater and more consistent improvements.  

\begin{figure}[h]
    \centering
    \begin{subfigure}{\textwidth}
        \centering
        \includegraphics[width=\textwidth]{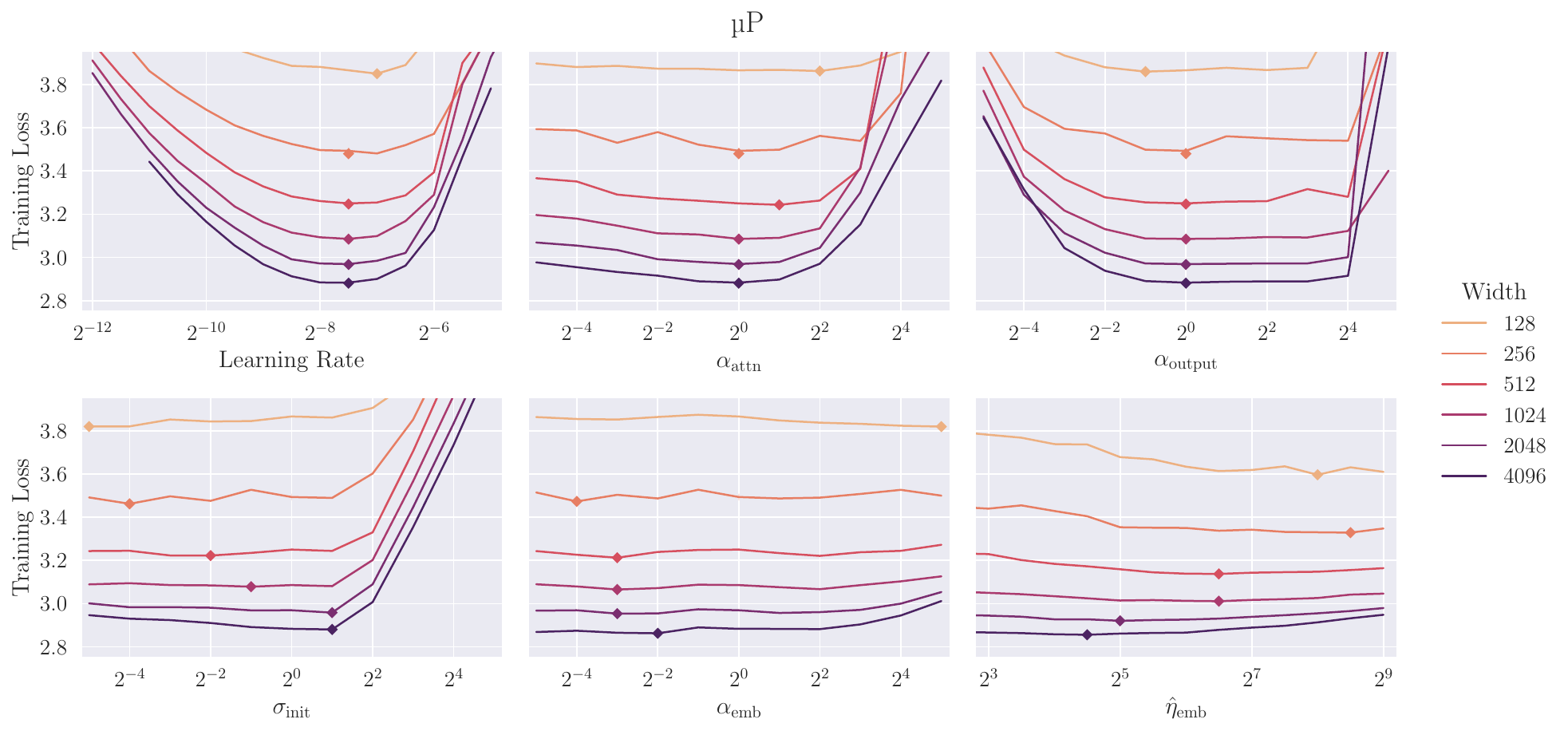}
    \vspace{1em}
    \end{subfigure}
    \begin{subfigure}{\textwidth}
        \centering
        \includegraphics[width=\textwidth]{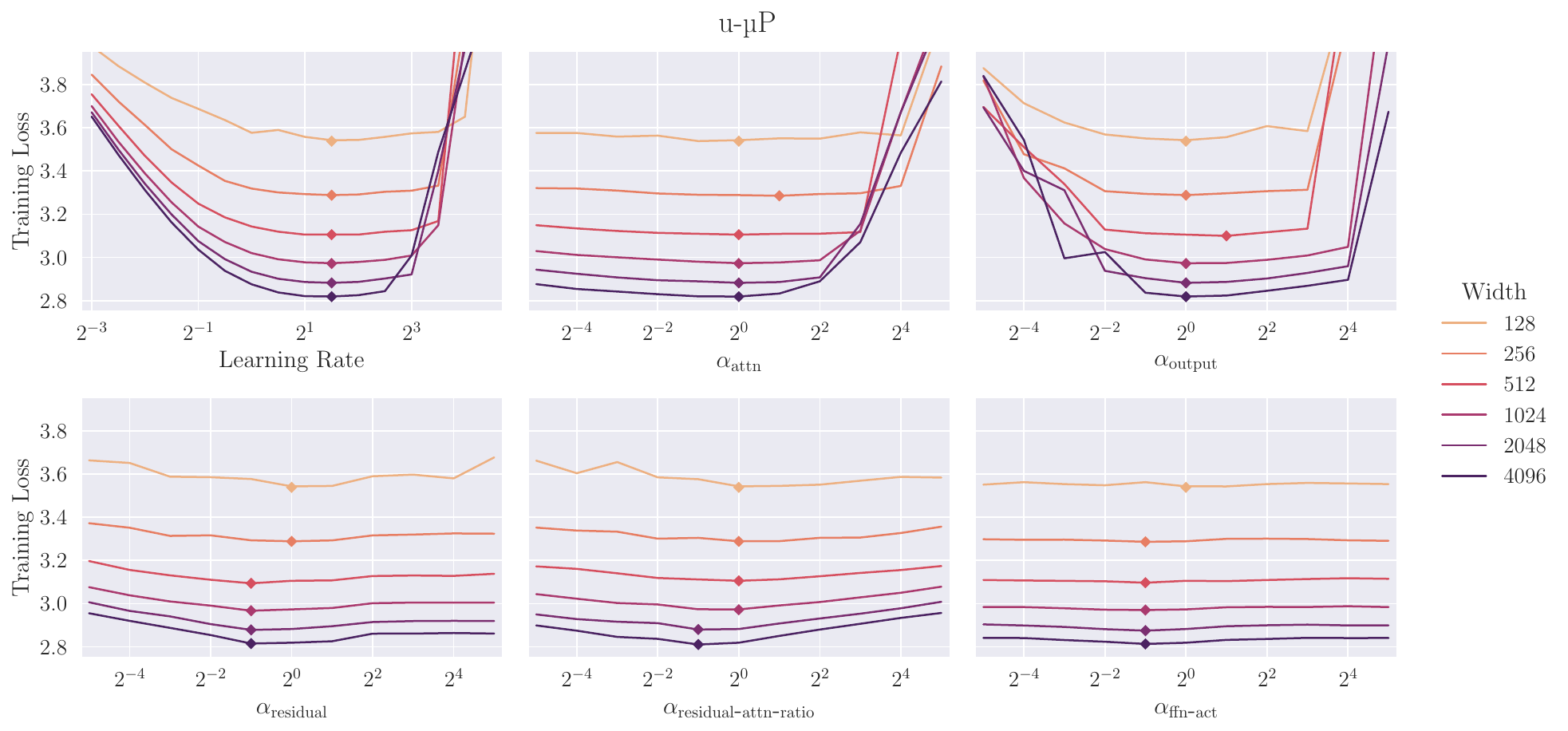}
    \end{subfigure}
    \caption{Transfer of model hyperparameters over width for \mup{} (top) and \umup{} (bottom). When one hyperparameter is being swept, all others are fixed at $1$, with the exception of Learning Rate $\eta=(2^{1.5}, 2^{-7.5})$ for (\umup{}, \mup{}).}
    \label{fig:experiments:hp_transfer_over_width}
\end{figure}

\paragraph{Other HP transfer over width} For our non-LR HP transfer results in \Cref{fig:experiments:hp_transfer_over_width}, we note that good transfer under \mup\ has not been demonstrated for all HPs used in the literature. This is particularly true for the $\hat{\eta}_\mathrm{emb}$ HP, which has poor transfer under \mup. Our investigation here led to our identification of the need to adjust the embedding LR scaling rule outlined in \Cref{sec:umup:emb_lr_rule}. In many cases users have not swept this HP, but instead swept the corresponding parameter multiplier $\alpha_\mathrm{emb}$. How this HP interacts with the embedding LR scaling problem identified (and our proposed fix) remains to be explored, though we note in \Cref{fig:experiments:hp_transfer_over_width} that it also appears to have poor transfer.

\begin{figure}[h]
    \centering
    \includegraphics[width=0.5\textwidth]{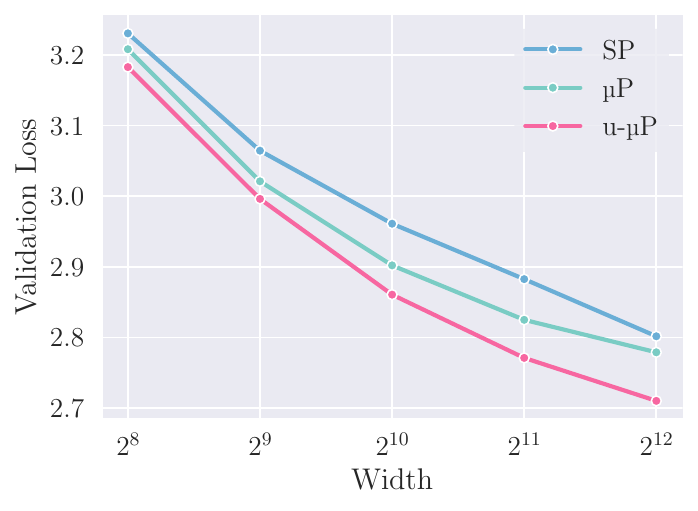}
    \caption{Transferring hyperparameters from width 256 up to 4096 using three different hyperparametrization schemes. \mup\ and \umup\ results are as seen in \Cref{fig:fig1}, whilst Standard Parametrization follows the initialization approach of Pythia \citep{Pythia}.}
    \label{fig:experiments:SP}
\end{figure}

\paragraph{Combined HP transfer}
Whilst \Cref{fig:experiments:hp_transfer_over_width} demonstrates the transfer of individual hyperparameters over width, \Cref{fig:experiments:SP} instead demonstrates the simultaneous transfer of all hyperparameters when co-optimized on the small-scale proxy model, as is done for \mut. The \mup\ and \umup\ points are taken from \Cref{fig:fig1}, with hyperparameters swept on a model of width 256 using a full random HP search and a simple learning rate sweep for \mup\ and \umup\ respectively. The Standard Parametrization scheme, as shown in \Cref{table:hp_sets} requires choosing a learning rate and a weight-initialization scheme. We follow the initialization scheme of Pythia \citep{Pythia}, and transfer learning rate using a heuristic scaling factor of $\nicefrac{\basewidth}{\mathrm{width}}$, as is done in \citep{LLAMA3}.

\FloatBarrier
\clearpage

\subsection{Numerical properties} \label{app:fp8_training}

\begin{figure}[h]
    \centering
    \begin{subfigure}{\textwidth}
        \centering
        \includegraphics[width=\textwidth]{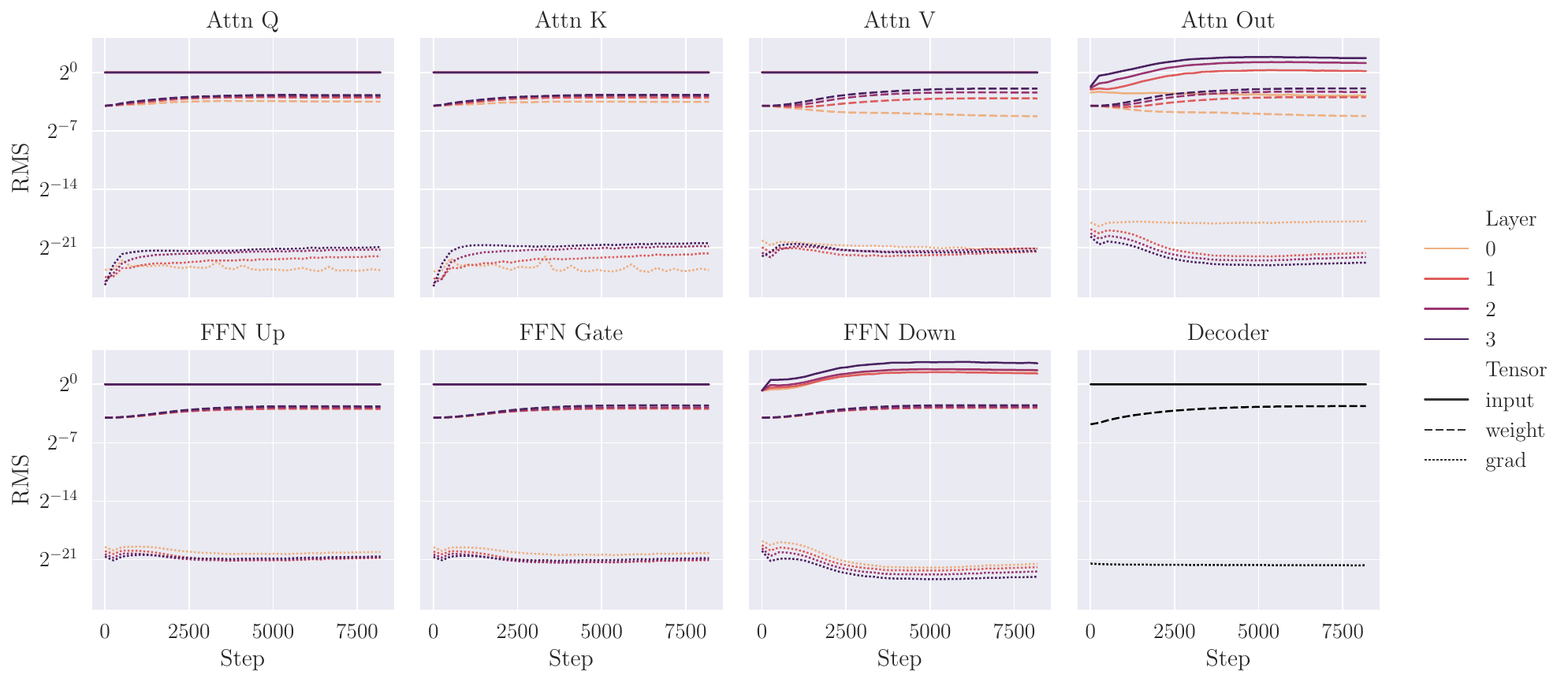}
    \end{subfigure}
    \begin{subfigure}{\textwidth}
        \centering
        \includegraphics[width=\textwidth]{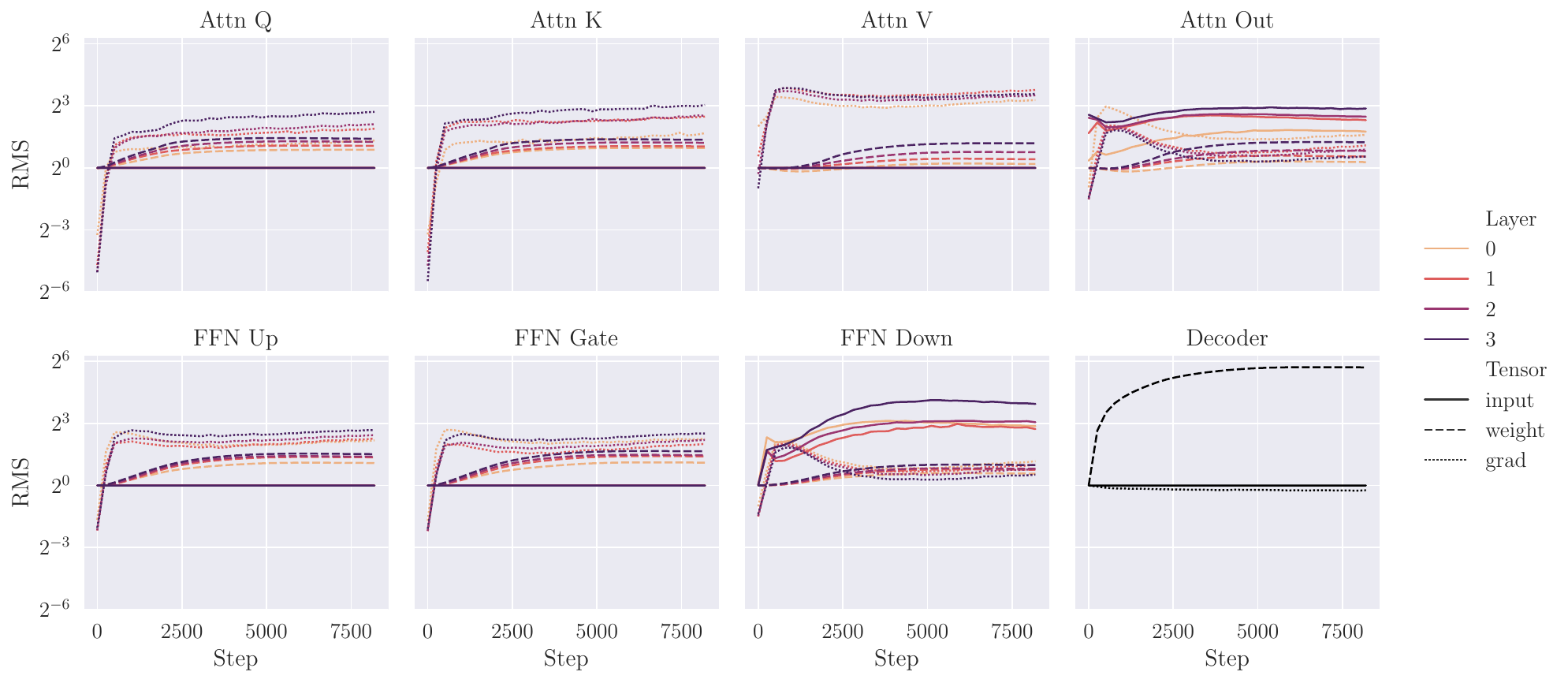}
    \end{subfigure}
    \caption{RMS during training, for all parametrized matmul inputs, for \mup{} (top) and \umup{} (bottom). Model width $256$, default hyperparameters, $\eta=(2^1, 2^{-8})$ for (\umup{}, \mup{}).}
    \label{fig:numerics:rms_during_training}
\end{figure}

Our analysis of the numerical properties of \umup\ focuses on the RMS of tensors that we wish to cast to FP8: linear module input activations, weights and output gradients.
From the RMS training statistics plots in \Cref{fig:numerics:scale} and \Cref{fig:numerics:rms_during_training} we note that

\begin{enumerate}
    \item \mup{} has gradients and weights with low RMS, at risk of FP8 underflow, whereas \umup{} starts with $\mathrm{RMS} \approx 1$.
    \item Many input activations do not grow RMS during training (due to a preceding non-trainable RMSNorm), however the attention out projection and FFN down projection have unconstrained input activations that grow considerably during training.
    \item The decoder weight grows during training. Since it is preceded by a RMSNorm, the model may require scale growth in order to increase the scale of softmax inputs. Other weights grow slightly during training.
    \item Gradients grow quickly but stabilize, except for attention out projection and FFN down projection, whose gradients shrink as the inputs grow.
\end{enumerate}

\begin{figure}[h]
    \centering
    \includegraphics[width=\textwidth]{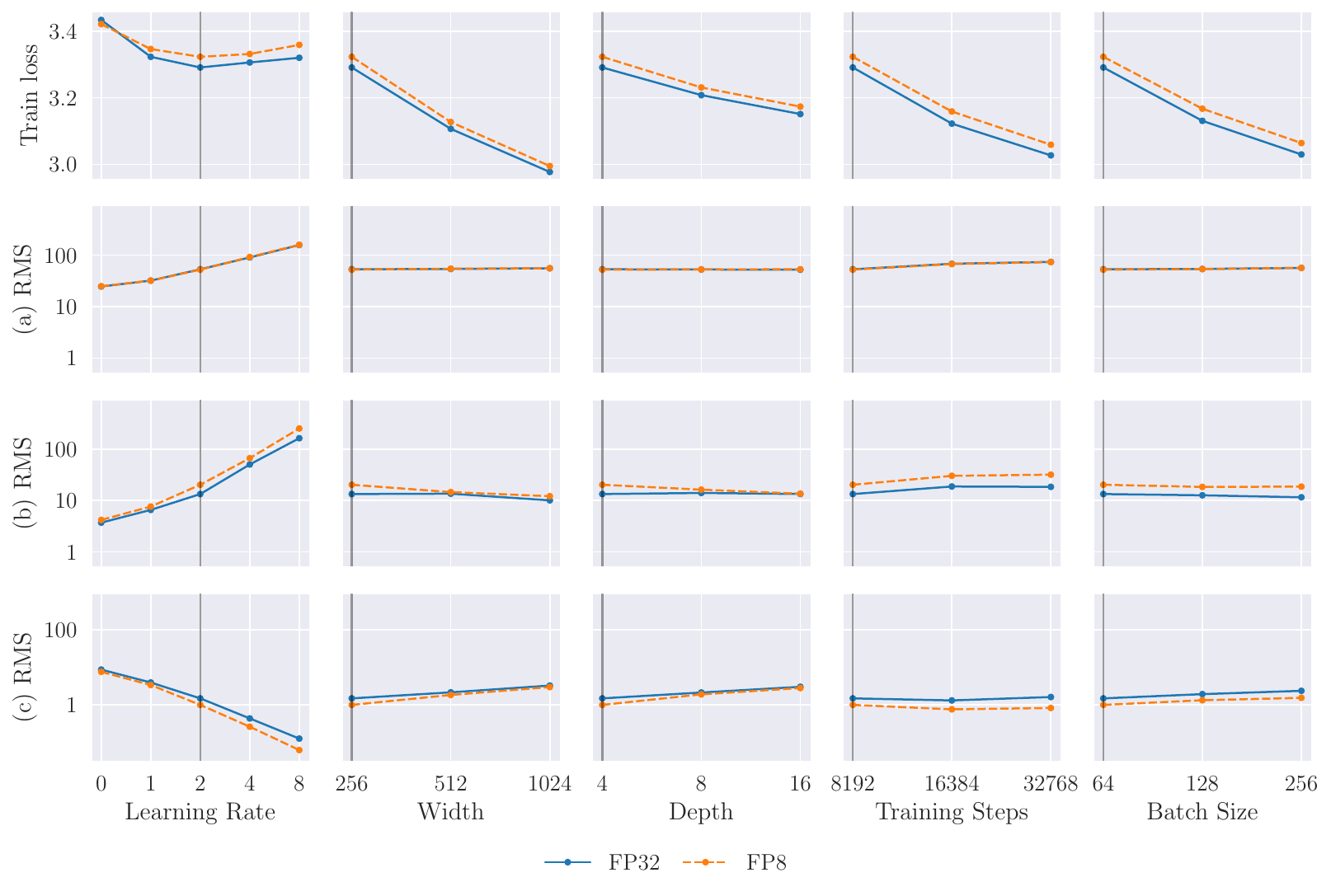}
    \caption{The effect of hyperparameters on FP8 training loss and on the end-training RMS of critical tensors: (a) decoder weight, (b) last-layer FFN down-projection input and (c) last-layer FFN down-projection output gradient. Only learning rate has a substantial effect on the end-training RMS. Vertical lines show the default setting of that hyperparameter, as used for all other plots.}
    \label{fig:numerics:scale_scaling}
\end{figure}

We also evaluate how RMS growth is affected by model and training hyperparameters in the tensors that showed the highest end-training RMS, shown in \Cref{fig:numerics:scale_scaling}. This shows that the main parameter affecting scale growth is learning rate, with end-training RMS increasing to the right of the optimal LR basin, as training becomes unstable. End-training RMS is remarkably stable as width, depth, training steps and batch size are independently increased.

\FloatBarrier
\clearpage



\subsection{Large-scale training details} \label{app:large_model_training}

Our large-scale training settings are given in \Cref{tab:large_scale_experiment_defaults}. These are largely the same as our standard experiments (\Cref{tab:experiment_defaults}), but with many more tokens used for training and scaling up to a larger model-size.
\begin{table}[h] 
    \centering
    \renewcommand{\arraystretch}{1.25}
    \begin{tabular}{|r|p{10cm}|}
    \hline
        Dataset & SlimPajama \citep{SlimPajama} \\
        Sequence length & $4096$ \\
        Vocab size & $65536$ \\
        Training set tokens & $600\mathrm{B}$ \\
        \hline
        Architecture & Llama \citep{Llama} \; (Transformer, PreNorm, RMSNorm, SwiGLU, RoPE, ``untied'' embeddings), non-trainable RMSNorm parameters. \\
        Width & $[2048, 3072, 4096]$ \; (1024 for proxy model) \\
        Depth & $[16, 24, 32]$ \; (8 for proxy model) \\
        Number of heads & $[16, 24, 32]$ \; (8 for proxy model) \\
        Head dimension & $128$ \\
        Total parameters & $[1.07\mathrm{B}, 3.12\mathrm{B}, 6.98\mathrm{B}]$\\
        \hline
        Batch size & $1024$ \\
        Training steps & $72000$ \; ($\sim$ 300B tokens; $20000$ for proxy model) \\
        LR schedule & Cosine to $10\%$, $500$ steps warm-up \\
        Optimizer & AdamW $(\beta_1, \beta_2, \epsilon) = (0.9, 0.95, 10^{-8})$ \\
        Weight decay & $2^{-13}$, independent \citep{Independent_WD} \\
        Dropout & $0.0$ \\
    \hline
    \end{tabular}
    \caption{Large-scale training settings.}
    \label{tab:large_scale_experiment_defaults}
\end{table}

We use mixed-precision during training with optimizer states in FP32 that are sharded via ZeRO stage 1 \citep{Zero}. We retain the model weights in BF16 and apply our FP8 scheme as described in \Cref{sec:umup:low_prec_training} to the tensors participating in matmul operations throughout the transformer block. All other tensors remain either in BF16 (embedding, readout layer, norm, activation function) or FP32 (Flash Attention \citep{Flash_Attention}).


Each model was trained on several Nvidia A100 (80GB) or H100 GPUs, with all FP8 experiments conducted on the H100 chips utilizing their native FP8 support. For the FP8 operations we use PyTorch's \texttt{torch.\_scaled\_mm} function as a backbone.

\FloatBarrier
\clearpage
\section{Unit-scaled op definitions}
\label{app:additional_unit_scaled_ops}

\begin{table}[h]
\vspace{-1em}
\caption{Implementations of unit-scaled ops, building on Table A.2. from the Unit Scaling paper \citep{Unit_Scaling}. These are considered part of \umup\ and should be used in the place of standard operations.}
\label{tab:ops_compendium}
\centering
\vspace{0.6em}
\renewcommand{\arraystretch}{1.25}
\begin{tabular}{lp{6.5cm}}
    \toprule
    Op & Unit Scaling factors \\
    \midrule
    
    $\operatorname{matmul}(x,w)=xw$
    &
    $\alpha=\frac{1}{\sqrt{\fanin}}, \beta_x=\frac{1}{\sqrt{\fanout}}, \beta_w=\frac{1}{\sqrt{\batchsize}}$
    \\[1.5em]

    $\operatorname{attention}(q,k,v)=$
    &
    $\alpha=\beta_q=\beta_k=\beta_v =$
    \\
    $\quad \mathrm{softmax}\left(\alpha_{\mathrm{attn}}  \,d_{\mathrm{head}}^{-1}\, (q k^\top)\, \odot c_\textrm{mask}\right) v$
    &
    $\quad 1/ \operatorname{log\_interpolate}\Bigl(\frac{1}{1 + \frac{4 d_{\mathrm{head}}}{\alpha_{\mathrm{attn}}^2}}, 1, \sqrt{\frac{\log(s)}{s}}\Bigr)$
    \\[2em]

    $\operatorname{gated\_silu}(x_\mathrm{in}, x_\mathrm{gate}) =$
    &
    $\alpha=\beta_{x_\mathrm{in}}=\beta_{x_\mathrm{gate}}=$
    \\
    $\quad x_\mathrm{in} \odot x_\mathrm{gate} \odot \operatorname{sigmoid}(\alpha_{\mathrm{ffn{\text -}act}}\,x_\mathrm{gate})$
    &
    $\quad 1/ \operatorname{log\_interpolate}\Bigl(\frac{1}{1 + \frac{1}{\alpha_{\mathrm{ffn{\text -}act}}^2}}, \frac{1}{\sqrt{2}}, \frac{1}{2}\Bigr)$
    \\[2em]

    $\operatorname{residual\_add}(x_\mathrm{resid.}, x_\mathrm{skip}) =$
    &
    $a=\frac{\tau}{\sqrt{\tau^2 + 1}}, \, b=\frac{1}{\sqrt{\tau^2 + 1}} \quad$
    \\
    $\quad a\, x_\mathrm{resid.} + b\,x_\mathrm{skip}$
    &
    (see \ref{subsubsec:umup_residual_in_full} for full details, inc. values for $\tau$, which depends on $\alpha_\mathrm{res}$ and $\alpha_\mathrm{res\text{-}attn\text{-}ratio}$.)
    \\[1.5em]

    $\operatorname{softmax\_xent}(x, t) =$
    &
    \\
    $\quad \operatorname{log\_softmax(\alpha_\mathrm{loss\text{-}softmax} \, x)}_t$
    &
    $\alpha=1, \; \beta = s / \sqrt{s-1}$
    \\[1.5em]

    $\operatorname{RoPE}(x)$
    &
    $\alpha=\beta=1 \quad$ (i.e. no scaling)
    \\[1.5em]

    $\operatorname{RMSNorm}(x)$ (non-trainable, see \citep{Exploration_Of_Mu_Transfer})
    &
    $\alpha=\beta=1 \quad$ (i.e. no scaling)
    \\[0.5em]
    
    \bottomrule
    \vspace{1em}
\end{tabular}
\end{table}

\FloatBarrier

The original Unit Scaling paper provides scaling factors for various ops, in order to make them unit-scaled. However, these ops do not cover every case required for the Llama architecture used in our experiments, nor do they cover our updated residual layer implementation. 
To address this, in this section we outline a series of new unit-scaled ops for each of our required architectural features, as well as existing unit-scaled ops, as given in \Cref{tab:ops_compendium}.

The presentation here is derived from that of the Unit Scaling Compendium given in \citep[Table A.2]{Unit_Scaling}. This makes reference to the factors $\alpha, \beta_1, \dots, \beta_k$. $\alpha$ is the output scaling factor in the forward pass, and $\beta_i$ are the scaling factors for the gradient of the op's inputs in the backward pass. For each op, a value or rule is provided for determining the required mult to ensure unit-scale. The correct value for these multipliers is derived by analyzing the scaling behavior of each op, given some reasonable distributional assumptions about the input and incoming gradient tensors (see \Cref{app:additional_background:us} for an example). Below we provide an in-depth overview of each new or modified unit-scaled op we introduce here.

\paragraph{Unit-scaled dot-product attention}

The Unit Scaling paper considers the attention layer scaling in terms of its separate components: the various matmul operations and the internal softmax. Linear operations are scaled using the standard rule, and the softmax scaling is given a $\alpha = \beta = s$ factor.

From an implementation perspective, the self-attention layer is more typically broken down into weight-matmuls and a fused scaled-dot-product attention operation. This is the case we handle here, accounting for three complicating factors not considered in the Unit Scaling paper:
\begin{enumerate}
    \item As we use a decoder-style transformer in our experiments, our softmax operation has a causal mask applied to its input.
    \item We follow the \mup\ guidance of using $1/ d_{\text head}$ scaling in our self-attention layer, rather than the usual $1 / \sqrt{d_{\text head}}$.
    \item We place a $\alpha_{\mathrm{attn}}$ multiplier immediately before the softmax, which is an HP that users may tune.
\end{enumerate}
As a result our dot-product attention takes the form:
\begin{align*}
    \operatorname{attention}(q,k,v) &= \mathrm{softmax}\left(\alpha_{\mathrm{attn\text{-}softmax}} \cdot d_{\mathrm{head}}^{-1} \cdot (q \cdot k^\top) \odot c_\textrm{mask}\right) \cdot v
\end{align*}
The addition of an HP before the softmax introduces an additional challenge for Unit Scaling, as our scaling multipliers will need to account for this value when preserving unit scale.

This operation is sufficiently complex that we found an empirical model of its scale to be more accurate than any mathematically-derived rule (future work may consider justifying our model mathematically). We find that the scale of dot-product attention is approximately
\begin{align*}
    \sigma(\operatorname{attention}(q,k,v)) &= \operatorname{log\_interpolate}\left(\frac{1}{1 + \frac{4 d_{\mathrm{head}}}{\alpha_{\mathrm{attn}}^2}}, 1, \sqrt{\frac{\log(s)}{s}}\right)
\end{align*}
where
\begin{align*}
    \operatorname{log\_interpolate}(\alpha, b_\mathrm{upper}, b_\mathrm{lower}) &= e^{\alpha \log(b_\mathrm{upper}) + (1 - \alpha) \log(b_\mathrm{lower})}.
\end{align*}
The corresponding scaling rule is therefore to divide by this factor in both the forward and backward pass, as outlined in \Cref{tab:ops_compendium}.

\paragraph{SwiGLU FFN}

Llama uses a SwiGLU \citep{GLU} layer for its FFN, which introduces two new operations for us to unit-scale: a SiLU \citep{SiLU} (a.k.a. swish \citep{Swish}) operation and an element-wise multiplication. We take a similar approach to our dot-product attention, and consider unit-scaling the following fused operation:
\begin{align*}
    \operatorname{gated\_silu}(x_\mathrm{in}, x_\mathrm{gate}) &= x_\mathrm{in} \odot x_\mathrm{gate} \odot \operatorname{sigmoid}(\alpha_{\mathrm{ffn{\text -}act}}\,x_\mathrm{gate})
\end{align*}
For the surrounding weight-matmuls we follow the standard Unit Scaling rules.

Again, we use an empirical model of the scale of this op, which is surprisingly similar to the dot-product attention model:
\begin{align*}
    \sigma(\operatorname{gated\_silu}(x_\mathrm{in}, x_\mathrm{gate})) &= \operatorname{log\_interpolate}\left(\frac{1}{1 + \frac{1}{\alpha_{\mathrm{ffn{\text -}act}}^2}}, \frac{1}{\sqrt{2}}, \frac{1}{2}\right),
\end{align*}
dividing through by this factor to get our scaling rule.

\paragraph{Residual layers}

Our implementation of residual layers for \umup\ is more complex than other operations, as adjustments are required to:
\begin{enumerate}
    \item Make pre-norm residual networks support Unit Scaling (see \Cref{sec:us_residuals}).
    \item Introduce our new, principled residual HPs (see \Cref{sec:umup_hparams}).
\end{enumerate}
Our residual layer scheme is presented in full in \ref{subsubsec:umup_residual_in_full}. For readers interested in our justification for this, see the sections noted above.

We also follow the example of Unit Scaling and delay the application of our residual multiplier in the backward pass to the base of the branch (see \citep{Unit_Scaling}, Figure 3c). This does not change the model, and enables unit-scale to be maintained on the residual branch regardless of the value of the multiplier.

\paragraph{RoPE embeddings}

We also require a unit-scaled implementation of Rotary Position Embeddings (RoPE \citep{RoPE}), which are applied just before the scaled dot-product attention operation. As RoPE essentially consists of pair-wise rotations of elements by different degrees, we observe no meaningful scale-change as a result of it's application, and hence leave it unchanged.

\paragraph{RMSNorm}

Following \citep{Exploration_Of_Mu_Transfer} we opt to use a non-trainable version of RMSNorm \citep{RMS_Norm}, in order to facilitate better transfer. As a result, we also leave this operation unchanged. Were a trainable RMSNorm to be used, the recipe would follow closely that of the LayerNorm presented in the original Unit Scaling Compendium.

\paragraph{Scale constraints}

One final, minor deviation from the scheme outlined in the Unit Scaling paper is the way in which we apply scale constraints (see their Section 5.2). The essence of scale constraints is that for perfect unit scaling, sometimes the ideal scale for the forward pass differs from those in the backward pass. In some special cases (e.g. at the ends of the network) the use of different scales can be valid, but in the general case a single scale must be agreed upon. The solution in the Unit Scaling paper is to use the geometric mean of the forward and backward scales.

We propose instead to simply use the forward scale over the backward scale(s) in these cases. We do so for the following reasons:
\begin{enumerate}
    \item For these architectures we find empirically that where there is a disparity in ideal forward and backward scales, it is not large.
    \item By taking the forward scale, we can ensure strict unit-scale in the forward pass.
\end{enumerate}
The value of the latter point is in terms of what it means for the interpretation of our \umup\ multiplier HPs. Consider the $\alpha_{\mathrm{ffn{\text -}act}}$ multiplier; with strict unit scale we can say that the standard deviation of activations immediately before this multiplier is 1. Therefore the standard deviation immediately after is $\alpha_{\mathrm{ffn{\text -}act}}$. As this multiplier is (by design) the last operation before the ffn activation function, we can say that the interpretation of $\alpha_{\mathrm{ffn{\text -}act}}$ is simply to set the input standard deviation to the FFN's activation function.
Similar arguments can be made for other \umup\ multiplier HPs. This interpretation only holds because we use the forward-scale in our constraints.

\section{A guide to using \umup} \label{app:using_umup_guide}

We bring together our \umup\ scheme presented in \Cref{sec:umup} to form a simple recipe for applying it to a model. The \umup\ scheme is designed and validated on a Llama-style architecture, so it may not be applicable or effective on other models, particularly those with substantially different architectures. Exploring this question is an important avenue for future work.

Before applying our scheme, users are encouraged to apply the following pre-requisites to their training setup, based on our analysis of effective \mut\ in \Cref{sec:challenges:mut}:

\begin{itemize}
    \item Remove trainable parameters from normalization layers
    \item Use the \textit{independent} form of AdamW
    \item Ensure training is in the under-fitting regime (i.e. avoid excessive data repetition)
\end{itemize}

Having done this, our recipe for using \umup\ is as follows:

\begin{enumerate}
    \item \textbf{Replace operations \& optimizers with \umup\ versions:}
    Each operation should be replaced by a unit-scaled version (these wrap the existing operations, with added static scales in the forward and backward passes). We have pre-calculated scales for common operations in \Cref{app:additional_unit_scaled_ops}. Parameters should be initialized with unit variance, and Adam(W) adjusted to use the scaling rules defined in \Cref{sec:umup:emb_lr_rule} (we refer to the optimizer as Adam in this section, but AdamW should be used if weight decay is required. Other optimizer scaling rules can be determined by the same process we outline). These features are all implemented in our library (see \Cref{app:us_lib_guide}).

    \item \textbf{Choose a set of HPs to sweep:}
    From the set of HPs outlined in \Cref{table:hp_sets}, select those to be swept. We recommend the extended set, though a basic LR sweep can be effective.
    
    \item \textbf{Decide on proxy model config:}
    The cost of proxy model training should be such that the sweeping process is much less than target model training, while still being as representative as possible. We base our recommendations on the results in \Cref{fig:lr_transfer}. In general, width is the most reliable feature to transfer. Training steps and batch size also give good transfer, so moderate changes here are permissible. Depth is the least reliable feature for transfer, so we only recommend modest changes in depth. We keep the number of warmup steps constant, but always decay to the same final LR when varying the number of steps.
    
    \item \textbf{Perform independent HP search:}
    Following the process outlined in \Cref{sec:experiments:hp_independence} and \Cref{app:umup_hp_search_algorithm}.
    
    \item \textbf{Train the target model:}
    This can be done in FP8 simply by placing casts on matmul inputs (though for our large-scale experiments we found the scales of two operations drifted enough over time that some lightweight dynamic re-scaling was required).
\end{enumerate}

The above functionality is provided in the Unit Scaling library, to avoid users having to implement it themselves, and to provide a reference implementation. We provide a guide to using this library in the following section.
\section{A guide to the \texttt{unit scaling} library} \label{app:us_lib_guide}

Our PyTorch \citep{pytorch} extension library, released under an open source license at \url{https://github.com/graphcore-research/unit-scaling}, accompanies this paper to provide standard and reference implementations of \umup{} operations and optimizers.

This section provides an overview of the functionality of the library; please consult the repository documentation for details. A good place to start is our demo of a simple \umup\ training implementation: \url{https://github.com/graphcore-research/unit-scaling/blob/main/examples/demo.ipynb}.

\subsection{Standard usage}

Compared with SP, \umup{} requires the insertion of appropriate scaling factors in the forward and backward pass, a different parameter initialization scheme and the application of learning rate scaling rules based on the role and shape of each parameter.

The library provides implementations of ops in \texttt{unit\_scaling.functional} with appropriate scaling rules (\Cref{tab:ops_compendium}). Non-homogeneous ops (\Cref{app:non-homogeneous}) have optional \emph{mults}, which are hyperparameters controlling shape of non-linear operations and the interpolation between mutiple inputs. Ops may also specify \emph{constraints}, which are used to satisfy the cut-edge rule (\Cref{app:cut_edge_rule}). Although this rule could be automated as a global graph transformation, the library makes constraint selection an explicit step for the user, while providing sensible defaults. For example, weight gradients are generally cut-edges, so are unconstrained.

Parameters are \emph{tagged} with their role in the model (as a ``bias'', ``norm'' parameter, ``weight'' or ``output''). The library achieves this by extending \texttt{torch.nn.Parameter} with an additional property \texttt{mup\_type}. This property is required for every parameter in a \umup{} model. Given this, and information on the overall depth of the model, the library applies the learning rate rules of \Cref{table:mup_umup_schemes} as a pre-optimizer transformation that modifies the learning rate for each parameter. This allows standard PyTorch optimizers to be used without modification.

PyTorch uses \emph{modules} to encapsulate parameter declaration, initialization and the calling of ops. The library makes available \umup{} versions of common modules, which declare tagged parameters, apply unit-scale initialization, and call unit-scaled ops, with appropriate default settings.

With these components, user code for training using \umup{} is very close to that of vanilla PyTorch (see an example in \Cref{fig:umup_example}).

\begin{figure}[h]
\codefig{umup_example.py}
\caption{Using the \texttt{unit scaling} library given the tensors \texttt{input\_} \& \texttt{target}.}
\label{fig:umup_example}
\end{figure}

\subsection{Extending the library}

As the set of deep learning ops of interest is always growing, the unit-scaling library is open for extension. For example, consider the possible implementation of unit-scaled \texttt{hardtanh(x) = clip(x, -1, 1)} in \Cref{fig:umup_extension_hardtanh}.

\begin{figure}[h]
\codefig{umup_extension_hardtanh.py}
\caption{Implementing new unit-scaled operations.}
\label{fig:umup_extension_hardtanh}
\end{figure}

The implementation follows a standard pattern:

\begin{enumerate}
\item Calculate the theoretical or empirical scaling factors for each forward and backward pass independently, based on independent unit-scaled Gaussian inputs.
\item Apply the optional constraint to select or combine these scaling factors, using the helper function \texttt{apply\_constraint}.
\item Call \texttt{scale\_bwd} on the inputs, and \texttt{scale\_fwd} on the outputs to compensate for scaling after the op or grad-op is executed.
\end{enumerate}

It can be checked empirically using random inputs and gradients (example in \Cref{fig:umup_testing_hardtanh}).

\begin{figure}[h]
\codefig{umup_testing_hardtanh.py}
\caption{Testing unit-scaled operations, using \texttt{constraint=None} to allow independent fwd and bwd scaling. The default constraint \texttt{"to\_output\_scale"} preserves forward-pass scale while constraining the forward and backward scales to be equal.}
\label{fig:umup_testing_hardtanh}
\end{figure}

\subsection{As a reference implementation}

The core technique of \umup{} is readily implementable in most deep learning frameworks; the primary requirement is for custom gradient operations in order to provide equivalents of \texttt{scale\_fwd} and \texttt{scale\_bwd}. We hope that the library provides a useful reference, as well as a set of tools and techniques for developing custom \umup{} support in other libraries and projects.

\section{Additional background material}

\subsection{The Maximal Update Parametrization} \label{app:additional_background:mup}

\paragraph{Theoretical background} We do not cover the theory underpinning \mup\ in this paper, presenting only its resulting scaling rules (\Cref{table:mup}). For readers interested in this theory, the extensive Tensor Programs series \citep{Tensor_Programs_I, Tensor_Programs_II, Tensor_Programs_IIb, Tensor_Programs_III, Tensor_Programs_IVb} builds up a framework from which \mup\ is derived \citep{Tensor_Programs_IV}. For those requiring a more accessible introduction, \citep{Spectral_Condition} show that \mup\ can be derived in a simpler and more general way by placing a spectral scaling condition on the norm of weights and their updates.

\paragraph{Approaches to HP sweeping in the literature}

\Cref{table:mup_hps} outlines the ways in which users of \mup\ in the literature have approached HP sweeping. These all follow the approach used in Tensor Programs V of a random sweep, sampling combinations from the joint space of all HPs. The authors of Tensor Programs V note that other more complex methods may be more efficient, but these are considered beyond the scope of their work and have not been used widely. A Bayesian search method was used for the development of MiniCPM~\citep{MiniCPM}, but the authors give no further details---as they use 400 runs in their sweep it is not clear that this approach makes HP search easier.

\begin{table}[h]
  \centering
  \caption{Sweeping configurations used for a selection of \mup\ models from the literature.
  The sweeping process is similar across models, the only differences being the choice of discrete or continuous distributions and their ranges.}
  \label{table:mup_hps}
  \begin{tabular}{cccccl}
      \toprule
      Model & \thead{proxy/target\\tokens used} & \thead{proxy/target\\model size} & \thead{sweep\\size} & \thead{base\\width} & \thead{HPs swept}
      \\
      \midrule
      T.P.V WMT14~\citep{Tensor_Programs_V} & 100\% & 7.1\% & 64 & \multirow{3}{*}{?} & $\eta, \alpha_\textrm{out}, \alpha_\textrm{attn}$
      \\
      T.P.V $\text{BERT}_\text{large}$~\citep{Tensor_Programs_V} & 10\% & 3.7\% & 256 & & $\eta, \eta_\textrm{emb}, \alpha_\textrm{out}, \alpha_\textrm{attn}, \alpha_\textrm{LN}, \alpha_\textrm{bias}$
      \\
      T.P.V GPT-3~\citep{Tensor_Programs_V} & 1.3\% & 0.6\% & 350 & & $\eta, \sigma, \alpha_\textrm{emb}, \alpha_\textrm{out}, \alpha_\textrm{attn}, \alpha_\textrm{pos}$
      \\
      MiniCPM~\citep{MiniCPM} & 0.008\% & 0.45\% & 400 & 256 & $\eta, \sigma, \alpha_\textrm{emb}, \alpha_\textrm{residual}$
      \\
      Cerebras-GPT~\citep{Cerebras_GPT} & 1.1\% & 1.5\% & 200 & 256 & $\eta, \sigma, \alpha_\textrm{emb}$
      \\
      S\mup ar~\citep{Supar} & 6.6\% & 6.4\% & 350 & 256 & $\eta, \sigma, \alpha_\textrm{emb}$
      \\
      \bottomrule
  \end{tabular}
\end{table}

\FloatBarrier

\subsection{Unit Scaling} \label{app:additional_background:us}

\paragraph{An example: the unit-scaled matmul op} Here we outline the procedure for calculating the scaling factor of a matmul op, which practitioners can use as a guide for scaling new ops that we do not cover in this paper (see \Cref{app:additional_unit_scaled_ops}).

There are two potential approaches here. The first is to derive scaling factors from an analysis of an op's dynamics. Specifically, given the assumption of unit-scaled inputs, the appropriate scaling factor is the reciprocal of the expected output scale. For a basic matrix-matrix matmul we have,
\begin{align*}
    &\operatorname{matmul}(X, W) = XW, \quad\quad X \in \mathbb{R}^{d_\mathrm{batch} \times d_\fanin},\; W \in \mathbb{R}^{d_\fanin \times d_\fanout},
\end{align*}
where weights and activations are sampled i.i.d. from a centered Gaussian:
\begin{align*}
    X_{ij} \sim \mathcal{N}(0, \sigma_X^2), \; W_{jk} \sim \mathcal{N}(0, \sigma_W^2).
\end{align*}
From this we can derive the expected output scale (i.e. $\sigma(\operatorname{matmul})$):
\begin{align*}
    \operatorname{matmul}(X, W)_{ik} = \sum_{j=1}^{d_\fanin} X_{ij}W_{jk},
    \\
    \sigma\left(\operatorname{matmul}(X, W)_{ik} \right) = \sqrt{d_\fanin} \, \sigma_W \, \sigma_X.
\end{align*}
Under Unit Scaling we have $\sigma_W = \sigma_X = 1$, and hence the scaling factor required to ensure a unit-scaled output is $1/\sqrt{d_\fanin}$. This gives our final unit-scaled matmul:
\begin{align*}
    &\operatorname{u-matmul}(X, W) = \operatorname{matmul}(X, W) / \sqrt{d_\fanin}
\end{align*}

The distributional assumptions made here hold at initialization, but do not over training. A more precise model for the asymptotic behavior of neural networks under training is given by the Tensor Programs framework, but for the purposes of numerics this precise treatment of scale at initialization appears to be sufficient.

The second, less ideal approach to calculating scaling factors is to use experimentation to infer this relationship empirically. In this case, one would sample random initializations and compute the output scale over a range of $d_\fanin$ values (or whatever HPs one expects the output scale to depend on), fitting a curve to the observed data.

\paragraph{Applying unit scaling} 

To apply Unit Scaling to a model and train in low-precision, the following steps are required:

\begin{enumerate}
    \item Scale parameter initializations to have zero-mean and unit variance.
    \item Replace operations with their unit-scaled equivalents (including and especially the loss, matmuls and residual-adds).
    \item \textit{Constrain} the scales of operations which are required to have the same forward and backward factors.
    \item Place a simple \texttt{.to(fp8)} cast on the inputs to matmuls.
\end{enumerate}

Step 3 relates to the problem of conflicting scales in the forward and backward passes. A single linear layer in a differentiated model requires 3 matmul ops in the forward and backward passes, each requiring a different scaling factor ($\frac{1}{\sqrt{d_\fanin}}, \frac{1}{\sqrt{d_\fanout}}, \frac{1}{\sqrt{d_\batchsize}}$). However, using these directly would give invalid gradients. The compromise here is that the activations and activation gradients have their scaling factors \textit{constrained} such that they are equal (the original Unit Scaling paper recommends taking the geometric mean; we modify this for \umup\ in \Cref{app:additional_unit_scaled_ops} to simply use the forward scale everywhere). Weight gradients can still be given their own scaling factor due to the \textit{cut-edge rule} (as explained in \Cref{app:cut_edge_rule}).

Step 4 reflects the key benefit of Unit Scaling. Unlike other methods it changes the learning dynamics of a model, but the advantage is that unit-scaled models then `naturally' generate well-scaled tensors. This means that low-precision arithmetic ideally becomes as simple as placing a cast operation before matmuls as outlined.
\section{Unit-scaled pre-norm residual layers}  \label{sec:us_residuals}

The popular pre-norm residual network architecture is simple to implement, but problematic to combine with Unit Scaling. It exhibits scale-growth in the skip-stream at initialization, due to the repeated addition of residual connections without subsequent normalization. Here we present a surprising and useful finding: that for any pre-norm model there exists a mathematically-equivalent model where this scale-growth is eliminated, through the careful re-scaling of residual connections.

Note that this section focuses on applying Unit Scaling to \textit{standard} pre-norm models. Only once we have addressed this problem are we able to do the same for \umup\ models, as shown in \Cref{subsec:residual_branch_multipliers}. Readers only interested in our final \umup\ residual implementation may skip ahead to \Cref{subsubsec:umup_residual_in_full}.




\subsection{Scale growth in pre-norm residual networks}

Let's consider a pre-norm residual network of depth $L$:
\begin{align}
    R_0(x) &= r_0 x, \label{eq:resnet_a}
    \\
    R_{l}(x) &= r_{l}f_l(R_{l-1}(x)) + R_{l-1}(x), \quad l=1,..,L \label{eq:resnet_b}
    \\
    R_{L+1}(x) &= f_{L+1}(R_L(x)) \label{eq:resnet_c}
\end{align}
with embedding multiplier $r_0$ and residual branch multipliers $r_l$ for $l=1,..,L$. To satisfy pre-norm, all $f_l$ are zero-homogeneous functions, i.e. $f_l(\lambda x) = f_l(x)$.

The scale of the skip-stream at initialization as a result of \Cref{eq:resnet_b} is
\begin{align}
    \sigma(R_l) &= \sqrt{r_l^2 \sigma(f_l)^2 + \sigma(R_{l-1})^2}
    > \sigma(R_{l-1}), \quad l=1,..,L
    \label{eq:residual_non_unit_scale}
\end{align}
assuming $r_l^2 \sigma(f_l)^2 > 0$. This shows that scale inevitably grows with the addition of each residual layer.

This scale-growth is clearly incompatible with unit scaling, which aims for $\sigma(R_l) = 1$ for all $l=0,..,L+1$.
In the following we present an elegant solution to this problem making use of a symmetry transformation available in pre-norm residual architectures.

\subsection{Residual symmetry in pre-norm architectures}

To resolve the problem of scale shift in residual networks demonstrated by \Cref{eq:residual_non_unit_scale}, we try a slightly more general ansatz:
\begin{align}
    \hat{R}_0(x) &= x,  \label{eq:general_resnet_a}
    \\
    \hat{R}_{l}(x) &= a_l f_l(\hat{R}_{l-1}(x)) + b_l\hat{R}_{l-1}(x),  \label{eq:general_resnet_b}
    \\
    \hat{R}_{L+1}(x) &= f_{L+1}(\hat{R}_L(x)) \label{eq:general_resnet_c}
\end{align}
with coefficients $a_l, b_l$. We want to choose these coefficients so that the outputs of $\hat{R}_{l}$ are unit-scaled if the outputs $f_l, \hat{R}_{l-1}$ are. A similar calculation as in \Cref{eq:residual_non_unit_scale} leads to the sufficient condition
\begin{align}
    a_l^2 + b_l^2 = 1, \label{eq:a_b_sq_sum}
\end{align}
which can be easily satisfied. Having restored Unit Scale, we are faced with another issue. It seems that \Cref{eq:general_resnet_a,eq:general_resnet_b,eq:general_resnet_c} describe a different network than \Cref{eq:resnet_a,eq:resnet_b,eq:resnet_c}, whereas ideally the relation from input to final output should be unchanged when converting the network to Unit Scaling. 

Note that the coefficients $a_l, b_l$ are not uniquely defined yet, so our mathematical intuition tells us that we should find an additional constraint to get a unique solution. To find this constraint, let us consider our original residual network in \Cref{eq:resnet_a,eq:resnet_b,eq:resnet_c} and analyze how the variance propagates through the network if we assume all the $f_l$ satisfy Unit Scaling and $\sigma(x) = 1$. Let $\sigma_{l-1}^2$ denote the variance of $R_{l-1}$. Then a simple inductive calculation shows that
\begin{align*}
    \sigma_{l-1}^2 =  \sum_{i=0}^{l-1}r_i^2.
\end{align*}
By \Cref{eq:resnet_b} the output of $R_l$ adds a quantity of scale $r_l$ from the residual connection and a quantity of scale $\sigma_{l-1}$ from the skip connection. Intuitively, the \textit{ratio} of these scales should be more important for the overall network dynamics than their absolute values. Thus our constraint becomes preserving the ratio of scales from the original model, through our choice of $a_l, b_l$:
\begin{align*}
    \frac{a_l}{b_l} = \frac{\sigma(r_l f_l)}{\sigma_{l-1}} = \frac{r_l}{\sqrt{\sum_{i=0}^{l-1}r_i^2}} =: \tau_l,
\end{align*}
which, recalling \Cref{eq:a_b_sq_sum}, (up to sign) uniquely defines our multipliers $a_l, b_l$ as 
\begin{align}
    a_l = \frac{\tau_l}{\sqrt{\tau_l^2 + 1}}, \quad b_l = \frac{1}{\sqrt{\tau_l^2 + 1}}
\end{align}
In summary, we propose the modified residual network
\begin{align}
    \hat{R}_0(x) &= x, \label{eq:umup_resnet_a}
    \\
    \hat{R}_{l}(x) &= \frac{\tau_l}{\sqrt{\tau_l^2 + 1}}f_l(\hat{R}_{l-1}(x)) + \frac{1}{\sqrt{\tau_l^2 + 1}}\hat{R}_{l-1}(x), \label{eq:umup_resnet_b}
    \\
    \hat{R}_{L+1}(x) &= f_{L+1}(\hat{R}_L(x)),  \label{eq:umup_resnet_c}
    \\
    \tau^2_l &= \frac{r^2_l}{\sum_{i=0}^{l-1}r_i^2}. \label{eq:umup_resnet_d}
\end{align}
Our main result of this section is that this network is indeed mathematically equivalent to the network defined in \Cref{eq:resnet_a,eq:resnet_b,eq:resnet_c}, under a simple additional structural assumption:
\begin{lem} \label{lem:residual_symmetry}
    Consider $R_l$, $\hat{R}_l$ defined as in \Cref{eq:resnet_b,eq:umup_resnet_b} respectively. Then $\hat{R}_l = R_l / \sqrt{\sum_{i=0}^l r_i^2}$ for all $l=0,..,L$.
\end{lem}
Remarkably, this result does not assume the individual network operations $f_l$ actually satisfy Unit Scaling. It is purely a consequence of the pre-norm residual structure. However, only under Unit Scaling can the factors $\tau_l$ be interpreted as the ratio of scales between skip and residual branch.

As a consequence of the lemma, the final residual output $R_L(x)$ is the same as in our original network up to a fixed multiplier. Due to the zero-homogeneity of the final output function $f_{L+1}$ this gives $\hat{R}_{L+1} = f_{L+1}\left(R_L(x)/\sqrt{\sum_{i=0}^l r_i^2}\right) = f_{L+1}(R_L(x)) = R_{L+1}$, proving the mathematical equivalence of our residual scheme.
Modern LLM architectures like Llama~\citep{Llama} are pre-norm residual networks of this kind. Hence they admit a faithful unit-scaled reparametrization.

\subsection{Proof of Lemma~\ref{lem:residual_symmetry}}
\begin{proof}
    This is proved by induction. For the base-case $l=1$, we have $\tau_1 = r_1/r_0$, giving
    \begin{align*}
        \hat{R}_{1}(x) &= \frac{\tau_l}{\sqrt{\tau_l^2 + 1}}f_1(x) + \frac{1}{\sqrt{\tau^2_l + 1}}x
        \\
        &= (r_1f_1(x) + r_0x) / \sqrt{r_0^2 + r_1^2}
        \\
        &= R_1 / \sqrt{r_0^2 + r_1^2}.
    \end{align*}
    Then if the statement holds for $l-1$ we have
    \begin{align*}
        \hat{R}_{l}(x) &= \frac{\tau_l}{\sqrt{\tau^2_l + 1}}f_l(\hat{R}_{l-1}(x)) + \frac{1}{\sqrt{\tau^2_l + 1}}\hat{R}_{l-1}(x)
        \\
        &= \frac{r_l}{\sqrt{\sum_{i=0}^{l}r_i^2}}f_l(\hat{R}_{l-1}(x)) + \frac{\sqrt{ \sum_{i=0}^{l-1}r_i^2}}{\sqrt{\sum_{i=0}^{l}r_i^2}}\hat{R}_{l-1}(x)
        \\
        &= \left(r_l f_l(\hat{R}_{l-1}(x)) + \sqrt{\sum_{i=0}^{l-1}r_i^2} \hat{R}_{l-1}(x)\right) / \sqrt{\sum_{i=0}^{l}r_i^2}
        \\
        &= \left(r_l f_l({R}_{l-1}(x)) + \sqrt{\sum_{i=0}^{l-1}r_i^2} \frac{{R}_{l-1}(x)}{\sqrt{\sum_{i=0}^{l-1}r_i^2}}\right) / \sqrt{\sum_{i=0}^{l}r_i^2}
        \\
        &= \left(r_l f_l({R}_{l-1}(x)) + {R}_{l-1}(x)\right) / \sqrt{\sum_{i=0}^{l}r_i^2}
        \\
        &= R_l(x) / \sqrt{\sum_{i=0}^{l}r_i^2}
    \end{align*}
\end{proof}

\subsection{Unit Scaling for transformer residuals} \label{subsec:unit_scaled_transformer_residuals}

The above scheme describes Unit Scaling for arbitrary pre-norm residual networks. We now apply it to the case of pre-norm transformer residual layers.

We can describe a transformer in terms of the residual network given in \Cref{eq:resnet_a,eq:resnet_b,eq:resnet_c}. Our $f_l$ functions alternate between self-attention layers and feed-forward layers. Implementations differ in the handling of how residual multipliers $r_l$ correspond to HPs. In many cases practitioners simply ignore these $r_l$, but for the sake of expressivity we assume the two types of residual layer each have their own HP, as well as the embedding. In other words,

\begin{equation*}
    r_l = \begin{cases}
        \alpha_{\mathrm{emb}} & l = 0
        \\
        \alpha_{\mathrm{attn{\text -}residual}} & l \textrm{ is odd }
        \\
        \alpha_{\mathrm{ffn{\text -}residual}} & l \textrm{ is even, and } l > 0.
    \end{cases}
\end{equation*}

To convert this to a Unit Scaled network we apply \Cref{eq:umup_resnet_a,eq:umup_resnet_b,eq:umup_resnet_c,eq:umup_resnet_d}, from which can derive the following closed-form expression for $\tau_l$:

\begin{equation*}
    \tau^2_l = \begin{dcases}
        \frac{\alpha^2_{\mathrm{attn{\text -}residual}}}{\alpha^2_{\mathrm{emb}} + \ell \alpha^2_{\mathrm{attn{\text -}residual}} + \ell \alpha^2_{\mathrm{ffn{\text -}residual}}} & l \textrm{ is odd }
        \\
        \frac{\alpha^2_{\mathrm{ffn{\text -}residual}}}{\alpha^2_{\mathrm{emb}} + (\ell + 1) \alpha^2_{\mathrm{attn{\text -}residual}} + \ell \alpha^2_{\mathrm{ffn{\text -}residual}}} & l \textrm{ is even}.
    \end{dcases}
\end{equation*}

where $\ell = \lfloor{\frac{l-1}{2}}\rfloor$.

This gives us a unit-scaled pre-norm residual implementation for a \textit{standard} transformer, which is mathematically equivalent to a non-unit-scaled version. In the next section we augment this by adding in two HPs, in a carefully-designed manner that satisfies our criteria for \umup\ HPs, giving us our full residual implementation.

\section{Justifying the \umup\ hyperparameter scheme} \label{sec:umup_hparams}

Here we justify our particular choice of \umup\ HP, as given in \Cref{table:hp_sets} (with their placement defined in \Cref{tab:ops_compendium}). We discuss this topic briefly in \Cref{sec:umup:principled_hps}, stating that all our HPs (excepting the LR) are $\alpha$ HPs, and under \umup\ they are now associated with operations instead of weights. All operations have an $\alpha$ HPs, unless they are unary and $k$-homogeneous for $k \ge 0$.

We begin this section by explaining why we apply this rule to the model and how it results in three of our \umup\ HPs. We then consider how best to hyperparametrize our residual layers, building on our criteria for HPs given in \Cref{sec:umup:principled_hps} and the unit-scaled pre-norm residual scheme in \Cref{sec:us_residuals}.

\subsection{Multipliers for non-homogeneous ops: $\alpha_{\mathrm{attn{\text -}softmax}},\; \alpha_{\mathrm{ffn{\text -}act}},\; \alpha_{\mathrm{loss{\text -}softmax}}$} \label{app:non-homogeneous}

In this section we derive the rest of our \umup\ multipliers. We want to identify the minimal set that can still express all different choices of pre-op scales in the model. The crucial observation is that every pre-scale multiplier $\alpha$ of a unary operation $h \mapsto f(\alpha h)$ can be propagated through the network if $f$ is $k$-homogeneous for some $k>0$, i.e. $f(\alpha x) = \alpha^k f(x)$, leaving the model and its optimization unchanged. We can iterate this along the computational path until either the next operation is non-homogeneous, non-unary (we are at the end of a residual path), or the next operation is 0-homogeneous (e.g. a norm).

In the first case the accumulated scales are absorbed in the pre-op scale of the non-homogeneous operation (where we introduce a multiplier), in the second case they are absorbed in the residual addition for that branch (where we again introduce a multiplier), and in the final case the scale disappears (so we start over). We now go through the Llama forward computation and follow this paradigm to identify our multipliers in \Cref{tab:scale_prop}.

\begin{table*}[ht]
\caption{A walkthrough of the Llama architecture, showing how our $\alpha_{\mathrm{attn{\text -}softmax}}$, $\alpha_{\mathrm{ffn{\text -}act}}$ and $\alpha_{\mathrm{loss{\text -}softmax}}$ multipliers are derived via an analysis of scale-propagation.}
\label{tab:scale_prop}
\centering
\vspace{0.6em}
\renewcommand{\arraystretch}{1.25}
\begin{tabular}{p{3.5cm}p{9.75cm}}
    \toprule
    \textbf{Op} & \textbf{Scale propagation behavior} \\
    \midrule
    Embedding & We show in \Cref{subsubsec:improved_residual_HPs} that the embedding multiplier can be absorbed in the residual multipliers, meaning one is not required here.
    \\\midrule
    Attention RMSNorm & This operation is $0$-homogeneous and thus we start over.
    \\
    Query \& key projection & Both are linear, meaning their scale is propagated. Multipliers are therefore not required.
    \\
    Query-key matmul & Again linear. As query \& key are both generated from the same input, this operation is $2$-homogeneous wrt. that input. Hence it also propagates scale.
    \\
    Softmax & The softmax operation is non-homogeneous. Thus the pre-op scale of the softmax becomes our first multiplier: $\alpha_{\mathrm{attn{\text -}softmax}}$.
    \\
    Value & The value layer is linear and hence propagates scale.
    \\
    Softmax-value matmul & Again linear and hence propagates scale.
    \\
    Attention projection & This operation is linear and lies at the end of the attention residual path. Hence there are no more multipliers in the attention block.
    \\
    Residual add & This operation is non-unary and hence receives our second (and third) multipliers: $\alpha_{\mathrm{res}}, \; \alpha_{\mathrm{res{\text -}attn{\text -}ratio}}$. The manner and motivation for using two multipliers here is justified in the next section.
    \\\midrule
    FFN RMSNorm & This operation is $0$-homogeneous and thus we start over.
    \\
    FFN input scale & The input layer is linear, hence it propagates scale.
    \\
    Sigmoid input & This function is non-homogeneous and thus we have our fourth multiplier: $\alpha_{\mathrm{ffn{\text -}act}}$.
    \\
    SiLU weight & This layer is also linear and propagates scale.
    \\
    Product & The entry-wise multiplication of the outputs of sigmoid, input layer and SiLU weight is homogeneous and thus propagates scale.
    \\
    FFN output & This layer is linear and at the end of the residual path. Hence there are no more multipliers in the FFN residual block.
    \\
    Residual add & See above.
    \\\midrule
    Output RMSNorm & This operation is $0$-homogeneous and thus we start over.
    \\
    Output head & This layer is linear, hence it propagates scale.
    \\
    Loss & The cross-entropy loss is non-homogeneous and leads to our final multiplier: $\alpha_{\mathrm{loss{\text -}softmax}}$.
    \\
    \bottomrule
\end{tabular}
\end{table*}

\subsection{Residual branch multipliers: $\alpha_{\mathrm{res}}, \; \alpha_{\mathrm{res{\text -}attn{\text -}ratio}}$}
\label{subsec:residual_branch_multipliers}

In this section we derive our two \umup\ residual HPs. We start with the basic, non-unit scaled model we began with in the previous section, outlined in \Cref{eq:resnet_a,eq:resnet_b,eq:resnet_c}. We described a set of $\alpha_{\mathrm{emb}}, \alpha_{\mathrm{attn{\text -}residual}}, \alpha_{\mathrm{ffn{\text -}residual}}$ HPs associated with this model in \Cref{subsec:unit_scaled_transformer_residuals}. However these HPs poorly satisfy our cardinality, independence and interpretability criteria from \Cref{sec:umup:principled_hps}, so in the \Cref{subsubsec:improved_residual_HPs} we present a re-parametrization of these HPs designed to better satisfy these points. In \Cref{subsubsec:umup_residual_in_full} we then combine these HPs with the final unit-scaled pre-norm residual scheme we derived in \Cref{sec:us_residuals}, resulting in our complete \umup\ residual scheme.

\subsubsection{Improved hyperparameters for transformer residuals} \label{subsubsec:improved_residual_HPs}


To avoid cluttered notation, in this section we rename
\begin{align*}
    &\alpha_{\mathrm{res}} = \alpha_r, \quad
    \alpha_{\mathrm{res{\text -}attn{\text -}ratio}} = \alpha_\rho\\
    \alpha_{\mathrm{emb}} = \alpha_e&, \quad
    \alpha_{\mathrm{attn{\text -}residual}} = \alpha_a \quad
    \alpha_{\mathrm{ffn{\text -}residual}} = \alpha_f.
\end{align*}
To make the presentation more clear, we derive our new HPs using the standard residual scheme from \Cref{eq:resnet_a,eq:resnet_b,eq:resnet_c}. For the actual unit scaled implementation one needs to transform the multipliers following \Cref{eq:umup_resnet_a,eq:umup_resnet_b,eq:umup_resnet_c,eq:umup_resnet_d}, which we do in Section~\ref{subsubsec:umup_residual_in_full}.


To facilitate our analysis, we can view the transformer residual output as the sum of three terms:
\begin{align*}
    R_L &= R_L^{(e)} + R_L^{(a)} + R_L^{(f)},
    \\
    R_L^{(e)} &:= \alpha_e x,
    \\
    R_L^{(a)} &:= \sum_{l=1}^{L/2} \frac{\alpha_a}{\sqrt{L/2}} f_{2l-1}(R_{2l-1}(x)),
    \\
    R_L^{(f)} &:= \sum_{l=1}^{L/2} \frac{\alpha_f}{\sqrt{L/2}} f_{2l}(R_{2l}(x)),
\end{align*}
and define the average residual scale,
\begin{equation*}
    \sigma(R_L^{(a,f)})^2 := \frac{\sigma(R_L^{(a)})^2 + \sigma(R_L^{(f)})^2}{2}.
\end{equation*}
Note that we have added in the \depthmup\ multipliers here, though a similar analysis can be performed for non-\depthmup\ models. As above, $f_l$ functions alternate between self-attention layers and feed-forward layers.

With respect to our interpretability criterion, we propose two new multipliers that correspond to dynamics in the network which we suggest are important to control at initialization. The first is the ratio of the average scale of the residuals' contributions to those of the embedding, $\alpha_r = \sigma(R_L^{(a,f)}) / \sigma(R_L^{(e)})$. The second is the ratio of the scale of the attention-residuals' contributions to those of the feed-forward-residuals, $\alpha_{\rho} = \sigma(R_L^{(a)}) / \sigma(R_L^{(f)})$. Not only do these two ratios control key dynamics of our model, but we can use them to replace our existing $(\alpha_e, \alpha_a, \alpha_f)$ multipliers. 

Let us first examine these two quantities for a standard (non-unit-scaled model). Residual functions of the same kind have the same expected output scale at initialization in pre-norm networks, meaning we can denote the output scale $\sigma(f_l(R_l))$ of all self-attention functions as $\sigma_a$, and of all feed-forward functions as $\sigma_f$. We thus have the following scales at the output:
\begin{align*}
    \sigma(R_L^{(e)}) &= \alpha_e \sigma(x),
    \\
    \sigma(R_L^{(a)}) &= \frac{\alpha_a}{\sqrt{L/2}}\, \sigma\!\left(\sum_{i=1}^{L/2} f_{2l-1}(R_{2l-1})\right) = \alpha_a \sigma_a,
    \\
    \sigma(R_L^{(f)}) &= \frac{\alpha_f}{\sqrt{L/2}}\, \sigma\!\left(\sum_{i=1}^{L/2} f_{2l}(R_{2l})\right) = \alpha_f \sigma_f,
    \\
    \sigma(R_L^{(a,f)}) &= \sqrt{\frac{(\alpha_a \sigma_a)^2 + (\alpha_f \sigma_f)^2}{2}}.
    \\
\end{align*}
Recalling our definitions of $\alpha_r, \alpha_{\rho}$ above, this gives us:
\begin{align*}
    \alpha_\rho &= \frac{\alpha_a}{\alpha_f} \frac{\sigma_a}{\sigma_f},
    \\
    \alpha_r &= \sqrt{\frac{(\alpha_a \sigma_a)^2 + (\alpha_f \sigma_f)^2}{2\,(\alpha_e \sigma(x))^2}},
    \\
    &= \sqrt{\frac{\alpha_\rho^2 + 1}{2}} \frac{\sigma_f}{\sigma(x)} \frac{\alpha_f}{\alpha_e}.
\end{align*}
The original $\alpha_a, \alpha_f$ multipliers can then be written in terms of $\alpha_r, \alpha_\rho$:
\begin{align*}
    \alpha_a &= \alpha_\rho \alpha_f \frac{\sigma_f}{\sigma_a}
    \\
    \alpha_f &= \alpha_r \alpha_e \frac{\sigma(x)}{\sigma_f} \sqrt{\frac{2}{\alpha_\rho^2 + 1}}
\end{align*}
We have replaced two of the three original multipliers, but still have a dependence on $\alpha_e$ here in our expressions for $\alpha_f$ and $R_L^{(e)}$, which we now remove by dividing it out of our residual branches and embedding. We use the hat ($\hat{\cdot}$) symbol to denote terms that have been divided-through by $\alpha_e$. This new system of equations is equivalent to our old one thanks to the zero-homogeneity of the final post-residual layer:
\begin{align*}
    R_{L+1}(x) &= f_{L+1}(R_L^{(e)} + R_L^{(a)} + R_L^{(f)})
    \\
    &= f_{L+1}((R_L^{(e)} + R_L^{(a)} + R_L^{(f)})/\alpha_e)
    \\
    &= f_{L+1}(\hat{R}_L^{(e)} + \hat{R}_L^{(a)} + \hat{R}_L^{(f)})
\end{align*}
This gives $\hat{R}_L^{(e)} = \alpha_e x / \alpha_e = x$, removing our first occurrence of $\alpha_e$. Following the division through $\hat{R}_L^{(a)}$ and $\hat{R}_L^{(f)}$, we obtain:
\begin{align*}
    \hat{R}_L^{(a)} &:= \sum_{l=1}^{L/2} \frac{\hat{\alpha}_a}{\sqrt{L/2}} f_{2l-1}(R_{2l-1}),
    \\
    \hat{R}_L^{(f)} &:= \sum_{l=1}^{L/2} \frac{\hat{\alpha}_f}{\sqrt{L/2}} f_{2l}(R_{2l}),
    \\
    \hat{\alpha}_a &= \alpha_\rho \hat{\alpha}_f \frac{\sigma_f}{\sigma_a},
    \\
    \hat{\alpha}_f &= \alpha_r \frac{\sigma(x)}{\sigma_f} \sqrt{\frac{2}{\alpha_\rho^2 + 1}}.
\end{align*}

This system of equations is the same as the original, but with the two $\alpha_e$ terms dropped, meaning our model's multipliers can be expressed in terms of only $\alpha_r$ and $\alpha_\rho$. Using the above equations, any pair of values for $(\alpha_r, \alpha_\rho)$ can be translated back into an equivalent set of values for $(\alpha_e, \alpha_a, \alpha_f)$ such that the output $R_{L+1}(x)$ is the same, meaning that our multipliers are no less expressive than the original set. This satisfies our desired criteria of minimizing the number of multipliers while maintaining expressivity.

We can simplify further in the case of unit-scaled models, which are designed such that $\sigma(x), \sigma_a, \sigma_f$ are all $1$ at initialization. In this case our re-parametrization becomes:
\begin{align}
    \hat{\alpha}_a &= \alpha_\rho \hat{\alpha}_f,
    \label{eq:us_residual_eqs_a} \\
    \hat{\alpha}_f &= \alpha_r \sqrt{\frac{2}{\alpha_\rho^2 + 1}},
    \label{eq:us_residual_eqs_b} \\
    \hat{\alpha}_e &= 1.
    \label{eq:us_residual_eqs_c}
\end{align}

This is the basis of our claim that Unit Scaling is what enables a more intuitive set of multipliers. Not only do the multipliers $\alpha_r$ and $\alpha_\rho$ represent important dynamics in the network at initialization (the ratio of residual-to-embedding scales, and the ratio of attention-to-feed-forward scales), but it's only via unit scaling that these equations become simple enough to implement in practice. Using equations \Cref{eq:us_residual_eqs_a,eq:us_residual_eqs_b,eq:us_residual_eqs_c} for a non-unit scaled network may still be effective, but the interpretation we've given to $\alpha_r$ and $\alpha_\rho$ no longer hold.

Our final desired property is an empirical one: that the most effective choice of one multiplier depends as little as possible on the choice of the other multiplier(s). We demonstrate that our multipliers satisfy this property better than the standard set of residual multipliers in \Cref{sec:experiments:hp_independence}.

\subsubsection{The full \umup\ residual scheme} \label{subsubsec:umup_residual_in_full}

Here we give the full definition of our \umup\ residual scheme, summarizing the results of previous sections. A general pre-norm transformer is implemented as:

\begin{align} \label{eq:final_full_residual_eq}
    R_0(x) &= c\,x,
    \\
    R_{l}(x) &= a_{l}f_l(R_{l-1}(x)) + b_lR_{l-1}(x), \quad l=1,..,L
    \\
    R_{L+1}(x) &= f_{L+1}(R_L(x)),
\end{align}
where $a_l, b_l$ and $c$ are scalar multipliers, and the $f_l$ alternate between self-attention and feed-forward layers. We consider our baseline set of \mup\ residual HPs here to be $(\alpha_{\mathrm{emb}}, \alpha_{\mathrm{attn{\text -}residual}}, \alpha_{\mathrm{ffn{\text -}residual}})$, which we implement (assuming \depthmup\ branch scaling) as:
\begin{align*}
    a_l &= \begin{dcases}
        \frac{\alpha_{\mathrm{attn{\text -}residual}}}{\sqrt{L/2}} & l \textrm{ is odd (self-attention)}
        \\
        \frac{\alpha_{\mathrm{ffn{\text -}residual}}}{\sqrt{L/2}} & l \textrm{ is even (feed-forward)}
    \end{dcases}
    \\
    b_l &= 1
    \\
    c &= \alpha_{\mathrm{emb}}.
\end{align*}
The corresponding \umup\ set of residual HPs is $(\alpha_{\mathrm{res}}, \alpha_{\mathrm{res{\text -}attn{\text -}ratio}})$, which we implement as:
\begin{align} \label{eq:final_full_residual_multipliers}
    a^2_l &= \frac{\tau^2_l}{\tau^2_l + 1}
    \\
    b^2_l &= \frac{1}{\tau^2_l + 1}
    \\
    c &= 1, \\
    \\
    \tau^2_l &= \begin{dcases}
        \frac{\hat{\alpha}^2_a}{\frac{L}{2} + \ell \hat{\alpha}^2_a + \ell \hat{\alpha}^2_f} & l \textrm{ is odd}
        \\
        \frac{\hat{\alpha}^2_f}{\frac{L}{2} + (\ell + 1) \hat{\alpha}^2_a + \ell \hat{\alpha}^2_f} & l \textrm{ is even}
    \end{dcases}, \quad \ell = \left\lfloor{\frac{l-1}{2}}\right\rfloor
    \\
    \hat{\alpha}^2_a &= \alpha^2_{\mathrm{res{\text -}attn{\text -}ratio}}\,\hat{\alpha}^2_f\
    \\
    \hat{\alpha}^2_f &= \frac{2}{\alpha_{\mathrm{res{\text -}attn{\text -}ratio}}^2 + 1}\,\alpha^2_{\mathrm{res}}\,.
\end{align}
This is the \umup\ residual scheme. It satisfies the three properties that we initially set out to achieve: the variance at initialization of our $R_l(x)$ is always 1, our HPs have a clear and useful interpretation, and our scheme is as expressive as the baseline (which is neither unit-scaled or has interpretable HPs).


\section{The cut-edge rule} \label{app:cut_edge_rule}

In the section we review the notion of \textit{constraints} used for scaling operations in a computational graph. For a more thorough, generalized treatment, please refer to Section 5.1 and Appendix E.4 of the original Unit Scaling paper \cite{Unit_Scaling}.

For simplicity, we will only discuss the cut-edge rule in the context of a typical neural network. For each operation $f$, parametrized by $\theta$ taking input $x$ and emitting output $y$, a user must choose how to scale $y$, $\nabla_x$ and $\nabla_\theta$ (gradient of loss w.r.t. $x$ and $\theta$ respectively). In the simplest case, where there are no further data dependencies, we can simply choose factors that preserve unit scale. In more complex scenarios, we must balance the need for each tensor to be unit-scaled and for gradients to be correct up to a constant factor.

In particular, a problem emerges in the presence of residual blocks in which $y = x + f(x; \theta)$. In these circumstances, $\nabla_x$ is computed as the sum of residual gradient $\nabla_f$ $\frac{\partial{f}}{\partial{x}}$ and  skip gradient $\nabla_y$. If we choose not to insert scaling factors into our graph, $\nabla_f$ $\frac{\partial{f}}{\partial{x}}$ and $\nabla_y$ will have some ratio of scale $r$. However, if we have chosen to rescale the gradient of operations in $f$, then $\nabla_f$ $\frac{\partial{f}}{\partial{x}}$ will have been rescaled by some $s$. This means the new ratio of $\nabla_f$ $\frac{\partial{f}}{\partial{x}}$ and $\nabla_y$ will be $r \cdot s$. Therefore,  when adding these together, $\nabla_x$ is no longer a correct gradient up to a constant factor.

How do you remedy this? If we can ensure that the scaling factors are the same for both the input gradients and outputs of an op, we will have $s=1$. This ensures that gradients for inputs to residual blocks are correct up to a constant factor.

How do you decide when you are free to preserve unit scale, and when to constrain scaling factors to be the same? We previously define the \textit{cut-edge rule} \cite{Unit_Scaling} for computational graphs where nodes represent forward pass operations and edges represent operation outputs. If an input edge is a \textit{cut-edge}, i.e., the number of connected components in the graph would increase upon deletion (examples in a typical transformer model: output of embedding gather, output of a residual add, output of final norm, output token logits, weights), there is no need to constrain the scales of the operation's output edge and the input edge gradient. For all other input edges (e.g., inputs to a residual add, intermediates computed along a residual branch), the scales of gradients and outputs should be constrained.
\FloatBarrier

\section{From \mup\ to \umup} \label{app:from_mup_to_umup}

Here we outline additional details to help readers follow the process of deriving \umup\ from the combination of Unit Scaling and \mup. Our first step of dropping $\sigma_W$ and $\basefanin$, and moving $\alpha_W$s to functions, results in \Cref{table:mup_to_umup_1}. This intermediate scheme does not yet satisfy Unit Scaling, but simplifies the HP rules in preparation for further changes. Note that we have also removed $\hat \eta_\mathrm{emb}$ as we don't include this HP in our \umup\ extended HP set. We have included residual scaling rules here, in accordance with \depthmup, which we intend \umup\ to satisfy, though our standard \mup\ implementation doesn't use it.

\begin{table}[h]
  \centering
  \caption{An intermediate scheme resulting from dropping those HPs from \mup\ which are not needed under \umup.}
  \label{table:mup_to_umup_1}
  \begin{tabular}{l @{\hspace{0.8\tabcolsep}} lcccc}
      \toprule
      \multicolumn{2}{c}{\multirow{2}{*}[-0.2em]{ABC-multiplier}} & & Weight Type \vspace{0.2em} & & \multirow{2}{*}[-0.2em]{Residual}
      \\\cline{3-5}
      \rule{0pt}{1em} & & Input & Hidden & Output &
      \\
      \midrule
      parameter & ($A_W$) & $1$ & $1$ & $\frac 1 {\fanin}$ & $\frac 1 {\sqrt{\depth}}$\textsuperscript{*}
      \rule{0pt}{1.2em}\\[0.75em]
      initialization & ($B_W$) & $1$ & $\frac 1 {\sqrt{\fanin}}$ & $1 $ & ---
      \\[0.65em]
      Adam LR & ($C_W$) & $\eta$ & $\eta \, \frac 1 {\fanin}$ & $\eta$ & $\frac 1 {\sqrt{\depth}}$\phantom{\textsuperscript{*}}
      \\[0.45em]
      \bottomrule
      \vspace{-0.4cm}
      \\
  \end{tabular}
\end{table}

\section{Low-precision and its trade-offs} \label{app:low_precision_and_its_trade_offs}

\paragraph{Number formats for deep learning}

The standard numerical representations used in deep learning are the set of formats defined by the IEEE 754 floating-point standard~\citep{IEEE_754}. IEEE floats comprise three elements: a sign bit, exponent bits, and mantissa bits. The number of exponent bits determines the \textit{range} of a format, while the mantissa determines the \textit{precision}\footnotemark.
We refer readers to the original Unit Scaling paper (\citep{Unit_Scaling}, Section 3.1) for a comprehensive overview of floating-point representations.

\footnotetext{Confusingly, the term \textit{low-precision} tends to indicate using <32 bit-width formats, so in this context \textit{precision} also reflects the number of exponent bits as well as the usual mantissa bits.}

The default format used for training is the single-precision floating-point format, commonly known as FP32, with some hardware providers automatically casting it to the smaller TF32 compute mode for accelerated arithmetic. The 16-bit FP16 and BF16 formats were later introduced, and more recently the FP8 E5 \& E4 formats \citep{HFP8, 8_Bit_Numerical_Formats, FP8_Formats}. The higher range of E5 has typically been used for gradients, while the higher precision of E4 has been seen as necessary for weights and activations. Our particular implementation of FP8 training is covered in \Cref{sec:umup:low_prec_training}. Other aspects of training such as the optimizer state and cross-device communication have also been put into FP8 \citep{FP8-LM}, though not all tensors are amenable to being run in the lowest precision \citep{LLM_INT8} without degradation. The use of multiple formats is known as \textit{mixed precision} \citep{Mixed_Precision}. A comparison of these formats is given in \Cref{table:formats}.

\begin{table*}[ht]
  \centering
  \caption{A comparison of deep learning formats. E indicates exponent bits, and M mantissa bits. The smaller formats typically give more FLOPS, at the expense of reduced range and/or precision.}
  \label{table:formats}
  \begin{tabular}{lcccccc}
      \toprule
      Format & E & M & | max | & | min normal | & | min subnormal | & FLOPS (vs TF32)
      \\
      \midrule
      FP32 & 8 & 23 & $3.4 \times 10^{38}$ & $1.2 \times 10^{-38}$ & $1.4 \times 10^{-45}$ & $< 1 \times \hphantom{<}$
      \\
      TF32 & 8 & 10 & $3.4 \times 10^{38}$ & $1.2 \times 10^{-38}$ & $1.1 \times 10^{-41}$ & $1 \times$
      \\
      BF16 & 8 & 7 & $3.4 \times 10^{38}$ & $1.2 \times 10^{-38}$ & $9.2 \times 10^{-41}$ & $2 \times$
      \\
      FP16 & 5 & 10 & $65504$ & $6.1 \times 10^{-5\hphantom{0}}$ & $6.0 \times 10^{-8\hphantom{0}}$ & $2 \times$
      \\
      FP8 E5 & 5 & 2 & $57344$ & $6.1 \times 10^{-5\hphantom{0}}$ & $1.5 \times 10^{-5\hphantom{0}}$ & $4 \times$
      \\
      FP8 E4 & 4 & 3 & $448$ & $1.6 \times 10^{-2\hphantom{0}}$ & $2.0 \times 10^{-3\hphantom{0}}$ & $4 \times$
      \\
      \bottomrule
  \end{tabular}
\end{table*}

\paragraph{The benefits of low-precision}

Using numerical representations with fewer bits facilitates the design of more efficient arithmetic in hardware, typically leading to a linear increase in peak FLOPS (as shown in \Cref{table:formats}). As large-scale training efforts are typically compute-bound due to the size of matmuls \citep{Megatron-LM_Cluster}, putting the inputs to these operations in low-precision formats has a substantial impact on training efficiency. Low-precision formats also reduce the other two common performance constraints: for memory-bandwidth-bound models they require fewer bits to be transmitted, and for memory-size-bound models they require fewer bits to be stored.

\paragraph{The challenges of low-precision}

Unfortunately, moving to low-precision formats also increases \textit{quantization error}. For values within the representable range this takes the form of \textit{rounding error}, and for values outside it, \textit{clipping error} (both overflow and underflow). Rounding error tends to be an intrinsic problem: the number of mantissa bits dictates the expected accuracy of representations and this cannot easily be changed. In contrast, clipping error is often eliminated by scaling a tensor so that its values lie within the range of a format. Note that a multiplicative change in values of this kind doesn't affect the (relative) rounding error, due to the exponential spacing of values. Most research into making low-precision work has focused on the problem of scaling tensors in this way.

Simply casting all tensors to FP16 or FP8 tends to impair training, largely due to clipping error. For FP16, this primarily affects gradients. \citep{Mixed_Precision} address this by introducing a fixed global \textit{loss-scale} HP, which multiplies the loss value in the backward pass, artificially up-scaling gradients to lie within FP16 range \citep{Mixed_Precision}. \textit{Automatic loss scaling} \citep{OpenSeq2Seq} builds upon this idea, making the loss-scale a dynamic value that is tuned during training.

The later BF16 format has the same range as FP32, making loss scaling unnecessary. For FP8 no such range-equivalent format can exist, so the problem of clipping error must be addressed. Most FP8 implementations have done so by moving from a global loss-scale to a local scale for each FP8 tensor. In pseudo-code, this takes the form:

\codefig{scaled_matmul.py}

where we assume that \texttt{matmul} takes inputs in FP8 and directly produces the output in higher precision.

The result of the \texttt{scale()} operation can either be a fixed scale determined before training \citep{8_Bit_Numerical_Formats}, or in the case of Transformer Engine \citep{FP8_Transformer_Engine}, computed dynamically as a function of the `absmax' of the input tensor (though they introduce a delay across time-steps, to facilitate an efficient fused kernel). Increasing granularity and computing scales dynamically using this kind of method inevitably adds complexity (from both a logical and implementation perspective), as well the potential for computational overhead. Unit Scaling generally avoids the need for matmul input scaling.
\section{Benchmarking scaled matrix multiplication implementation in PyTorch} 
\label{app:scaled_mm_benchmarking}

Given that the end-goal of leveraging u-mup's low-precision properties is to speed up training and reduce memory usage, it's reasonable to ask why we don't investigate this experimentally. The answer relates to the relative immaturity of the FP8 training software stack - a lack of open, efficient FP8 kernels for compute and communication mean significant additional engineering effort is required to attain expected speedups over the full model.

Here we show that \umup's static scaling factors add no overhead to matmuls in FP8, and hence ought to be able to reach close to the maximal FP8 throughput attainable for the full model.

\begin{figure}[!h]
    \centering
    \begin{subfigure}{\textwidth}
        \centering
        \includegraphics[width=\textwidth]{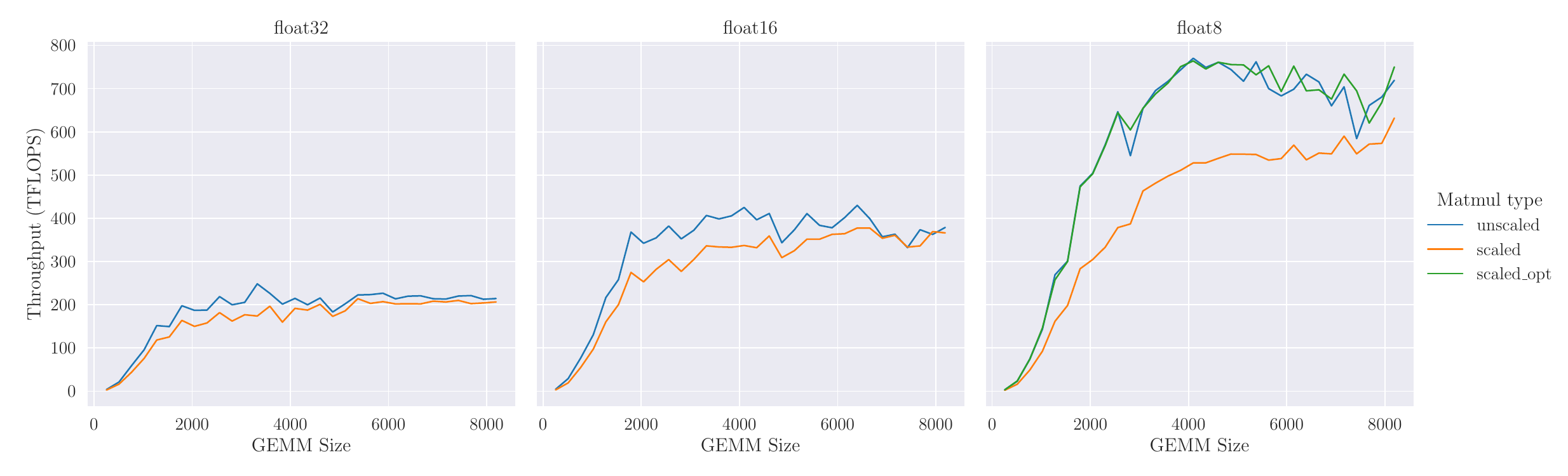}
    \end{subfigure}
    \caption{Square matrix multiplication throughput in TFLOPs with and without scaling factors applied to the output across 32-, 16-, and 8-bit float dtypes on NVIDIA H100 PCIe. Naive implementation in PyTorch.}
    \label{fig:scaled-mm-benchmarking}
\end{figure}

Standard strategies for FP8 training require expensive statistics gathering (e.g., amax) per tensor. A key benefit of \umup\ for FP8 training is that it instead provides us with static scaling factors to rescale operation outputs. Even a naive implementation in pytorch can achieve a minimal drop in hardware utilization.

Figure \ref{fig:scaled-mm-benchmarking} demonstrates hardware utilization for FP8, FP16, and FP32 matrix multiplications on a single NVIDIA H100 PCIe card. For FP16 and FP32, \texttt{torch.matmul} is used, whereas \texttt{torch.\_scaled\_mm} is used for FP8. Comparing "scaled" to "unscaled" matrix multiplication demonstrates a 30$\%$, 20$\%$, and 10$\%$ drop in hardware utilization for each data type respectively. In the case of FP8, where the drop in utilization is most pronounced, utilization can be recovered by passing the scaling factor as a scale associated with one of the two input tensors.

It should be noted that as of PyTorch version 2.3, \texttt{torch.\_scaled\_mm} always computes amax as well as the matrix multiplication. The performance of FP8 matrix multiplications could be higher without this overhead.

The above analysis focuses on throughput; significant memory savings are also possible through the use of FP8, though how this affects the total memory footprint depends on various additional variables and the overall distributed training setup. The following factors are play a significant role: typically the main memory bottlenecks are the optimizer states, which are kept in full precision. This footprint can be reduced by applying ZeRO sharding \citep{Zero}, though for significant gains the number of data parallel processes needs to be sufficiently large and ZeRO stage 2 or 3 are required. In these settings the memory footprint of activations and gradients becomes significant, and quantizing these to lower precision promises further memory savings, though may be non-trivial \citep{FP8-LM}.
\section{Attention output RMS grows with model depth} 
\label{app:per_tensor_rms_summary}

\begin{figure}[!h]
    \centering
    \begin{subfigure}{\textwidth}
        \centering
        \includegraphics[width=\textwidth]{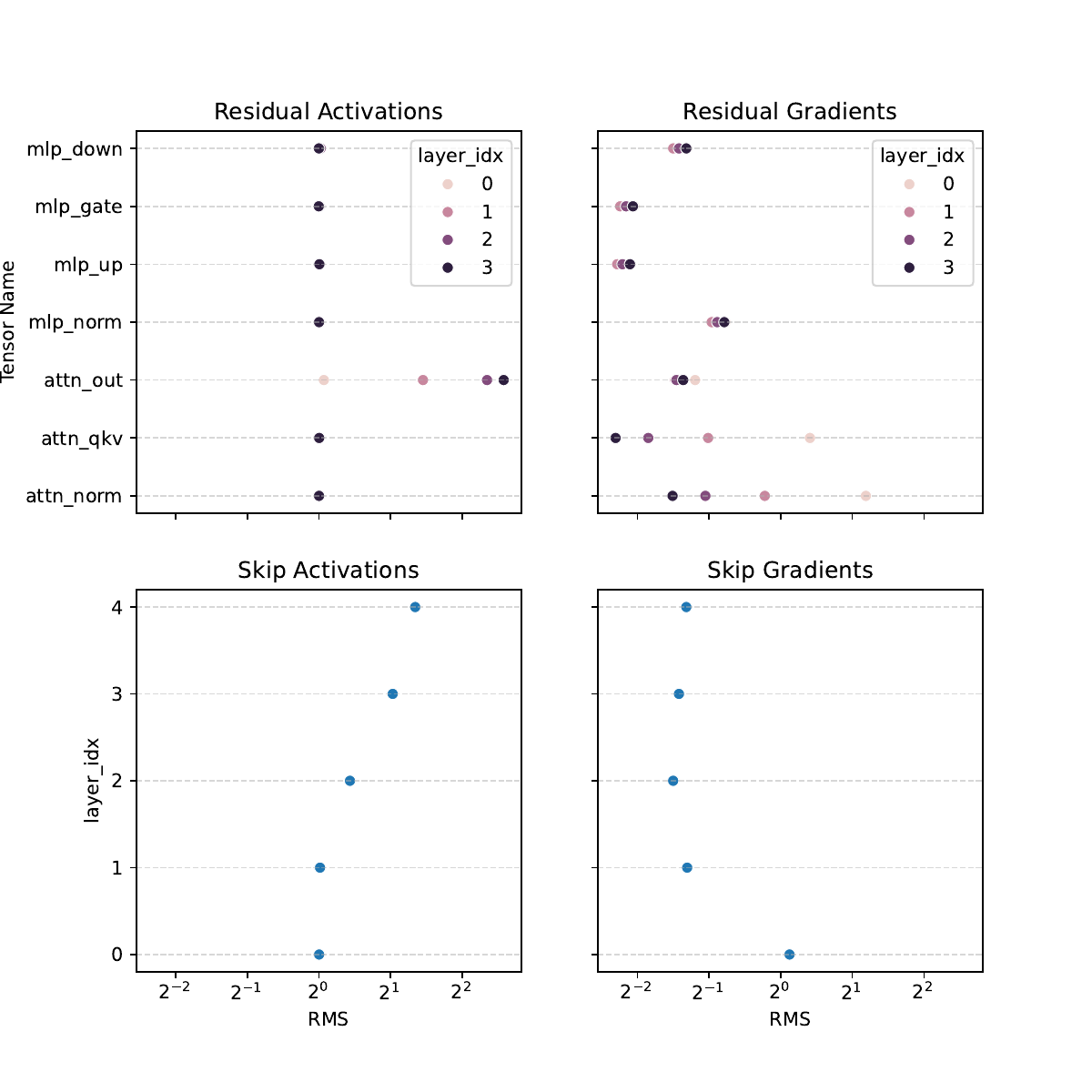}
    \end{subfigure}
    \caption{Scale of intermediate tensors grows with depth at initialization. Top left: Intermediate activation tensor RMS along the residual branch. Only the attention outputs after the first layer are not unit-scaled. Bottom left: Skip activation tensor RMS. Scale growth in attention outputs drives growth in skip activation scales. Note that \texttt{layer\_idx}$=0$ corresponds to the embedding output, and \texttt{layer\_idx}$=4$ corresponds to the final layer outputs. Top right: Intermediate gradient tensor RMS along the residual branch. Growth in the attention output scale drives growth in attention qkv gradient scales. Bottom Right: Skip gradient tensor RMS. The scale of output activations induces a global rescaling of the gradients.}
    \label{fig:acts-grads-rms}
\end{figure}

A core assumption in deriving per-op scaling factors is that each input to an operation has zero mean, unit-variance, and uncorrelated elements at initialization. This is trivially true for weights and by extension the token embeddings taken as input to the transformer trunk. However, this is not guaranteed for intermediate results and gradients if an operation in the computational graph induces correlation in the elements. In such a scenario our scaling factors will not return unit-variance outputs as we will not have corrected for these correlations in the inputs. As we then increase the depth of the network, where the same operation is left to amplify correlations, we can end up with variance in intermediate results and gradients scaling with depth

Figure \ref{fig:acts-grads-rms} illustrates this phenomenon in a unit-scaled four-layer Llama model with width=256. All activation tensors in the residual branches are unit-scaled, except for the output of the attention layers. We also see that the variance of attention outputs grows with depth. Since Llama models use pre-norm on the residual-branch, residual-branch inputs will revert to unit-scale again until they reach another instance of the correlation-inducing operation. As we add under-scaled attention layer results back to the skip-branch, our skip tensor variances grow with depth as our residual-add assumes unit-variance inputs. This has a knock-on effect on the global scaling of the gradients since the Jacobian of the final norm will scale the gradient by the inverse of the final skip tensor variance.

So which operation induces correlation in the attention output at initialization? For the default case where all multipliers are set to 1, our $1/d$ scaling of attention logits results in a sufficiently high temperature that attention probabilities are effectively uniform. With causal masking, we effectively take a running mean across the value tensor along the sequence dimension. As a result, each subsequent token representation is correlated with the last. Since we derive appropriate scaling factors for the first layer, we do not see scale growth emerging until the second layer, where correlations accumulate during the next effective running mean.

We leave it to future work to offer a solution to scale growth created by correlation in intermediate tensors. We note that this is scale growth emergent at initialization, but we also see scale growth in other intermediate tensors during training. Whether scale growth during training is related to the phenomenon outlined here remains to be seen.

\end{document}